%% file: neurips_2025.tex
\documentclass{article}
\pdfoutput=1

\usepackage[utf8]{inputenc} %
\usepackage[T1]{fontenc}    %
\usepackage{hyperref}       %
\usepackage{url}            %
\usepackage{booktabs}       %
\usepackage{amsfonts}       %
\usepackage{nicefrac}       %
\usepackage{microtype}      %
\usepackage{xcolor}         %
\usepackage{comment}
\usepackage{ulem}
\usepackage{wrapfig}

\usepackage{enumitem}
\usepackage{multirow}
\usepackage{subfig}
\usepackage{graphicx}
\usepackage{float}
\usepackage{amsmath}
\usepackage{arydshln} 
\usepackage{xcolor}   
\usepackage{array}
\usepackage{tabularx}
\usepackage{subcaption}
\usepackage[subtle]{savetrees}

\input{preamble}

\input{math_commands}

\definecolor{darkgreen}{RGB}{0,128,0}

\newcommand{\weaver}{\textsc{Weaver}}

\title{Shrinking the Generation-Verification Gap\\with Weak Verifiers}

\author{%
\centering
\begin{tabular}[]{@{}c@{}}
Jon Saad-Falcon$^{\dag}$\thanks{Equal contribution, Corresponding Authors: \texttt{<jonsaadfalcon,kelly.buchanan,mfchen>@stanford.edu}} , E. Kelly Buchanan$^{\dag *}$, Mayee F. Chen$^{\dag *}$, \\
Tzu-Heng Huang$^\ddag$,
Brendan McLaughlin$^\dag$, Tanvir Bhathal$^\dag$, Shang Zhu$^\S$, Ben Athiwaratkun$^\S$,\\ 
Frederic Sala$^\ddag$, Scott Linderman$^\dag$, Azalia Mirhoseini$^\dag$, Christopher Ré$^\dag$\\\\
$^\dag$ Stanford University\\
$^\ddag$ University of Wisconsin-Madison\\
$^\S$ Together AI
\end{tabular}
}

\newif\ifcomments
\commentstrue
\ifcomments\newcommand{\comments}[1]{#1}\else\newcommand{\comments}[1]{}\fi

\definecolor{clrgp}{rgb}{.9,0,.9}
\definecolor{red}{rgb}{.8,0,0}
\definecolor{blue}{rgb}{0,0, 0.8}
\definecolor{gray}{rgb}{0.41, 0.41, 0.41}
\definecolor{forestgreen}{rgb}{0.13, 0.55, 0.13}
\definecolor{subtle}{RGB}{152,78,163}

\usepackage{tcolorbox}

\usepackage{cleveref}

\usepackage{verbatim}
\usepackage{afterpage}

\AtBeginDocument{%
  \afterpage{\setlength{\textfloatsep}{0pt}}%
}

\begin{document}

\maketitle

\input{sections/abstract}

\input{sections/introduction}

\input{sections/related_work}

\input{sections/preliminaries}

\input{sections/methods}

\input{sections/experiments}

\input{sections/discussion}

\input{sections/acknowledgements}

\bibliography{references}
\bibliographystyle{plain}

\input{sections/appendix}

\end{document}

%% file: preamble.tex
\usepackage[capitalize,noabbrev]{cleveref}
\usepackage[textsize=tiny]{todonotes}
\usepackage{mathtools}  %
\usepackage{multirow}
\usepackage{tabularx} %
\usepackage{natbib}
\usepackage[margin=1in]{geometry}

\crefname{equation}{Eq.}{Eqs.}
\crefname{figure}{Fig.}{Figs.}
\crefname{section}{Sec.}{Secs.}
\crefname{subsection}{Sec.}{Secs.}
\crefname{theorem}{Thm.}{Thms.}
\crefname{appendix}{Appx.}{Appx.}
\crefname{lemma}{Lemma}{Lemmas}
\crefname{algocf}{Alg.}{Algs.}
\Crefname{algocf}{Algorithm}{Algorithms}
\hypersetup{
  colorlinks,
  linkcolor={blue},
  citecolor=blue,
  urlcolor=blue,
  breaklinks=true,   %
  linktoc=all,
}

%% file: math_commands.tex
\usepackage{amsmath,amsfonts,bm}

\def\eqref#1{equation~\ref{#1}}

\def\1{\bm{1}}

\DeclareMathAlphabet{\mathsfit}{\encodingdefault}{\sfdefault}{m}{sl}
\SetMathAlphabet{\mathsfit}{bold}{\encodingdefault}{\sfdefault}{bx}{n}

\newcommand{\Dtest}{\mathcal{D}^{\text{test}}}
\newcommand{\Ddev}{\mathcal{D}^{\text{dev}}}

\newcommand{\V}{\mathcal{V}}

\newcommand\SmallMatrix[1]{{%
  \tiny\arraycolsep=0.3\arraycolsep\ensuremath{\begin{bmatrix}#1\end{bmatrix}}}}

%% file: sections/abstract.tex
\begin{abstract}

Verifiers can improve language model (LM) capabilities by scoring and ranking responses from a pool of generated candidates. 
Currently, high-quality verifiers are either unscalable (e.g., humans) or limited in utility (e.g., tools like Lean for formal proofs). 
While LM judges and reward models have become broadly useful as general-purpose verifiers, a significant performance gap remains between them and oracle verifiers (i.e. verifiers with perfect accuracy).
To help close this gap, we introduce \weaver{}, a framework for designing a strong verifier by combining multiple weak, imperfect verifiers. 
First we find that weighted ensembles of verifiers, which typically require learning from labeled data, significantly outperform unweighted combinations due to differences in verifier accuracies. To reduce the dependency on labeled data, \weaver{} leverages weak supervision to estimate each verifier’s accuracy and combines their outputs into a unified score that better reflects true response quality. 
However, directly applying weak supervision algorithms poses several challenges, including inconsistent verifier output formats and handling low-quality verifiers. \weaver{} addresses these challenges by using dataset statistics to normalize outputs and filter specific verifiers.
We study the effectiveness of \weaver{} in test-time repeated sampling settings, where a model generates multiple candidate responses and selects one from among them. 
Our evaluations demonstrate that \weaver{} significantly improves over $Pass@1$---the performance when simply selecting the first candidate response---across several reasoning and math tasks, achieving o3-mini-level accuracy with Llama 3.3 70B Instruct (a much cheaper non-reasoning model) as the generator, and an ensemble of 70B or smaller judge and reward models as the verifiers (87.7\% average). This gain mirrors the jump achieved 
between GPT-4o and o3-mini (69.0\% vs. 86.7\%), which required extensive finetuning and post-training interventions. 
To reduce the computational costs of running verifier ensembles for \weaver{}, we train a compact 400M cross-encoder using \weaver{}'s combined output scores. This distilled model retains 98.7\% of \weaver{}’s full accuracy while reducing verification compute by up to 99.97\%.

\end{abstract}

%% file: sections/introduction.tex
\section{Introduction}
\label{sec:introduction}

A core challenge in deploying language models (LMs) is \textit{verification}: determining the quality or correctness of a model's response. 
This problem arises across various components of the LM  pipeline, including dataset curation, model alignment, and inference-time decision-making.  
Verification relies on \textit{verifiers}—functions that score responses. 
When combined with repeated sampling---generating multiple candidate responses from a LM---a perfect verifier can be used to select a correct candidate response, significantly enhancing model capability on tasks such as math, code, and reasoning \citep{snell2024scalingllmtesttimecompute, brown2024largelanguagemonkeysscaling, puri2025probabilisticinferenceapproachinferencetime}. 
For example, Llama 3.1 8B Instruct can match Llama 3.1 70B Instruct and even GPT-4o performances on MATH500 \citep{hendrycks2021measuring} and MiniF2F \citep{zheng2022minif2fcrosssystembenchmarkformal} when paired with perfect verifiers for these mathematics tasks.  
However, without a perfect verifier, a \textit{generation-verification gap} emerges \citep{song2025mindgapexaminingselfimprovement}: a LM can generate a correct response, but we fail to identify it. 

The generation-verification gap is prevalent across many tasks across mathematics, coding, scientific reasoning, instruction-following, and more.
For some of these settings, we have access to \textit{oracle verifiers} that can perfectly identify correct responses. A prominent example is Lean, a formal theorem prover that can be used for problems such MiniF2F \citep{zheng2022minif2fcrosssystembenchmarkformal}. However, this is often a limited setup, as not all mathematical proofs can be processed by Lean. 
Alternatively, humans could judge LM responses but manual evaluation is often expensive, noisy, and difficult to scale \citep{hosking2024humanfeedbackgoldstandard, clark2021thatshumangoldevaluating, karpinska2021perilsusingmechanicalturk}.
In contrast, LMs prompted as judges \citep{chiang2024chatbot} and reward models \citep{lambert2024rewardbenchevaluatingrewardmodels, singhi2025solveverifycomputeoptimalproblem, liu20251bllmsurpass405b} can be applied off-the-shelf to tasks like mathematics, coding, scientific reasoning, instruction-following \citep{hendrycks2021measuring, rein2024gpqa, jain2024livecodebenchholisticcontaminationfree, alpaca_eval}. 
However, these \textit{weak verifiers} produce noisy, inconsistent scores, often exhibit poor calibration, and suffer from high false positive rates \citep{stroebl2024inferencescalingflawslimits}. 
We ask: \textit{to what extent can we leverage weak verifiers to improve accuracy in the repeated sampling regime?}

We explore \textit{scaling verification}, specifically how to combine \textit{multiple weak verifiers} to improve response selection for repeated sampling.
As new pre-trained models become available, the pool of weak verifiers continues to expand and offer diverse, complementary sources of signal that could improve response selection if they can be aggregated effectively.
Recent work has explored scaling verification through techniques such as self-verification or averaging LM judge scores \citep{lifshitz2025multiagentverificationscalingtesttime, zhao2025sample, chen2025sets} although other work has found limitations to scaling test-time compute when utilizing weak verifiers for response selection \citep{stroebl2024inferencescalingflawslimits}.  
We observe three key challenges towards ensembling weak verifiers:

\input{tables_and_figures/main_figure}

\begin{enumerate} [itemsep=0.00pt,topsep=0pt,leftmargin=12pt]
    \item \textbf{Naively aggregating weak verifiers is insufficient for reliable verification.} Weak verifiers such as LM-based judges or reward models produce noisy, biased, and poorly calibrated scores, leading to inconsistent performance. \citep{stroebl2024inferencescalingflawslimits, lambert2024rewardbenchevaluatingrewardmodels, chiang2024chatbot}. 
    While using a naive unweighted average of verifier scores is straightforward, it implicitly assumes uniform verifier quality, causing low-quality verifiers to dominate and degrade the overall accuracy~\citep{verga2024replacingjudgesjuriesevaluating, xu2024perfectblendredefiningrlhf, eisenstein2023helping}. Moreover, while previous work has hypothesized that more sophisticated weighted ensembles should perform better, this claim has not been studied~\citep{lifshitz2025multiagentverificationscalingtesttime}. 
   
    \item \textbf{Effective ensembling with limited labeled data is challenging.} More sophisticated ensembling techniques typically learn verifier weights from labeled data, but such data is expensive and difficult to obtain. 
    \textit{Weak Supervision} (WS), a family of statistical techniques developed for data labeling, offers a potential solution through algorithms that aggregate multiple weak signals---such as crowd-worker annotations and expert-defined heuristics---while only requiring a small amount of labeled data \citep{ratner2016data, ratner2019training, fu2020fast}.  
    In traditional WS, practitioners can design and shape each weak signal to ensure sufficient quality (i.e., iteratively tweaking program-based heuristics), and guarantees of WS hinge on a baseline level of quality.
    Our weak signals, however, are fixed pre-trained language model verifiers, which have wildly varying accuracy---especially when applied to out-of-distribution tasks---and can emit incompatible outputs (logits, binary scores, Likert scores)~\citep{lambert2024rewardbenchevaluatingrewardmodels} that we cannot easily tweak. Due to these conditions, WS algorithms may not perform well when directly applied to verification.
    \item \textbf{Verification is expensive to deploy at inference.}
    Verification can dominate inference-time costs \citep{singhi2025solveverifycomputeoptimalproblem, liu20251bllmsurpass405b}, since each verifier must process both the problem and its candidate response(s) \cite{lightman2023letsverifystepstep}, often evaluating intermediate steps \cite{lightman2023letsverifystepstep} and multiple solution paths \cite{snell2024scalingllmtesttimecompute}. 
    In fact, achieving gains over unverified generation (i.e. majority voting) can require $10\times$ to $128\times$ the inference compute per query \citep{singhi2025whentosolve, lifshitz2025multiagentverificationscalingtesttime, zhao2025sample, chen2025sets}. 
    
\end{enumerate}

In this work, we introduce \weaver{}, a framework for aggregating weak verifiers without supervised finetuning on ground truth labels (Figure \ref{fig:main_figure}). 
First, we demonstrate that if we have access to a large corpus of labeled training data (e.g., 50,000 query-response pairs), we can learn weighted ensembles that can outperform naive averaging by up to 11.2\% points. This is because weighted ensembles take advantage of wide variability in verifier accuracy. However, in many real-world scenarios, we do not have access to such quantities of labeled data.
Second, to reduce the dependency on labeled data, we adapt Weak Supervision to the verification setting by addressing challenges around inconsistent outputs and low-accuracy verifiers. 
\weaver{} filters out uninformative verifiers, normalizes verifier scores, and builds a latent variable model over these scores and the unknown true labels to estimate the verifier accuracies to be used as weights for the ensemble~\citep{ratner2016data, hall2003generalized}.

Empirically, given a repeated sampling budget and a set of verifiers, \weaver{} improves over repeated sampling with unweighted averaging of verifier scores by 17.1\% and with majority voting by 13.5\% (Table \ref{tab:verifier_ablations}; Figure \ref{fig:fp_and_pass@1}).
Compared to an LM's $Pass@1$, \weaver{} allows us to improve performance by 17.9\% for 8B models and 14.5\% for 70B models across reasoning and mathematics tasks (Tables \ref{tab:verifier_ablations} and \ref{tab:8B_verifier_ablations}).
\textit{This mirrors the performance jump from GPT-4o to o3-mini} (73.9\% vs. 88.2\%)---but only via increased sampling at test time rather than parameter tuning or post-training procedures. 
We also study how \weaver{} scales along different axes of test-time compute: generation, verifiers, model size, and inference budget (Section~\ref{sec:scaling_verification_with_weaver}). 
We find that even as we increase the number of generations, many standard verification baselines (e.g. majority voting) quickly plateau (Figure \ref{fig:fp_and_pass@1}). 
Naive ensembling saturates more slowly, but its gains are limited by sensitivity to the model choice and the number of verifiers. 

Finally, to mitigate the compute costs of calling multiple weak verifiers for each response, we extend \weaver{} by training a 400M-parameter cross-encoder verifier using \weaver{}'s selected responses.
We demonstrate that using a distilled \weaver{} cross-encoder as a verifier \textit{retains 98.7\% of the accuracy gains} from the learned verifier ensemble  while reducing compute costs by three orders of magnitude -- \textit{saving 99.97\% inference FLOPS} while still capturing an effective verification strategy (Section \ref{sec:weaver_distillation}). 
Overall, our findings highlight that more reliable, scalable verification is possible even in the absence of ground-truth labels---paving the way for improved data filtering, model alignment, and inference-time decision-making.

%% file: tables_and_figures/main_figure.tex
\begin{figure}[t]%
   \centering
   \includegraphics[width=1.0\linewidth]
{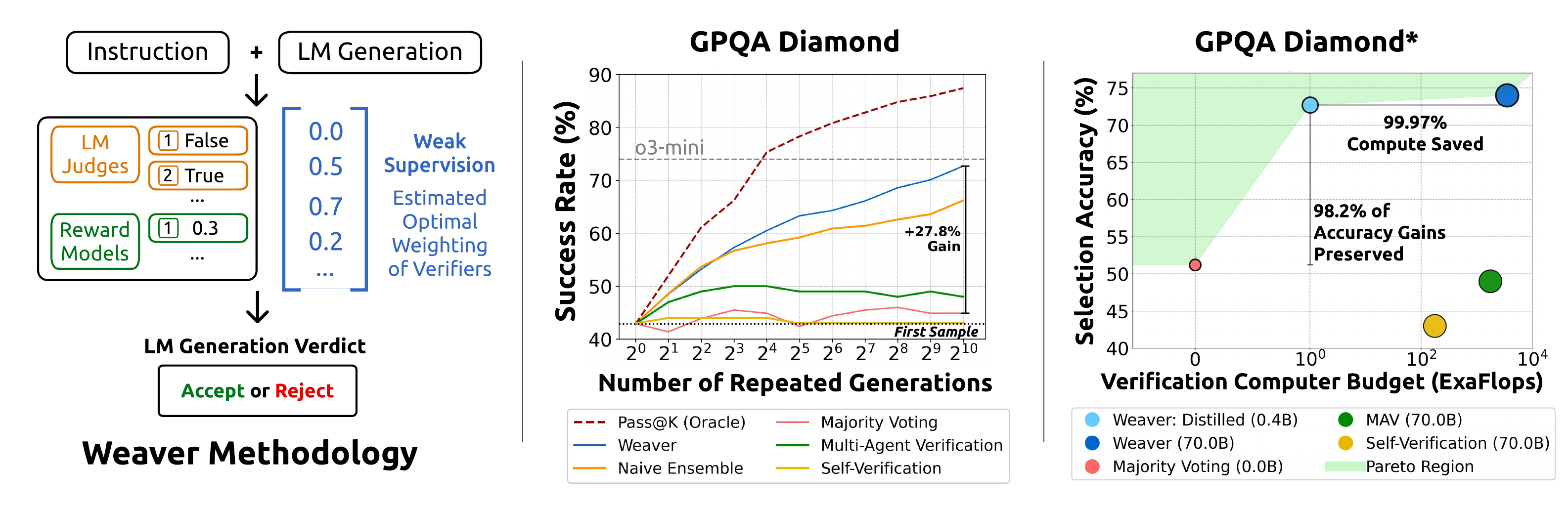}
   \caption{
   \textbf{\weaver{} Framework}: 
   We propose \weaver{}, a framework combining multiple weak verifiers to effectively scale repeated sampling without parameter finetuning on ground truth labels \textbf{(left)}. %
   \weaver{} significantly outperforms majority voting and shrinks a model's \textit{generation-verification gap} by 14.5\%, on average, for GPQA Diamond and other datasets (Table \ref{tab:verifier_ablations}) \textbf{(middle)}. 
   By distilling \weaver{} from an ensemble of 70B verifiers to a single 400M cross-encoder, we can preserve 98.2\% of the accuracy gains of \weaver{} while reducing inference compute cost by 99.97\% \textbf{(right)}. 
   }
   \label{fig:main_figure}
\end{figure}

%% file: sections/related_work.tex
\section{Related Work}
\label{sec:related_work}

\noindent\textbf{LM Judges and Reward Models}: 
Both LM judges and reward models are promising approaches for evaluating language model outputs, but their high false positive rates limit their reliability \citep{stroebl2024inferencescalingflawslimits}. 
LM judges can evaluate outputs without additional training \citep{liu2023g, wang2023chatgpt, fu2023gptscore}, using approaches from simple prompting to chain-of-thought reasoning \citep{liu2023g} to specialized fine-tuning \citep{saad2023ares, tang2024minicheckefficientfactcheckingllms} to multi-LM inference architectures \citep{saad2024archon, kalra2025verdictlibraryscalingjudgetime}. However, they face {poor generalization} across contexts \citep{es2023ragas, saad2023ares, ravi2024lynxopensourcehallucination} and {systematic biases} in position and self-preference \citep{chen2024humans, pan2024human, zheng2023evaluation}. 
Similarly, while reward models have become central to model alignment \citep{bradley1952rank, christiano2017deep, skyworkreward2024}, they struggle with {noisy training signals} from low inter-annotator agreement \citep{askell2021general, ouyang2022training, wang2024secretsrlhflargelanguage, dubois2024alpacafarm} and {learned biases} favoring attributes like response length \citep{lambert2023alignment, singhal2023long, dubois2024length}. 
Recent work has improved individual verifier reliability through better data collection, chain-of-thought reasoning, and natural language unit tests \citep{wang2023helpsteermultiattributehelpfulnessdataset, zhang2024generativeverifiersrewardmodeling, saadfalcon2024lmunitfinegrainedevaluationnatural}, yet fundamental challenges persist \citep{eisenstein2023reward, chaudhari2024rlhfdecipheredcriticalanalysis}. 
\weaver{} advances beyond these approaches by combining multiple verification signals with adaptive weighting, thus leveraging the complementary strengths of weak verifiers while suppressing noise and reducing false positives. 

\noindent{}\textbf{Weak Supervision}: \weaver{} builds upon statistical techniques from weak supervision, which emerged as a framework for programmatically generating training labels by aggregating multiple weak sources \citep{ratner2016data, ratner2020snorkel}. 
While a majority of the work focuses on classification tasks \citep{ratner2019training, fu2020fast, chen2022shoring}, recent advances have expanded to handle multi-task settings \citep{shin2021universalizing} and structured prediction \citep{vishwakarma2022lifting}.  Weak Supervision has also been applied to LM prompting~\citep{arora2022askanythingsimplestrategy} and routing~\citep{guhasmoothie}.
\weaver{} applies Weak Supervision to answer verification, treating binary imperfect verification signals (e.g. reward models and LM judges) as weak supervision voters that classify candidate solutions as correct or incorrect. 
This novel application combines predictions by converting these diverse signals into binary verdicts, enabling \weaver{} to learn better verification strategies from weak but complementary verifiers.

\noindent{}\textbf{Verification as another compute axis and aggregation:}
Recent work has explored verification as a new scaling axis \cite{lifshitz2025multiagentverificationscalingtesttime,liu2025can,zhao2025sample,singhi2025whentosolve, stroebl2024inferencescalingflawslimits,chen2025sets}. However this work limits their analysis to one verifier, and instead scale how many times to verify \cite{zhao2025sample}. Approaches that do leverage multiple verifiers often rely on substantial amounts of labeled data for aggregation or creating specialized verifiers \cite{kirchner2024prover, lifshitz2025multiagentverificationscalingtesttime}. With \weaver{}, we show that it is possible to combine verifiers without ground truth labels, even when they are not specialized. Other work has focused on combining multiple verifiers for post-training the base model using RLHF \cite{wang2024transforming, eisenstein2023helping,wang2025improvingmodelalignmentcollective}.

%% file: sections/preliminaries.tex
\section{Preliminaries}
\label{sec:preliminaries}
First, we define the problem of how to select among repeated samples. We then define verifiers and key evaluation metrics, including the generation-verification gap. 

\textbf{Problem Definition} \;\; Let $q \in \mathcal{Q}$ be a input text query, and let $r \in \mathcal{R} \sim \mathcal{M}(q)$ be a corresponding response sampled from language model $\mathcal{M}$ with non-zero temperature. For a given query-response pair $(q, r)$, we define $y: \mathcal{Q} \times \mathcal{R} \rightarrow \{0, 1\}$ such that $y(q, r)$ is the correctness label of $r$ for $q$. 

We are given an unlabeled test dataset $\Dtest = \{(q_i, \bm{r_i})\}_{i=1}^n$, where $\bm{r_i} = \{r_{ij}\}_{j=1}^K$ consists of $K$ repeatedly sampled responses from $\mathcal{M}$ for each $q_i$. 
We also assume access to a small labeled development dataset $\Ddev \subset \Dtest$, comprising $1\%$ of the test set (e.g. 5 to 10 query-answer pairs), which is used to estimate global statistics such as the task difficulty probability, $\Pr(y_{ij} = 1)$.
We do not have access to true labels $y{ij} := y(q_i, r_{ij})$ for any $i, j$ in $\Dtest \setminus \Ddev$.

For each $(q_i, \bm{r}_i) \in \Dtest$, our goal is to select a correct response $j^\star \in [K]$ that satisfies $y_{ij^\star} = 1$. We can broadly describe this selection rule using a scoring function $f: \mathcal{Q} \times \mathcal{R} \rightarrow \mathbb{R}$, namely $j^\star := \arg\max_j f^\star(q_i, r_{ij})$. 

\textbf{Using verifiers} \;\; A verifier, either a reward model or an LM prompted as a judge, can be expressed as a scoring function on query-response pairs $v: \mathcal{Q} \times \mathcal{R} \rightarrow \mathbb{R}$. For reward models, the verifier score is continuous, while for LM judges, the verifier score is typically discrete (for our setup, we use $[0, 1]$ and $\{0, 1\}$, respectively). 
We assume that we have access to multiple verifiers $\V = \{ v_1, \dots, v_m\}$. We apply each of the $m$ verifiers to each $(q_i, r_{ij})$, for a total of $nmK$ scores on $\Dtest$, with $s_{ijk} := v_k(q_i, r_{ij})$. We aim to use $\mathcal{V}$ to construct a verification strategy $f$.

\textbf{Evaluation metrics} \;\; 
The $Pass@1$ metric is the probability that an LM's first response is correct.
$Pass@K$ generalizes this metric and is defined as the probability that there exists a correct response among $K$ generated responses: $Pass@K = \frac{1}{n} \sum_{i = 1}^n \mathbf{1}(\exists j \in [K]: y_{ij} = 1)$. 
This metric is independent of the verification strategy, and depends on the choice of $\mathcal{M}$, $K$, and the task dataset. 
The success rate of a verification strategy $\hat{f}$ is $\frac{1}{n} \sum_{i = 1}^n y_{i\hat{j}}$, where $\hat{j} = \arg\max_{j \in [k]} \hat{f}(q_i, r_{ij})$. Success rate is dependent on the verification strategy and bounded by Pass@K, and equality is obtained with oracle verification (i.e., $\hat{f} = f^\star$ can always select a correct $j$ as long as it exists). 
 
We define the \textit{generation-verification gap} as Pass@K - Success Rate. A large positive gap indicates that although correct answers are generated, the verification strategy fails to select them consistently. We aim to close this gap and will use it to evaluate verification strategies.

%% file: sections/methods.tex
\section{\weaver{}: A Framework for Weak Verifier Aggregation}
\label{sec:methods}

In Section~\ref{sec:weighted_vs_naive}, we demonstrate that naively averaging multiple verifier scores to select responses significantly underperforms weighted ensembles; however, common methods for computing weights require labeled data \citep{schapire2013explaining, ying2015decision}. 
We introduce \weaver{} (Section~\ref{sec:weaver_alg}), a method for weighted aggregation of verifier scores with minimal data that draws inspiration from Weak Supervision. 
Unlike prior work, \weaver{} adapts weak supervision to verification by addressing challenges unique to verifier aggregation, such as inconsistent score formats and the presence of low-quality or adversarial verifiers. 
To our knowledge, this is the first framework to successfully apply weak supervision to ensemble verifier scores for response selection.

\subsection{How to aggregate multiple verifiers: weighted vs unweighted ensembles} \label{sec:weighted_vs_naive}

A straightforward approach for using multiple verifiers is a naive ensemble---selecting the response with the highest average verifier score: $f(q_i, r_{ij}) = \frac{1}{m} \sum_{k=1}^m s_{ijk}$.
This approach~\cite{lifshitz2025multiagentverificationscalingtesttime} does not consider the relative accuracy of verifiers. However, we observed that there is significant variation in the success rates of individual verifiers---spanning a range of up to 37.5\%---suggesting that naive ensembles could be suboptimal (Table \ref{tab:verifier_accuracies_ranges}).

An alternative is to use a weighted ensemble. One approach is to use a labeled dataset to identify and use the top-performing verifier, effectively assigning a weight of $0$ to discarded verifiers. Other strategies include using Logistic Regression or a Naive Bayes classifier, where the scoring function $f(q_i, r_{ij})$ is the probability $\Pr(y_{ij} = 1 | s_{ij1}, \dots, s_{ijm})$. These classifiers are fit using labeled data and can be either modeled as a logistic function or factorized using Bayes' rule and independence assumptions, respectively. 

In Figure~\ref{fig:MV_vs_Naive_vs_Weighted}, we compare a naive ensemble with weighted ensembles for several tasks, using Llama 3.3 70B Instruct to generate responses and using a collection of 33 7B-72B reward models and LM judges as verifiers (Appendix \ref{app:models_and_datasets}). 
We see that using a weighted ensemble can achieve up to 11.2 points higher success rate than the naive ensemble. However, all weighted ensembles shown are ``oracle'' methods: they are computed using $y_{ij}$ for all $i\in [n], j \in [K]$, although in practice these labels are unknown for $\Dtest$. In fact, when we instead use $0.01 n$ labeled samples, accuracy drops by 20.1\% on average (Table \ref{tab:LR_and_NB_across_datasizes}). 
This raises the question of how to best construct weighted ensembles with limited labeled data.

\input{tables_and_figures/MV_vs_Naive_vs_Weighted}

\subsection{\weaver{}: weighted ensembling of verifier scores with minimal labeled data} \label{sec:weaver_alg}

We first describe the WS method we use in \weaver{} to construct a weighted ensemble over binary verifier scores. 
Because verifiers often produce scores in inconsistent formats and exhibit low accuracies---challenges not typically encountered in traditional WS---we introduce a binarization and verifier discarding strategy in Appendices \ref{app:practical_considerations} and \ref{app:implementation_decisions} to discard low-quality verifiers and ensure that only sufficiently reliable binary scores are used as input to the WS method.

\subsubsection{Weak Supervision Algorithm}
In Weak Supervision, the input is an unlabeled dataset, where each entry has multiple binary ``votes'' on the true label. Applied to our setting, each entry is a query-response pair, forming a dataset of size $nK$, and verifier scores $s_{ijk}$ are binarized into votes $\bar{s}_{ijk} \in \{0, 1\}$ for all $i, j, k$. Our goal is to predict the probability that a response is correct, 
$\Pr(y_{ij} = 1 | s_{ij1}, \dots, s_{ijm})$ for all $i, j$.

\paragraph{WS model} We can view all $y_{ij}$ across query-response pairs as samples of an unknown random variable $Y$ and each $\bar{s}_{ijk}$ across $i, j$ as samples of a random variable $S_k$. WS then defines a latent variable graphical model over the random binary vector $\{Y, S_1, \dots, S_m\}$, where $Y$ is latent while $S_1, \dots S_m$ are observable. While existing WS methods assume various models, one common assumption is that $S_i \perp S_j | Y$ for each $S_i, S_j$. That is, $S_i$ and $S_j$ are conditionally independent given $Y$; intuitively, each verifier is assumed to capture independent aspects of the correctness of the response (Figure \ref{fig:Verifiers_vs_Generations_Tradeoff} in Appendix \ref{app:scaling_candidate_generations}). 
Under this assumption, we can write the posterior probability of a correct generation as the following, for some given binary verifier scores $\{\bar{s}_1, \dots, \bar{s}_m\}$:
\begin{align}
\Pr(Y = 1 | S_1 = \bar{s}_1, \dots, S_m = \bar{s}_m) = \frac{\prod_{i = 1}^m \Pr(S_i = \bar{s}_i | Y = 1) \Pr(Y = 1)}{\Pr(S_1 = \bar{s}_1, \dots, S_m = \bar{s}_m)}.
\label{eqn:weaver_posterior}
\end{align}
The weighted ensemble score for each query-response pair can thus be written in terms of: 1) $\Pr(S_1 = \bar{s}_1, \dots, S_m = \bar{s}_m)$, which is intractable to compute from the data for large $m$; 2) $\Pr(Y = 1)$, which can be estimated from $\Ddev$; and 3) $\Pr(S_i = \bar{s}_i | Y = 1)$, or equivalently $\Pr(S_i = 1 | Y = 1)$, which is the verifier's ``accuracy parameter''---this cannot be computed directly since we do not have access to $Y$. 
Next, we discuss how to estimate these accuracy parameters, $\Pr(S_i = 1 | Y = 1)$, without labels.

\paragraph{WS parameter estimation} We outline a parameter estimation technique first introduced in \citep{ratner2020snorkel}. Due to the assumption that $S_i \perp S_j | Y$, the following equation holds:
\begin{align}
\Pr(&S_i, S_j) = \Pr(S_i, S_j | Y = 1) \Pr(Y = 1) + \Pr(S_i, S_j| Y = 0) \Pr(Y = 0) \nonumber \\
&= \Pr(S_i | Y = 1) \Pr(S_j| Y = 1) \Pr(Y = 1) + \Pr(S_i | Y = 0) \Pr(S_j | Y = 0) \Pr(Y = 0). \label{eq:indep}
\end{align}

Note that $\Pr(S_i, S_j)$ can be computed from the known verifier scores, and $\Pr(Y=1)$ is estimated from $\Ddev$. Then, \eqref{eq:indep} is a quadratic equation over the accuracy parameters. We can write this equation for every pair $S_i, S_j$, and for every pair of values $\{0, 1\}^2$ they can take. Furthermore, we can write another type of equation over the accuracy parameters:
\begin{align}
\Pr(S_i = 1) = \Pr(S_i = 1 | Y = 1)\Pr(Y = 1) + \Pr(S_i = 1 | Y = 0) \Pr(Y = 0). \label{eq:decomp}
\end{align}

This is a consistency property that holds regardless of the conditional independence assumption, and we can write this equation for each of the $m$ $S_i$'s. Because we know that the accuracy parameters should follow equations~\ref{eq:indep} and~\ref{eq:decomp}, we can construct an objective function that aims to minimize the difference between the left and right hand sides of these equations. We write this efficiently in matrix notation. Let $P \in \mathbb{R}^{2\times 2}$ be a diagonal matrix with diagonal $[\Pr(Y=0) \; \Pr(Y=1)]$. Define $\mu \in \mathbb{R}^{m \times 2}$ to be the matrix of accuracy parameters, and define $O \in \mathbb{R}^{2m \times 2m}$ to be a matrix over the joint probabilities of pairs of $S_i, S_j$; more formally:
\begin{align}
    &\mu_{2i-1:2i, 1:2} = \SmallMatrix{
        \Pr(S_i = 0 | Y = 0) & \Pr(S_i = 0 | Y = 1) \\ \Pr(S_i = 1 | Y = 0) & \Pr(S_i = 1 | Y = 1)
    }, \;\; O_{2i-1:2i, 2i-1:2i} = \SmallMatrix{
        \Pr(S_i = 0) & 0 \\ 0 & \Pr(S_i = 1)
    } \; \forall i \in [m] \nonumber \\
    &O_{2i-1:2i, 2j-1: 2j} = \SmallMatrix{
        \Pr(S_i = 0, S_j = 0) & \Pr(S_i = 0, S_j = 1) \\ \Pr(S_i = 1, S_j = 0) & \Pr(S_i = 1, S_j = 1)
    } \; \forall i \neq j \in [m]
\end{align}

Let $\text{off-diag}$ denote the elements of a matrix that lie outside its $2\times 2$ block diagonal. Then, to estimate $\mu$ that satisfies both equations~\ref{eq:indep} and~\ref{eq:decomp}, we have the following objective:
\begin{equation}
 \text{minimize}_{\mu} \bigl\|\,O_{\text{off-diag}} - (\mu\,P\,\mu^T)_{\text{off-diag}}\bigr\|^2 + \bigl\|\,\mathrm{diag}(O) - \mu\,P\,\mathbf{1}^T\bigr\|^2
\label{eqn:ws_loss}
\end{equation}
We optimize~\ref{eqn:ws_loss} using gradient descent to estimate the verifier accuracy parameters. These estimates are then used in \cref{eqn:weaver_posterior} to select the response with the highest estimated posterior. 
To further improve modeling of verifier accuracies, we explore whether partitioning the query distribution by empirical difficulty can yield better weak supervision estimates. 
As detailed in Appendix~\ref{app:clustering_exploration}, we cluster queries based on the observed ratio of correct to incorrect generations, and fit a separate \weaver{} model within each difficulty bucket.
We provide more details in \cref{app:ws_discrete_knowndiff}.

%% file: tables_and_figures/MV_vs_Naive_vs_Weighted.tex
\begin{figure}[H]
   \centering
   \includegraphics[width=\linewidth]{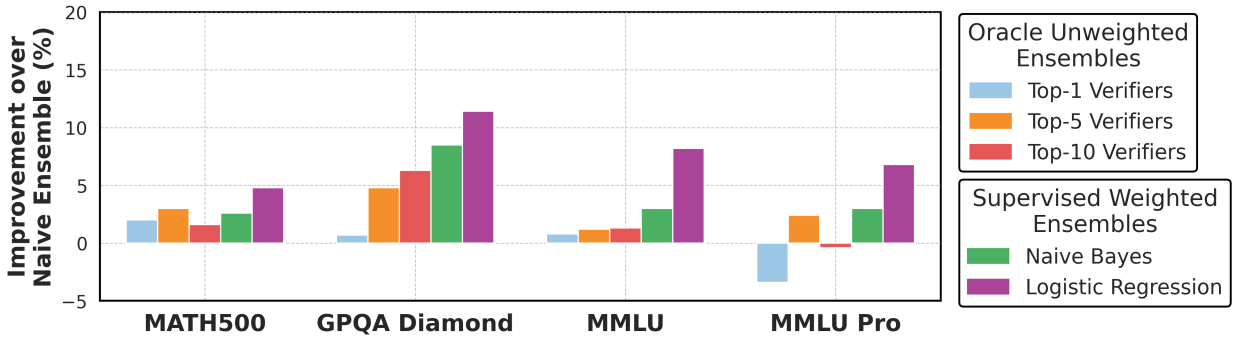}
   \caption{\textbf{Weighted Verifier Ensembles Outperform Naive Verifier Ensembles}: 
   By using oracle data to keep the best verifiers (i.e. \textit{top-$K$ verifier ensembles}) or learn aggregation weights for verifiers (i.e. \textit{supervised weighted ensembles}), we can improve beyond naive combinations of the verifiers available by 3.6\% and 7.8\%, on average, respectively.
   }
   \label{fig:MV_vs_Naive_vs_Weighted}
\end{figure}

%% file: sections/experiments.tex
\section{Results}
\label{sec:results}

In section~\ref{sec:weak_verifier_ensembling}, we provide empirical results on \weaver{}'s performance compared to other approaches for selecting responses in repeated sampling. In section~\ref{sec:scaling_verification_with_weaver}, we study how \weaver{}'s performance scales along several axes: the number of responses, model size, verifier counts, and inference compute. 

\textbf{Datasets, Verifiers, and Baselines} \;\; Our reward models range in size from 8B to 72B, are all open-source, and are obtained from RewardBench \cite{lambert2024rewardbenchevaluatingrewardmodels}, a popular evaluation tool for reward models. We prompt open-source language models from Chatbot Arena \citep{chiang2024chatbot} to serve as judges. Unless specified, we use Llama 3.3 70B Instruct to generate responses and use all 33 reward models and judges. We evaluate on MATH500, GPQA Diamond, MMLU College, and MMLU Pro. See~\cref{app:models_and_datasets} for more details.

We compare \weaver{} against verifier-free baselines as well as standard verification strategies. First Sample, also known as Pass@1, only uses the first response and does not scale test-time compute or verification. Majority Voting involves repeated sampling but not verification, picking the most common final answer from the responses \citep{brown2024largelanguagemonkeysscaling, snell2024scalingllmtesttimecompute, chen2024more}. We compare against the highest scoring reward model and a naive ensemble of the top-10 reward models on RewardBench. We also evaluate two recently proposed methods that scale verification but do not use different verifier models or weighted ensembles: Self-Verification~\citep{zhao2025sample} and Multi-Agent Verification~\citep{lifshitz2025multiagentverificationscalingtesttime}. Lastly, we report the oracle Pass@K rate, which establishes an upper bound for the success rate of these verification strategies. 

\subsection{\weaver{} Shrinks the Gap with Frontier LMs}
\label{sec:weak_verifier_ensembling}
In \cref{tab:verifier_ablations}, we evaluate \weaver{} along with baseline verification methods, the first sample performance of frontier LMs, and the Pass@100 metric. We use LlaMA 3.3 70B Instruct to generate $K=100$ responses per query.
We find that \weaver{}'s weighted ensembling of multiple verifiers allows us to outperform majority vote by $15.5\%$ and come within $4.2\%$ of the Pass@100 oracle metric. 
Furthermore, \weaver{} rivals the performance of frontier reasoning models---coming within $0.5\%$ of OpenAI's o3-mini~\citep{OpenAIo3mini2025}---even though we use a non-reasoning model for generation.

\input{tables_and_figures/verifier_ablations}

\subsection{\weaver{} Improves Compute-Accuracy Trade-Off for Scaling}
\label{sec:scaling_verification_with_weaver}
By proposing to combine multiple weak verifiers instead of one, we introduce yet another axis for test-time scaling. In this section, we study how well scaling verification with \weaver{} interacts with common previously studied axes for verification, summarized in \cref{tab:scaling_axes}. 

\begin{table}[h]
\centering
\caption{\textbf{Scaling Dimensions for Generation and Verification Models}}
\newcommand{\colAwidth}{0.3\textwidth}
\newcommand{\colBwidth}{0.2\textwidth}
\newcommand{\colCwidth}{0.28\textwidth}
\newcommand{\colDwidth}{0.1\textwidth}
\small
\setlength{\tabcolsep}{4pt} 
\begin{tabular}{
    >{\raggedright\arraybackslash\hspace{0pt}}p{\colAwidth}
    >{\raggedright\arraybackslash\hspace{0pt}}p{\colBwidth}
    >{\raggedright\arraybackslash\hspace{0pt}}p{\colCwidth}
    >{\raggedright\arraybackslash\hspace{0pt}}p{\colDwidth}
}
\toprule
\textbf{Scaling Dimension} & \textbf{Base Model} & \textbf{Verifier Type} & \textbf{Visuals} \\
\midrule
\textbf{Sample Count}: More Generations
& Temperature-based sampling 
& Majority Vote, Weak Verifier, Top-K, \weaver{} 
& \autoref{fig:fp_and_pass@1} \\
\textbf{Model Size}: Larger Models 
& Llama 8B → 70B
& RM-8B → RM-70B
& \autoref{tab:model_size_ablation} \\
\textbf{Verifier Count:} More Models
& Llama 8B/70B
& RMs and LM Judges
& \autoref{fig:Weaver_Verifier_Scaling} \\ 
\textbf{Inference Compute:} More FLOPs for Gen./Ver.
& Temp-based sampling
& Weak Verifiers + \weaver{} 
& \autoref{fig:Weaver_Scaling_Laws} \\
\bottomrule
\end{tabular}
\label{tab:scaling_axes}
\end{table}

\noindent \textbf{(1) Scaling Candidate Generations}: we study the performance of verification methods as we increase the number of repeated samples in~\cref{fig:fp_and_pass@1}. 
Based on prior work~\cite{bradley1952rank, chen2021evaluating}, as the number of responses increases, we are more likely to see a correct response (i.e. Pass@K increases), and hence more likely to select a correct response given a good verification strategy. However, differences in verification translate into different scaling rates. 
We evaluate the performance of \weaver{} and baselines for $K = 2^0$ to $2^{10}$, comparing to o3-mini and Pass@K as well.
Across all tasks, \weaver{} yields the most substantial gains when scaling the number of generations. 
\weaver{} consistently narrows the generation-verification gap with the oracle upper bound (Pass@K) while alternative verification strategies plateau after a few generations. 
The effect is particularly pronounced on difficult tasks like GPQA. 
We detail the scaling trends observed in \cref{fig:fp_and_pass@1} in \cref{app:scaling_trends_of_weaver}.

\input{tables_and_figures/FP_and_Pass1}

\noindent \paragraph{(2) Scaling Model Sizes:}    
In~\cref{tab:model_size_ablation}, we study how \weaver{} applied on smaller models (both verifiers and for generating responses) can allow us to match the performance of larger models, enabling weak-to-strong verification. 
We consider an 8B setting---using LlaMA 3.1 8B to generate responses along with 8B verifiers---and compare this to a 70B setting (LlaMA 3.3 70B Instruct, 8B-72B verifiers) as well as o3-mini. We see that \weaver{} applied at the 8B scale comes within $1.6\%$ of the majority vote baseline at the 70B scale, and \weaver{} at 70B surpasses o3-mini by $1.0\%$, demonstrating a weak-to-strong verification phenomenon.
Verifier calibration details are available in Appendix \ref{app:scaling_verifier_count}.

\input{tables_and_figures/model_size_ablation}

\noindent \paragraph{(3) Scaling Verifier Count:} 
Two axes for scaling verification are \textbf{(1)} the number of verifiers used and \textbf{(2)} the number of scores sampled from each verifier.
\cref{fig:Weaver_Verifier_Scaling} shows how performance changes as we ensemble 1 to 15 verifiers using both naive averaging and \weaver{}. Verifiers are greedily added in order of individual accuracy, from highest to lowest.
Aggregating more verifiers improves performance by up to $8.5\%$ over the top-1 verifier. 
As shown in \cref{fig:Weaver_Verifier_Scaling}, \weaver{} consistently outperforms naive ensemble averaging across both \textit{Oracle Top-5 Verifiers} and \textit{Total Verifiers} configurations for verifier ensembling, with improvements ranging from +2.4\% to +10.1\% across all datasets. 
The performance gains are particularly pronounced on GPQA Diamond (+10.1\%) and MMLU Pro (+5.1\%), demonstrating \weaver{}'s effectiveness in aggregating verifier signals through learned weights rather than simple averaging.
However, gains diminish as more models are added—reflecting the classic ensemble bias-variance tradeoff: initial improvements stem from variance reduction, while additional verifiers contribute redundant signal due to correlated biases on hard examples~\cite{abebuchanan2024}.
We compare alternative score calibration strategies beyond \weaver{}'s binary transformation in Appendix~\ref{app:individual-verifier-optimization}, and find that the default binarization yields the strongest downstream selection performance.
We also explore scaling the number of scores per verifier—via prompt tuning or temperature variation—in Appendix~\ref{app:scaling_verifier_count}. While this yields modest improvements, increasing verifier count remains the more effective strategy. That said, both methods are complementary and can be combined for further gains.

\input{tables_and_figures/Weaver_Verifier_Scaling}
\input{tables_and_figures/Weaver_Scaling_Laws}

\noindent \paragraph{(4) Scaling Test-Time Compute:} We study how performance scales in the total compute used for both verification and repeated generations.  
Figure \ref{fig:Weaver_Scaling_Laws} shows the relationship between inference-time compute and success rate for different generation-verification systems. For each method, we scale the number of generations exponentially from 1 to 100 and plot the required inference compute for generation and verification together versus the success rate. Note that \cref{fig:Weaver_Scaling_Laws} differs from~\cref{fig:fp_and_pass@1}, since Majority Voting requires $0$ verification inference calls while \weaver{} requires 30+ calls for the weak verifiers.
We find that \weaver{} achieves the highest maximum success rate;
notably, majority voting plateaus at around $2^2$ to $2^3$ ExaFLOPs per query while \weaver{} continues scaling until 512 ExaFLOPs.  However, the additional compute required for \weaver{} can be prohibitive. 
We explore how to reduce this computational burden while retaining \weaver{}'s performance in the next section.

\section{\weaver{} Distillation: Improving Verification Efficiency at Inference}
\label{sec:weaver_distillation}

We explore distillation strategies for fine-tuning a smaller LM as a task-specific verifier. 
In particular, we train \textit{cross-encoders}; the input is a concatenated query-response pair, while the output is \weaver{}'s pseudolabel generated from Weak Supervision, namely $\Pr(y_{ij} = 1 | s_{ij1}, \dots, s_{ijm})$ (see Section \ref{sec:methods}).
For the model, we selected ModernBERT-Large (396M) \citep{modernbert}.
For more details, please see Appendix \ref{app:weaver_distillation}.

\input{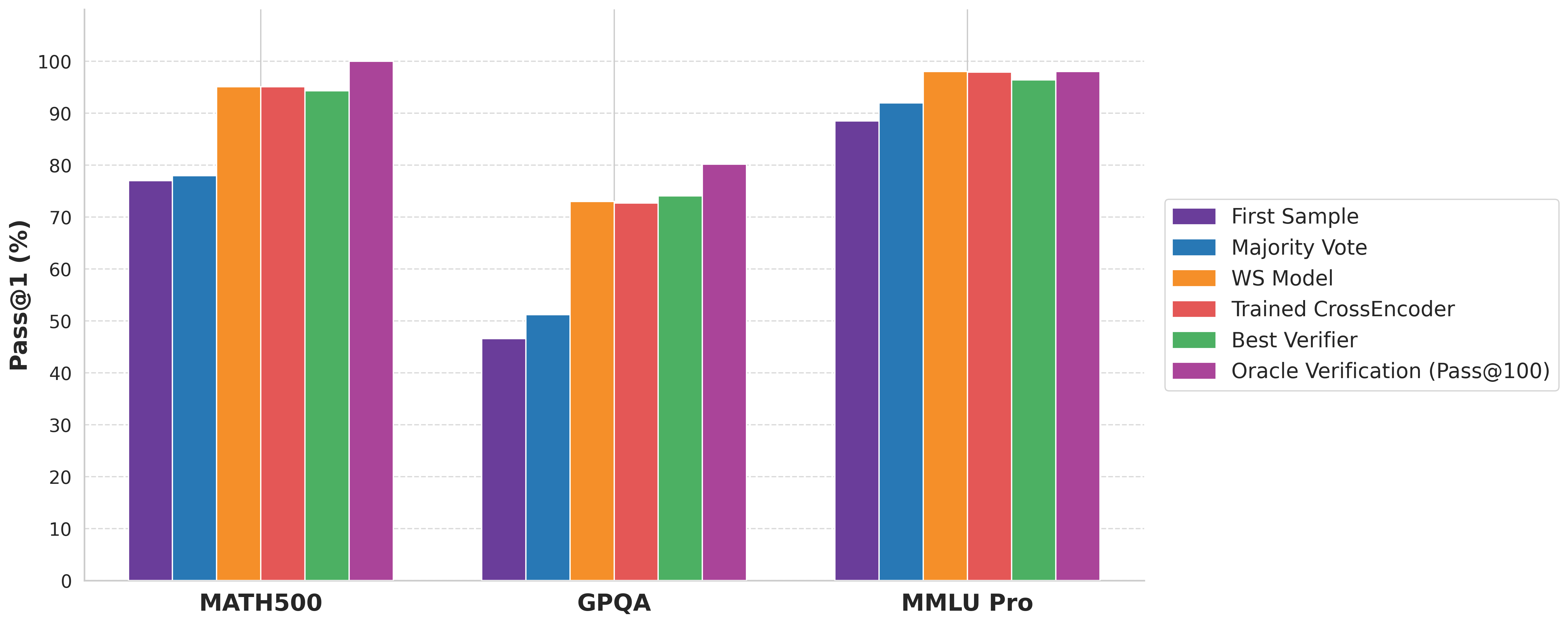}

Figure \ref{fig:CrossEncoderPerformances} shows the performance of \weaver{} on the Llama-70B generations against the cross-encoder on GPQA Diamond.
Across tasks, we find that the distilled cross-encoder is able to capture 98.2\% of the performance of \weaver{}.
When running \weaver{} with all the verifiers, it costs 35.35 exaFLOPs for each query's set of 100 samples.
Running a 400M cross-encoder costs 1.01 exaFLOPs for evaluating 100 samples and \textit{reduces compute cost by more than three orders of magnitude, saving 99.97\% of the FLOPs originally required for running the 70B verifiers}.
We also outperform majority voting by 23.2\% while only incurring a 0.57\% increased inference cost over only generating the responses.
We see similar results for additional datasets in Figure \ref{fig:Verifiers_vs_Generations_Tradeoff} (\cref{app:weaver_distillation}).

These results suggest that, through distillation, we can capture the combined strengths of the weak verifiers used for \weaver{}, and deploy generalizable and lightweight cross-encoders that use only a fraction of the parameters used for generation. 
This reduces our hardware constraints considerably; \textit{rather than utilizing an 8-GPU node per 70B verifier (i.e. Nvidia H200s with 80B memory), we only require a single A100 GPU with 32GB of memory for our cross-encoder}.

%% file: tables_and_figures/verifier_ablations.tex
\begin{table}[h]
\centering
\caption{\textbf{\weaver{} Outperforms Baseline Verification Methods and Shrinks Gap with Frontier LMs.}%
}
\resizebox{\textwidth}{!}{%
\setlength{\tabcolsep}{5pt}{
\begin{tabular}{ccccccccc}
\toprule
 & \multirow{2}{*}{\bf \begin{tabular}[c]{@{}c@{}}  \\ \\Methodology\end{tabular} } &
\multirow{2}{*}{\bf \begin{tabular}[c]{@{}c@{}}  \\\\Generations ($K$) \end{tabular} } & 
\multicolumn{4}{c}{\bf Datasets} &

\multirow{2}{*}{\bf \begin{tabular}[c]{@{}c@{}}  \\\\Average\end{tabular}} \\
\cmidrule(lr){4-7}
& & & \begin{tabular}[c]{@{}c@{}}  MATH\\500\end{tabular} &  \begin{tabular}[c]{@{}c@{}}  GPQA\\Diamond\end{tabular} & \begin{tabular}[c]{@{}c@{}}  MMLU\\College\end{tabular} & \begin{tabular}[c]{@{}c@{}}  MMLU\\Pro\end{tabular} \\
\midrule
\multirow{6}{*}{\rotatebox[origin=c]{90}{\scriptsize \begin{tabular}[c]{@{}c@{}} \textbf{Baselines}\end{tabular}}} & First Sample & 1 &    78.0\% &	42.9\% &	82.6\% &	69.9\% & 68.4\%  \\
& Majority Voting & 100 & 83.0\% &	47.4\% &	84.1\% &	74.4\% & \underline{72.2\%}	    \\ 
& Highest Scoring RM on RewardBench \citep{INF-ORM-Llama3.1-70B,lambert2024rewardbenchevaluatingrewardmodels} & 100 &	78.2\%	&	49.7\%	&		86.0\%	&		77.0\%	& \textbf{72.7\%}	     \\
& Naive Ensemble of Top-10 RMs on RewardBench \citep{lambert2024rewardbenchevaluatingrewardmodels} & 100 & 75.4\% & 41.3\% & 88.1\% & 71.4\% & 69.1\%   \\
& Self-Verification \citep{zhao2025sample} & 100 &  78.1\% &	43.1\% &	82.0\% &	69.5\% &	{66.9\%}
\\ 
& Multi-Agent Verification \citep{lifshitz2025multiagentverificationscalingtesttime} & 100 & 81.3\% &	47.8\% &	84.1\% &	72.6\% &	{71.6\%} \\
\midrule
 & \begin{tabular}[c]{@{}c@{}}  
 \weaver{} 
 \end{tabular} 
 & 100 &  93.4\% & 72.1\% & 94.9\% & 90.2\% & \textbf{87.7\%}   \\
 \midrule
\multirow{5}{*}{\rotatebox[origin=c]{90}{\scriptsize \begin{tabular}[c]{@{}c@{}} \textbf{Frontier}\\\textbf{Approaches}\end{tabular}}}
& GPT-4o \citep{openaigpt4} & 1	& 77.4\% & 35.9\% & 87.1\% & 75.4\% & 69.0\% \\
& Claude 3.7 Sonnet \citep{Anthropic2025Claude37} & 1 & 69.2\% & 48.0\% & 86.1\% & 78.1\% & {70.4\%} \\
& Llama 4 Maverick \citep{meta2025llama4} & 1 & 87.6\% & 68.9\% & 91.1\% & 81.0\% & {82.2\%} \\
& o3-mini \citep{OpenAIo3mini2025} & 1	& 94.4\% & 74.0\% & 92.2\% & 86.0\% & \underline{86.7\%} \\
& Oracle Verification (Pass@100) & 100	&	98.6\% & 81.0\% & 96.0\% & 92.0\% & \textbf{91.9\%} \\
\bottomrule
\end{tabular}
}
} 
\label{tab:verifier_ablations}
\end{table}

%% file: tables_and_figures/FP_and_Pass1.tex
\begin{figure}[h]
\centering
\includegraphics[width=\linewidth]{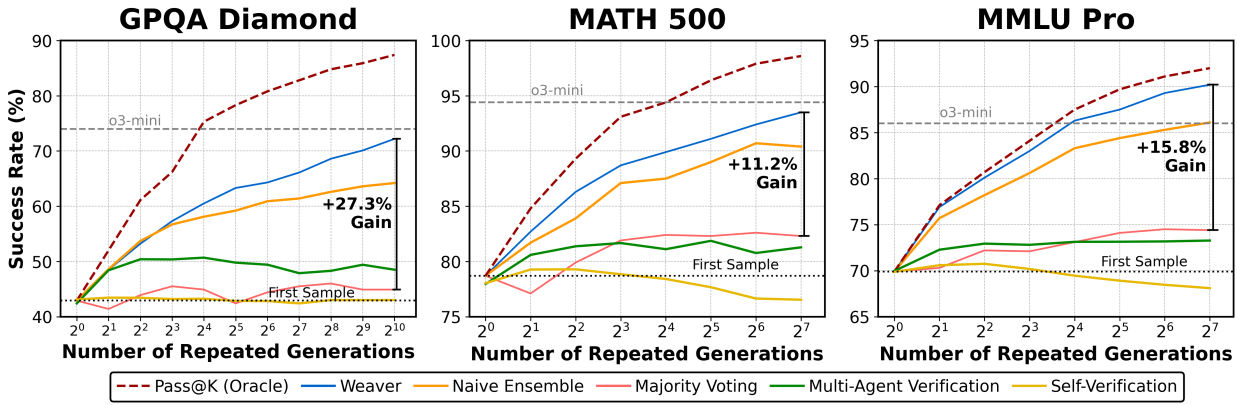}
\caption{\textbf{Scaling Generations Boosts Performance with \weaver{}}: 
   The generation-verification gap shrinks when increasing $K$ and leveraging \weaver{}, outperforming alternative verification methods by an average 18.3\%.
  }
\label{fig:fp_and_pass@1}
\end{figure}

%% file: tables_and_figures/model_size_ablation.tex
\begin{table*}[h]
  \centering
  \small
  \caption{
  \textbf{\weaver{} Reduces Gap between Model Classes: 8B and 70B, 70B and Frontier LM} %
  }  %
  \resizebox{\textwidth}{!}{%
  \setlength{\tabcolsep}{5pt}{
  \begin{tabular}{cccccccc}
    \toprule
    \multirow{3}{*}{\bf  \begin{tabular}[c]{@{}c@{}}  Generator\\Model\end{tabular}} &
    \multirow{3}{*}{\bf  \begin{tabular}[c]{@{}c@{}}  Verifier\\Model\end{tabular}} &
    \multirow{3}{*}{\bf \begin{tabular}[c]{@{}c@{}}  Aggregation\\Strategy\end{tabular}} & 
    \multicolumn{4}{c}{\bf Datasets} &  
    \multirow{3}{*}{\bf Average} \\
    \cmidrule(lr){4-7}
    & & & MATH & \begin{tabular}[c]{@{}c@{}}  GPQA\\Diamond\end{tabular} & \begin{tabular}[c]{@{}c@{}}  MMLU\\College\end{tabular} & \begin{tabular}[c]{@{}c@{}}  MMLU\\Pro\end{tabular} \\
    \midrule
  \multirow{2}{*}{Llama 3.1 8B Instruct} & N/A & Majority Vote & 69.0\% &		30.5\% &		72.7\% &	56.4\% & 57.2\%  \\    
 & 8B and below & \weaver{}	&	80.0\%	&	47.1\%	&	85.7\%	&	67.2\%	& \textbf{70.0\%}  \\
    \cmidrule(lr){3-8} 
    \multicolumn{3}{r}{\bf $\Delta$ w. \weaver{}} & +11.0\% & +16.6\% & +13.0\% & +10.2\% & +12.8\% \\
    \midrule
  \multirow{2}{*}{Llama 3.3 70B Instruct} & N/A & Majority Vote & 83.0\% &	44.9\% &		84.1\% &		74.4\% & 71.6\%   \\    
    & 72B and below & \weaver{} & 93.4\% & 72.2\% & 94.9\% & 90.2\% & \textbf{87.6\%} \\
    \cmidrule(lr){3-8} 
    \multicolumn{3}{r}{\bf $\Delta$ w. \weaver{}} & +10.4\% & +27.3\% & +10.8\% & +15.8\% & +16.0\% \\
    \midrule 
    o3-mini & N/A & First Sample & 94.4\% & 74.0\% & 92.2\% & 86.0\% & \textbf{86.7\%} \\
    \bottomrule
  \end{tabular}
  }}
\label{tab:model_size_ablation}
\end{table*}
\vspace{-0.5cm}

%% file: tables_and_figures/Weaver_Verifier_Scaling.tex
\begin{figure}[h]
   \centering
   \includegraphics[width=\linewidth]{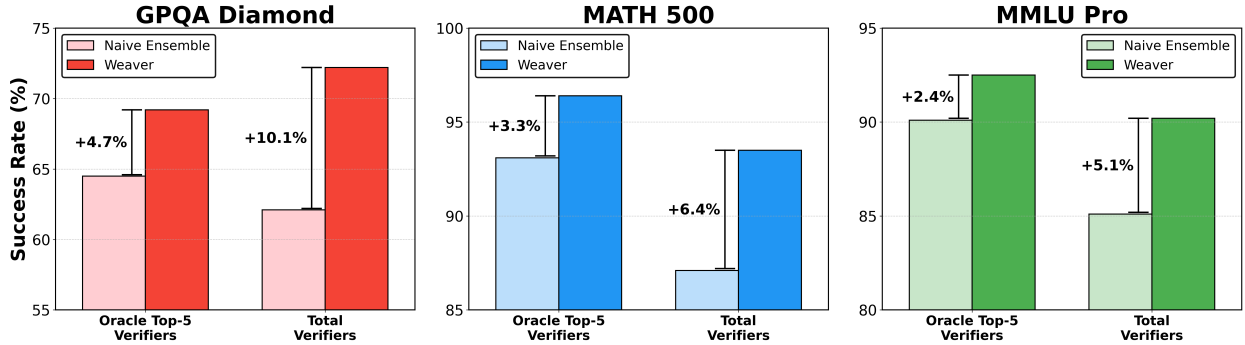}
   \caption{\textbf{\weaver{} Outperforms Naive Ensemble across Oracle Top-5 Verifiers and Total Verifiers Configurations:}
   Results are shown for \weaver{} ensembles and naive ensembles of the \textit{Oracle Top-5 Verifiers} (highest-performing verifiers on dataset selected using ground truth) and \textit{Total Verifiers} (all available verifiers). 
   \weaver{} consistently outperforms naive ensemble averaging, with improvements ranging from +2.4\% to +10.1\%.
   }
   \label{fig:Weaver_Verifier_Scaling}
\end{figure}
\vspace{-0.5cm}

%% file: tables_and_figures/Weaver_Scaling_Laws.tex
\begin{figure}[h]
   \centering
   \includegraphics[width=\linewidth]{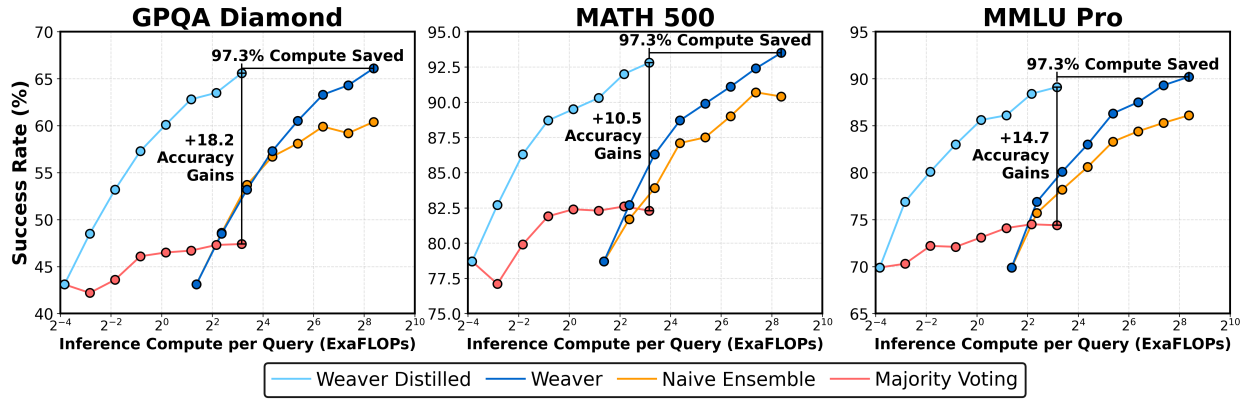}
\caption{\textbf{\weaver{} Improves the Accuracy-Compute Performance Trade-Offs.}
Success rate ($\%$) as a function of total inference compute per query (generation and verification compute, log scaled) for different verification strategies.
Each point represents a different number of candidate generations (from $2^0$ to $2^{7}$).
\weaver{} achieves the highest accuracy while requiring more compute than Majority Voting but demonstrates continued scaling benefits, while \weaver{} Distilled maintains most of \weaver{}'s performance gains with 97.3\% compute savings and substantial accuracy improvements over baseline methods.
} 
   \label{fig:Weaver_Scaling_Laws}
\end{figure}

%% file: tables_and_figures/CrossEncoderPerformances.tex
\begin{wrapfigure}{r}{0.45\textwidth}
   \centering \includegraphics[width=\linewidth]{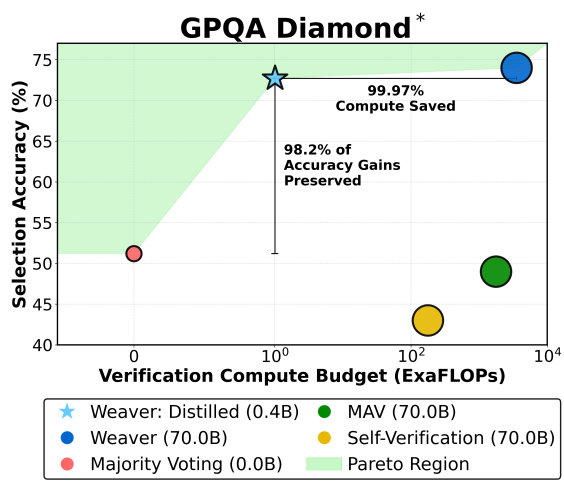}
   \caption{
   \textbf{Distilling \weaver{} into a 400M Cross-Encoder Almost Entirely Captures the Performance of \weaver{}, Yielding  99.97\% Compute Savings.} {$^*$}We train/evaluate on an 80:20 split. 
   }  \label{fig:CrossEncoderPerformances}
\end{wrapfigure}

%% file: sections/discussion.tex
\section{Discussion}
\label{sec:discussion}

Several research directions remain to be explored with \weaver{}:

\begin{enumerate}

\item \noindent \textbf{Specialized Verifier Development}: Our work highlights the varying effectiveness of different weak verifier categories across task domains. Future research should investigate specialized verifier architectures tailored to specific tasks, such as enhanced mathematical reasoning capabilities for numerical problems \citep{yang2024qwen25mathtechnicalreportmathematical} or improved code execution simulation for programming tasks \citep{jain2024livecodebenchholisticcontaminationfree, quan2025codeelobenchmarkingcompetitionlevelcode}.

\item \noindent \textbf{Dataset Distribution}: For particularly difficult datasets, such as AIMO 2024, \weaver{} has trouble selecting a correct answer since there are so few correct responses compared to the other datasets (\autoref{tab:8B_positive_negative_ratios}; \autoref{tab:70B_positive_negative_ratios}). 
By scaling the number of generated responses, we are able to improve performance by increasing the absolute number of correct responses (Figure \ref{fig:fp_and_pass@1}) but improved generation techniques, verifier scoring, and aggregation techniques can help us better close the gap with $Pass@K$ for these harder tasks.

\item \noindent \textbf{\weaver{} for RLHF}: With \weaver{} generated predictions, we can improve the quality of labels used for RLHF on reasoning and mathematics, improving beyond an individual verifier. 
Previous work has explored reward model ensemble approaches towards RLHF, minimizing poor performing RMs while maximizing the complementary strengths of accurate RMs \cite{wang2024transforming, eisenstein2023helping}.
Fine-tuning generation models can further improve the accuracy-compute trade-off of \weaver{} beyond solely distilling a lightweight verifier (Section \ref{sec:weaver_distillation}), improving the positive/negative generation ratio and thus making verification an easier task for \weaver{} models. 

\item \noindent \textbf{Multi-Modal Verification}: Extending \weaver{} to multimodal tasks involving images, audio, or video would broaden its applicability but introduces new challenges in verification across modalities \citep{phan2025humanitysexam, wang2020vatexlargescalehighqualitymultilingual}. Research is needed on how verification signals can be effectively combined across different data types.
\end{enumerate}

\section{Conclusion}
\label{sec:conclusion}

In this paper, we present \weaver{}, a framework that addresses the fundamental challenge of scaling test-time compute through effective verification strategies. 
Our contributions advance the state of knowledge in three key dimensions. 
First, we establish that weighted aggregation of weak verifiers substantially outperforms both individual verifiers and majority voting across reasoning and mathematics tasks, with weighted aggregation exceeding majority voting by an average of 12.3\% across all tasks explored (Table~\ref{tab:verifier_ablations}; Figure~\ref{fig:MV_vs_Naive_vs_Weighted}). 
Second, we developed a principled approach for unsupervised estimation of verifier accuracies using weak supervision, enabling effective ensemble weighting without fine-tuning on costly ground-truth annotations. This allows us to close the generation-verification gap by 12.8\% for 8B models and 16.0\% for 70B models (Table~\ref{tab:model_size_ablation}). By leveraging \weaver{} with 70B models, we marginally outperform frontier closed-source models such as OpenAI's o3-mini (87.7\% vs. 86.7\%) on average across the tasks explored (Table~\ref{tab:verifier_ablations}).
Third, we improve the accuracy-compute trade-off by distilling \weaver{} into lightweight 400M-parameter cross-encoders. These distilled models retain 98.2\% of \weaver{}'s performance while reducing inference compute by 99.97\% (Section~\ref{sec:weaver_distillation}). This enables high-throughput, cost-efficient verification without sacrificing accuracy, and demonstrates that scalable verification can be achieved without repeatedly querying large models.
These findings suggest that strategically combining and distilling weak verifiers enables scalable, label-efficient, and compute-efficient verification—paving the way for better data filtering, model alignment, and inference-time decision-making without additional training of the base generator model.

%% file: sections/acknowledgements.tex
\section{Acknowledgements}

We thank the members of the Hazy Lab, Linderman Lab and Scaling Intelligence Lab for their constructive feedback during the composition of the paper. 
In particular, we would like to thank 
Daniel Biderman, Bradley Brown, Ryan Ehrlich, Sabri Eyuboglu, Anna Goldie, Neel Guha, Simon Guo, Jordan Juravsky, Hermann Kumbong, Jerry Liu, Avanika Narayan, Anne Ouyang, Benjamin Spector, Shayan Talaei, Benjamin Viggiano, and Michael Zhang. 
We also thank Marlowe and Together AI for providing compute resources that enabled our experiments. 

We gratefully acknowledge the support of NIH under No. U54EB020405 (Mobilize); NSF under Nos. CCF2247015 (Hardware-Aware), CCF1763315 (Beyond Sparsity), CCF1563078 (Volume to Velocity), and 1937301 (RTML); US DEVCOM ARL under Nos. W911NF-23-2-0184 (Long-context) and W911NF-21-2-0251 (Interactive Human-AI Teaming); ONR under No. N000142312633 (Deep Signal Processing); Stanford HAI under No. 247183; Google DeepMind; Google Research; Google Cloud; NXP; Xilinx; LETI-CEA; Intel; IBM; Microsoft; NEC; Toshiba; TSMC; ARM; Hitachi; BASF; Accenture; Ericsson; Qualcomm; Analog Devices; Salesforce; Total; the HAI-GCP Cloud Credits for Research program; the Stanford Data Science Initiative (SDSI); members of the Stanford DAWN project: Meta, Google, and VMWare; and members of the Stanford SEAMS project: IBM and Felicis.
The U.S. Government is authorized to reproduce and distribute reprints for Governmental
purposes notwithstanding any copyright notation thereon. Any opinions, findings, and conclusions
or recommendations expressed in this material are those of the authors and do not necessarily
reflect the views, policies, or endorsements, either expressed or implied, of NIH, ONR, or the U.S.
Government.

%% file: sections/appendix.tex
\newpage
\appendix
\onecolumn

\section{Table of Contents}
\begin{enumerate}
    \item \textbf{\weaver{} Methodology} (Appendix \ref{app:weaver_methodology})
    \begin{enumerate}
        \item Discrete Weak Supervision Model with Known Difficulty (Appendix \ref{app:ws_discrete_knowndiff})
        \item Adapting Weak Supervision to the Verification Setting (Appendix \ref{app:practical_considerations})
        \item Filtering Out Low-Quality Verifiers (Appendix \ref{app:filtering_out_verifiers})
        \item Adaptation Method (Appendix \ref{app:implementation_decisions})
        \item Clustering by Difficulty to Improve \weaver{} (Appendix \ref{app:clustering_exploration})
    \end{enumerate}
    \item \textbf{Experiments}: (Appendix \ref{app:experiments})
    \begin{enumerate}
        \item Models and Datasets (Appendix \ref{app:models_and_datasets})
        \item Verification Baselines (Appendix \ref{app:verification_baselines})
        \item Scaling Trends of \weaver{} (Appendix \ref{app:scaling_trends_of_weaver})
        \item Scaling Candidate Generations (Appendix \ref{app:scaling_candidate_generations})
        \item Scaling Verifier Count (Appendix \ref{app:scaling_verifier_count})
        \item Weaver Distillation (Appendix \ref{app:weaver_distillation})
        \item Individual Verifier Optimization (Appendix \ref{app:individual-verifier-optimization})
    \end{enumerate}
    \item \textbf{Miscellaneous} (Appendix \ref{app:miscellaneous})
    \begin{enumerate}
        \item Compute Requirements (Appendix \ref{app:compute_requirements})
    \end{enumerate}
\end{enumerate}

\newpage
\section{\weaver{} Methodology}
\label{app:weaver_methodology}

\subsection{Weak Supervision Model}
\label{app:ws_discrete_knowndiff}

We can construct a data generating model over response correctness $y$ and the binary verifier outputs $\bar{s}$. The model is defined as:
\begin{align*}
\text{Response Correctness:} \quad & y_{ij}  \sim \text{Bernoulli}(\pi) \; \forall i \in [n], j \in [K], \\
\text{Verifier Score:} \quad & \bar{s}_{ijk} \mid y_{ij} \sim
\begin{cases}
\text{Bernoulli}(w_{k,1}), & \text{if } y_{ij} = 1, \\
\text{Bernoulli}(1 - w_{k,0}), & \text{if } y_{ij} = 0,
\end{cases} \; \forall i \in [n], j \in [K], k \in [m]
\end{align*}

where:
\begin{itemize}
\item \(\pi\) is the probability that a response is correct.
\item \(w_{k,1}\) is the true positive rate (TPR) of verifier \(k\), and \(w_{k,0}\) is the true negative rate (TNR), which we refer to as the verifier's \textit{accuracy parameters.}
\end{itemize}

Here, each verifier $k$ emits a binary score $\bar{s}_{ijk} \in \{0, 1\}$,  which is assumed to be a noisy indicator of whether response $y_{ij}$ is correct. The likelihood of the verifier's binary output \(X_{ijk} \in \{0,1\}\) is:
\[
\Pr(S_k = \bar{s}_{ijk} \mid Y = y_{ij}) = 
\begin{cases}
w_{k,1}, & \text{if } y_{ij} = 1 \text{ and } \bar{s}_{ijk} = 1, \\
1 - w_{k,1}, & \text{if } y_{ij} = 1 \text{ and } \bar{s}_{ijk} = 0, \\
w_{k,0}, & \text{if } y_{ij} = 0 \text{ and } \bar{s}_{ijk} = 0, \\
1 - w_{k,0}, & \text{if } y_{ij} = 0 \text{ and } \bar{s}_{ijk} = 1.
\end{cases}
\]

We are interested in estimating the correctness of a response \( y_{ij} \in \{0, 1\} \) based on assessments from multiple verifiers \(\bar{s}_{ij} = \{\bar{s}_{ij1}, \bar{s}_{ij2}, \dots, \bar{s}_{ijm}\} \). Applying Bayes' Rule, we get:
\begin{align}
\Pr(y_{ij} = 1 \mid S = \bar{s}_{ij}) = \frac{\Pr(S = \bar{s}_{ij} \mid y_{ij} = 1) \Pr(y_{ij} = 1)}{\Pr(\bar{s}_{ij})}
\label{eqn:y1_posterior}
\end{align}

\text{where} $ \Pr(\bar{s}_{ij}) = \sum_{y' \in \{0,1\}} \Pr(\bar{s}_{ij} \mid y_{ij} = y') \Pr(y_{ij} = y')$.\par

\cref{eqn:y1_posterior} requires evaluating the full conditional likelihood:
\[
\Pr(S = \bar{s}_{ij} \mid y_{ij}) = \Pr(S_1 = \bar{s}_{ij1}, S_2 = \bar{s}_{ij2}, \dots, S_m = \bar{s}_{ijm} \mid y_{ij}),
\]
which is a joint distribution over \(m\) binary random variables. Since each verifier \(S_{k} \in \{0, 1\}\) is binary, then there are $2^m$ possible verifier output configurations for \(S \in \{0, 1\}^m\).
This results in \(2^{m} - 1\) free parameters per class label to construct the distribution $\Pr(S = \bar{s}_{ij} \mid y_{ij})$. %

\paragraph{Conditional Independence Assumption}
To avoid this exponential blowup, we can assume that the verifiers provide \textit{conditionally independent} outputs:
\[
P(S \mid y) = \prod_{k=1}^{m} P(S_k \mid y),
\]
which reduces the number of parameters from \(O(2^m)\) to \(O(m)\) and enables efficient inference, under the assumption that each verifier provides unique information about the correctness of a response.

Then, \cref{eqn:y1_posterior} simplifies to a Naive Bayes-style estimator:
\begin{align}
  \Pr\bigl(y_{ij}=1 \mid S = \bar{s}_{ij} \bigr)
  \;=\; 
  \frac{\Pr(S_1 = \bar{s}_{ij1},\dots, S_m = \bar{s}_{ijm} |y_{ij}=1) \Pr(y_{ij}=1)}{\Pr(S = \bar{s}_{ij})}  \nonumber \\
  \;=\; 
  \frac{
    \Pr(y_{ij}=1)\,\prod_{k=1}^m\,\Pr\bigl(S_k = \bar{s}_{ijk} \mid y_{ij}=1\bigr)
  }{
    \sum_{y'\in\{0,1\}}
      \Pr(y_{ij}=y')\,\prod_{k=1}^m\,\Pr\bigl(S_k = \bar{s}_{ijk} \mid y_{ij}=y'\bigr)
  }
\label{eqn:naive_bayes}
\end{align}

The parameters in \cref{eqn:naive_bayes} include:
\begin{itemize}
\item The prior probability of correctness \( \pi = \Pr(y_{ij} = 1) \).
\item The verifier-specific conditional likelihoods $   P(S_k \mid y_{ij})$. \\
\end{itemize}

\subsubsection{Parameter Estimation}

\paragraph{Supervised Setting}
When ground-truth labels \(y_{ij}\) are available, parameter estimation reduces to computing empirical frequencies. 
We can estimate the prior as:
\[
\hat{\pi} = \frac{1}{N} \sum_{i,j} \mathbf{1}\{y_{ij} = 1\}
\]

For each verifier \(k\), we could estimate:
\begin{align*}
\hat{w}_{k,1} &= \frac{\sum_{i,j} \mathbf{1}\{y_{ij} = 1\} \cdot \mathbf{1}\{S_k = 1\}}{\sum_{i,j} \mathbf{1}\{y_{ij} = 1\}}, \\
\hat{w}_{k,0} &= \frac{\sum_{i,j} \mathbf{1}\{y_{ij} = 0\} \cdot \mathbf{1}\{S_k = 0\}}{\sum_{i,j} \mathbf{1}\{y_{ij} = 0\}}.
\end{align*}

\paragraph{Weak Supervised Setting}
When a few labeled $y_{ij}$ are available, we can use it to estimate $\pi$, but we still need to estimate the verifier accuracy parameters $w_{k,1}, w_{k,0}$ to compute $\prod_{k=1}^m \Pr(S_k = \bar{s}_{ijk}|y_{ij}=1)$. %
Instead of using labeled data, we estimate accuracy parameters using moment matching. In particular, we match observable second moments of verifier outputs to the model-implied moments under conditional independence assumptions, based on an approach from~\cite{HazyResearchMetal}.

\paragraph{Pairwise Statistics.}
For each pair of verifiers \(k_1, k_2\) and binary outputs \(a,b\in\{0,1\}\), we can express the joint probability of their outputs using the marginalization rule and the conditional independence assumption:
\begin{align}
&\Pr(S_{k_1} = a, S_{k_2} = b)\label{eq:indep2} \\
&= \Pr(S_{k_1} = a | Y = 1) \Pr(S_{k_2} = a | Y = 1) \Pr(Y = 1) + \Pr(S_{k_1} = b | Y = 0) \Pr(S_{k_2} = b | Y = 0) \Pr(Y = 0) \nonumber 
\end{align}
where the conditional distributions for verifier $k$ are:
\[
  \Pr(S_k =a \mid y=1)
  \;=\;
  \begin{cases}
    w_{k,1}, & a=1,\\
    1-w_{k,1}, & a=0,
  \end{cases}
  \quad
  \Pr(S_k =a\mid y=0)
  \;=\;
  \begin{cases}
    1-w_{k,0}, & a=1,\\
    w_{k,0}, & a=0.
  \end{cases}
\]

\paragraph{Marginal Statistics.}

Similarly, each verifier's marginal distribution can be written as:
\begin{align}
\Pr(S_k = 1) = \Pr(S_k = 1 | Y = 1)\Pr(Y = 1) + \Pr(S_k = 1 | Y = 0) \Pr(Y = 0) \label{eq:marginal}
\end{align}

Note that this equation holds true regardless of the conditional independence assumption.

\paragraph{Estimation method}

\begin{itemize}
\item Construct the second order moment matrix  \(O \in \mathbb{R}^{(2m)\times(2m)}\), where:
\begin{align}
     O_{2i-1:2i, 2i-1:2i} &= \SmallMatrix{
        \Pr(S_i = 0) & 0 \\ 0 & \Pr(S_i = 1)
    } \; \forall i \in [m] \nonumber \\
    O_{2i-1:2i, 2j-1: 2j} &= \SmallMatrix{
        \Pr(S_i = 0, S_j = 0) & \Pr(S_i = 0, S_j = 1) \\ \Pr(S_i = 1, S_j = 0) & \Pr(S_i = 1, S_j = 1)
    } \; \forall i \neq j \in [m]
\end{align}

\item Construct the conditional probability matrix \(\mu \in \mathbb{R}^{(2V)\times 2}\), where each row encodes: 
\[
  \mu_{2k+a,\,b}
  \;=\;
  \Pr\bigl(S_{k}=a \mid y=b\bigr)
\] 
\[
\mu =  \begin{bmatrix} w_{k_1, 0} & 1 - w_{k_1, 1}\\ 1 - w_{k_1, 0} & w_{k_1, 1}\\ \dots & \dots \\ w_{k_m, 0} & 1 - w_{k_m,1} \\ 1 - w_{k_m, 0} &  w_{k_m,1}\end{bmatrix}
\]
\item Label prior matrix \(P\in\mathbb{R}^{2\times 2}\) is a diagonal matrix:
\[
  P
  \;=\;
  \begin{bmatrix}
    \Pr(y_{ij}=0) & 0\\
    0 & \Pr(y_{ij}=1)
  \end{bmatrix}.
\]
\end{itemize}

Then, equation~\ref{eq:indep2} is equivalent to $O = \mu P \mu^\top$ on the entries off of the $2 \times 2$ block diagonal, and equation~\ref{eq:marginal} is equivalent to $\text{diag}(0) = \mu P \mathbb{1}^\top$. Therefore, we optimize the following loss to compute $\mu$.

\[
  \text{minimize}_{\mu} \bigl\|\,O_{\text{off-diag}} - (\mu\,P\,\mu^T)_{\text{off-diag}}\bigr\|^2 + 
  \bigl\|\,\mathrm{diag}(O) - \mu\,P\,\mathbf{1}^T\bigr\|^2,
\]

By solving via gradient-descent, we obtain estimates of the verifier accuracy parameters \(\{w_{k,1}, w_{k,0}\}\).

\subsubsection{Inference: computing response correctness probabilities}
Once the accuracy parameters $\{w_{k,1}, w_{k, 0} \}$ are estimated and $P(y_{ij}=1)$ is computed from a small labeled development dataset, we can compute posterior correctness probabilities for each response:
\begin{align*}
\Pr(y_{ij}= 1 | S = \bar{s}_{ij} ) \propto \Pr(y_{ij}=1) \prod_{k=1}^m \Pr(S_k = \bar{s}_{ijk}| y_{ij}=1) \\
\Pr(y_{ij}= 0 | S = \bar{s}_{ij}) \propto \Pr(y_{ij}=0) \prod_{k=1}^m \Pr(S_k = \bar{s}_{ijk} | y_{ij}=0)
\end{align*}

Normalizing these, we have a full posterior $P(y_{ij}=1| S = \bar{s}_{ij})$, which provides a score with which we can select a response for each query.

\subsection{Adapting Weak Supervision to the Verification Setting}
\label{app:practical_considerations}

Given a set of verifiers, we elaborate on the design choices behind the weak supervision model described in \cref{sec:methods}.  In particular, we describe challenges around normalization, binarization  and filtering out low-quality verifiers. In Section~\ref{app:implementation_decisions}, we describe \weaver{}'s approach to normalizing, binarizing, and filtering verifiers. 
 
\subsubsection{Normalization}
Verifier outputs often differ substantially in scale, range, and distribution. For instance, some verifiers output unbounded real-valued scores (e.g., log-likelihoods), while others output normalized probabilities or learned regression values.  Some standard losses under this framework include ranking losses, binary classification losses and regression losses. Examples include:

\begin{itemize}
\item \textbf{Bradley--Terry (ranking) loss:} \( \mathcal{L}(s_1, s_0) = \log(1 + \exp(s_0 - s_1)) \)
\item \textbf{Logistic (binary classification) loss:} \( \mathcal{L}(s, y) = \log(1 + \exp(-y' s)), \quad y' = 2y - 1 \)
    \item \textbf{Squared (regression) loss:} \( \mathcal{L}(s, y) = (s - y)^2 \)
\end{itemize}

To combine out of the box verifiers, which may be trained under different constraints, verifier scores must be comparable.
We note that standard losses imposes different but related invariance assumptions:
\begin{itemize}
\item \textbf{Ranking and binary classification losses}: These losses focus on the relative order of the scores, and therefore, the relative ranking of the scores is preserved under positive affine transformations \( s \mapsto \alpha s + \beta \), where \( \alpha > 0 \). %

\item \textbf{Regression losses}: These losses directly penalize the difference between predicted scores and target values, so the absolute scale of the scores is important. Because many verifier outputs approximate correctness labels \( y \in [0, 1] \), it is essential to constrain scores to the same interval to ensure meaningful comparisons. 
\end{itemize}

\subsubsection{Binarization}

The weak supervision algorithm described in the prior section requires binary verifier outputs. This is naturally suited for judge-style verifiers—such as language models prompted to answer yes/no questions—which output discrete ${0, 1}$ labels. However, many verifiers, especially reward models, emit continuous scores and often vary in scale and calibration. This raises a key design question: should we input these scores to the weak supervision model as-is, or should we binarize them?

Using continuous scores retains fine-grained information about the confidence of each verifier. This can improve ranking-based performance metrics such as AUC and may allow the weak supervision model to better resolve disagreements among verifiers. However, it introduces challenges when combining signals across verifiers with inconsistent calibration or scale: a score of 0.8 may have different meanings for different verifiers. As seen in Figure \cref{fig:histogram_verifier_accuracy_gpqa} different verifiers exhibit different score distributions even when evaluating the same set of responses. Some verifiers are sharply bimodal, others skew heavily toward low or high scores, and some produce nearly flat or noisy distributions.

To address this, we evaluate several binarization strategies that convert continuous verifier scores into discrete labels. Figure~\ref{fig:lr_binarization_auc} compares the AUC performance of a logistic regression model trained on verifier outputs across four binarization methods: no binarization (continuous scores), a fixed threshold at 0.5, class balance-based thresholds, and quantile binarization. We observe that while continuous scores can achieve strong AUC when sufficient training data is available, simple binarization strategies—especially those that account for score distribution skew—perform comparably and are more robust under limited supervision.

Figure~\ref{fig:lr_binarization_select_acc} shows only modest differences across binarization strategies for selection accuracy. We note in highly imbalanced datasets, as in the case GPQA, simple quantile-based binarization performs particularly well, likely because it adjusts for the skewed distribution of scores, i.e. it discards ambiguous mid-range scores and retains only the most confident signals.

\input{tables_and_figures/binarization}.

\subsubsection{Filtering out Low-Quality Verifiers}
\label{app:filtering_out_verifiers}

Verifiers with low accuracy or extreme marginals (e.g., near-constant outputs) not only degrade ensemble performance but also undermine the stability and identifiability of Weak Supervision algorithms, worsening the estimation error. \textit{How do we discard verifiers that have low signal?}
\begin{itemize}[leftmargin=*,itemsep=0pt,parsep=0pt,topsep=0pt,partopsep=0pt]
\item Skewed marginals: Consider a dataset where $\Pr(y = 1) \approx 0.5$ and we have a verifier with $\Pr(S_k = 1) \approx 0.99$. A skewed verifier with an extreme marginal (e.g., from naive thresholding for binarization) and near-constant outputs adds little information to the ensemble. It primarily increases noise in the objective in Eq.~\ref{eqn:ws_loss} and should thus be discarded. Yet, not at all verifiers that have extreme marginals add little signal; for instance, if instead $\Pr(y = 1) \approx 0.99$, a skewed verifier could be highly accurate. Therefore, the definition of a low-quality verifier depends on the distribution of correct responses.
\item Breaking symmetry in the WS objective: a common assumption of Weak Supervision is that a majority of the verifiers have better-than-random accuracy~\cite{fu2020fast}. Otherwise, there is a possibility that the WS algorithm can yield non-unique solutions; the terms in Eq.~\ref{eqn:ws_loss} are the joint probabilities over pairs of verifiers as well as their marginals, which do not uniquely determine if a verifier satisfies $w_{k, 1}, w_{k, 0} > 0.5$ or not. Therefore, it is critical to remove as many low-accuracy verifiers as possible to ensure that the estimation procedure converges to a unique solution.
\end{itemize}

\input{sections/implementation_details}

\section{Experiments}
\label{app:experiments}

\subsection{Models and Datasets}
\label{app:models_and_datasets}

\noindent \textbf{Benchmarks}: We evaluate our models with several benchmarks for instruction-following, reasoning, mathematics, and coding: MATH500 \citep{hendrycks2021measuring}, GPQA \citep{rein2024gpqa}, MMLU \citep{hendrycks2021measuring}, MMLU Pro \citep{wang2024mmlu}, and BBH \citep{suzgun2022challengingbigbenchtaskschainofthought}.
We provide an overview of each dataset in \autoref{tab:benchmarks_overview}.
For MMLU, we selected the college-level questions for evaluation: biology, chemistry, physics, mathematics, computer science, and medicine.
For MMLU Pro, we take a random sample of 500 queries out of the 12K queries available.
For BBH, we take four tasks from the dataset of 6K queries available: Penguins in a Table, Causal Judgement, Logical Deduction (Five Objects), and Tracking Shuffled Objects (Five Objects).

\noindent \textbf{Models}: We evaluate candidate generations using a range of \textit{weak verifiers}—models with imperfect but better-than-random accuracy. Our verification system \(\mathcal{V}\) includes two primary classes of weak verifiers: \textit{Reward Models} and \textit{LM Judges}.

\begin{itemize}
\item \noindent \textbf{Reward Models}:
A \textit{reward model} (RM) is a trained language model that assigns a scalar score to candidate responses based on how well they align with human preferences \citep{lambert2024rewardbenchevaluatingrewardmodels, song2025prmbenchfinegrainedchallengingbenchmark}. 
Given a query and a candidate response, the RM outputs a value \(V_{ij} \in [0, 1]\) representing the estimated quality of candidate \(j\) according to criteria such as correctness, helpfulness, and safety.
\begin{itemize}
    \item Examples of reward models include those from the RewardBench leaderboard~\citep{lambert2024rewardbenchevaluatingrewardmodels}, such as INF-ORM~\citep{INF-ORM-Llama3.1-70B}, QRM Gemma~\citep{dorka2024quantile}, and Skywork Reward~\citep{liu2024skywork}. We also include \textit{process reward models} (PRMs), which score the reasoning process itself—emphasizing step-by-step logic and coherence—rather than just the final answer~\citep{cui2025process, yuan2024implicitprm}.
    \item For our study, we selected the top-20 reward models from RewardBench and the top-20 process reward models from Process Reward Bench~\citep{song2025prmbenchfinegrainedchallengingbenchmark} at both 8B and 70B parameter scales. We exclude any RM or PRM that fails to provide a positive learning signal—i.e., those whose rankings perform no better than random selection on benchmark train sets (Appendix ~\ref{app:models_and_datasets}). 
    The diverse training objectives and datasets used for these reward models introduce systematic biases that affect their verification capabilities \citep{lambert2024rewardbenchevaluatingrewardmodels, song2025prmbenchfinegrainedchallengingbenchmark}, with different loss functions—including Bradley-Terry loss for pairwise preferences \citep{bradley1952rank}, margin loss for fixed score differences \citep{rosset2003margin}, and pairwise ranking loss for relative ordering \citep{cao2007learning}.
    \item Previous work has noted that it is nontrivial to combine the outputs of reward models and judges as they provide logits and binary decision rules \citep{verga2024replacingjudgesjuriesevaluating, xu2024perfectblendredefiningrlhf}. 
    Instead, we find that we can normalize all RM scores to the range \([0, 1]\) using robust percentiles: the bottom 5th percentile is mapped to 0 and the top 95th percentile to 1. 
    For models that provide multiple scoring dimensions (e.g., ArmoRM~\citep{ArmoRM}), we use only their primary output. 
\end{itemize}

\item \noindent \textbf{LM Judges}:
An \textit{LM judge} is a language model used to assess the correctness of a candidate response by generating a binary verdict: \(V_{ij} \in \{0, 1\}\), where 1 indicates that the response is judged correct. These models typically apply chain-of-thought (CoT) reasoning to arrive at their decisions~\citep{wei2023chainofthoughtpromptingelicitsreasoning}. Each LM judge takes a query and a response as input and outputs a single binary verdict.
\begin{itemize}
    \item We use well-known chat models from ChatBotArena~\citep{chiang2024chatbot} as LM judges, which are known for their general-purpose reasoning capabilities. To ensure consistency and determinism, we use greedy decoding (temperature \(T = 0\)) when generating judgments.
\end{itemize}

\end{itemize}

\input{tables_and_figures/benchmarks_overview}

\input{tables_and_figures/Growth_of_RMs}

\input{tables_and_figures/8B_positive_negative_ratios}
\input{tables_and_figures/70B_positive_negative_ratios}

\input{tables_and_figures/models_overview}

\input{tables_and_figures/verifier_accuracies_ranges}

\subsection{Verification Baselines}
\label{app:verification_baselines}

\subsubsection{Verifier-Free Approaches}

\noindent \textbf{First Sample (Pass@1)}: This baseline uses only the first generated response without any verification or selection mechanism. It represents the standard approach where models generate a single response and provides a lower bound for performance comparison. This method does not scale test-time compute or employ verification.

\noindent \textbf{Majority Voting}: A verifier-free approach that generates multiple candidate responses and selects the most frequent final answer across all responses \citep{brown2024largelanguagemonkeysscaling, chen2024llmcallsneedscaling, snell2024scalingllmtesttimecompute}. 
This method leverages repeated sampling but does not use verification models to assess response quality. 
Instead, it relies on the assumption that correct answers will appear more frequently than incorrect ones across multiple generations.

\subsubsection{Alternative Verification Strategies}

\noindent \textbf{Naive Unweighted Aggregation}: 
We consider three oracle configurations using the top-1, top-5, and top-10 verifiers (ranked by their agreement with ground-truth labels). Across all datasets, these oracle ensembles substantially outperform baselines. On average, the best-performing unweighted ensembles exceed first-sample performance by $20.3\%$ and outperform majority voting by $15.0\%$ (see \autoref{fig:MV_vs_Naive_vs_Weighted}). For more difficult benchmarks such as GPQA and MMLU Pro, the top-5 and top-10 ensembles consistently outperform top-1, suggesting that verifier diversity is especially beneficial on challenging examples. However, these oracle ensembles rely on access to ground truth to rank verifiers, limiting their use in practice and motivating the need for learned, unsupervised weighting. %

\noindent \textbf{Naive Bayes}: 
We implement a Naive Bayes classifier that models the probability of response correctness given verifier scores: $P(y_{ij} = 1 | s_{ij1}, ..., s_{ijm}) = \frac{P(s_{ij1}, ..., s_{ijm} | y_{ij} = 1) P(y_{ij} = 1)}{P(s_{ij1}, ..., s_{ijm})}$. 
Under the conditional independence assumption, this factorizes as $P(s_{ij1}, ..., s_{ijm} | y_{ij} = 1) = \prod_{k=1}^m P(s_{ijk} | y_{ij} = 1)$. 
We estimate the parameters using labeled data from the development set. 
This approach provides a probabilistic framework for aggregating verifier outputs but requires labeled data for parameter estimation.

\noindent \textbf{Logistic Regression}: 
We train a logistic regression classifier where the input features are the verifier scores $[s_{ij1}, ..., s_{ijm}]$ and the output is the correctness of each response: $P(y_{ij} = 1 | \mathbf{s}{ij}) = \sigma(\mathbf{w}^T \mathbf{s}{ij} + b)$, where $\sigma$ is the sigmoid function. 
The weights $\mathbf{w}$ and bias $b$ are learned using labeled training data. 
This supervised approach can capture more complex relationships between verifier outputs than naive averaging but requires substantial labeled data for effective training.

\noindent \textbf{Multi-Agent Verification (MAV) \citep{lifshitz2025multiagentverificationscalingtesttime}}: 
This approach combines multiple "Aspect Verifiers" (AVs) - off-the-shelf LLMs prompted to verify specific aspects of candidate outputs through binary True/False approvals. 
Unlike reward models, AVs require no additional training and can be easily combined through voting mechanisms. 
The MAV framework uses BoN-MAV (Best-of-N with Multi-Agent Verification), which: \textbf{(1)} samples $n$ candidate outputs from a generator LLM, \textbf{(2)} collects binary approvals from multiple aspect verifiers that vary across three dimensions (base LLM, aspect to verify, and verification strategy), and \textbf{(3)} selects the output with the most approvals. 
In our implementation, we use Llama 3.3 70B Instruct as the judge model rather than Gemini 1.5 Flash/Pro as used in the original paper.

\noindent \textbf{Self-Verification \citep{zhao2025sample}}: This method implements a sophisticated sampling-based search approach where models verify their own responses through detailed natural language analysis. The approach goes beyond simple self-critique by using structured verification prompts that: \textbf{(1)} rewrite candidate responses in rigorous mathematical theorem-lemma-proof format, \textbf{(2)} systematically scan for errors through step-by-step analysis, and \textbf{(3)} compare responses to localize potential mistakes. 
The method leverages two key principles: comparing across responses provides signals about error locations (since models struggle with error recall but can identify errors when given their locations), and different output styles are optimal for different tasks (chain-of-thought for generation, rigorous mathematical format for verification). 
This approach differs from naive self-verification by using structured, multi-step verification protocols rather than simple correctness judgments.

\input{tables_and_figures/LR_and_NB_across_datasizes}

\newpage
\subsection{Scaling Trends of \weaver{}}
\label{app:scaling_trends_of_weaver}

Scaling laws describe how performance metrics such as accuracy, sample efficiency or compute cost change as we scale controllable resources, i.e. the number of trials \(K\), model capacity.
\citep{kaplan2020scalinglawsneurallanguage} showed that, for fixed‐parameter Transformer language models, the cross-entropy loss decreases as a power-law in both model size and data. This framework has since been extended to explore optimal tradeoffs between model and data scaling \cite{hoffmann2022trainingcomputeoptimallargelanguage}, as well as inference-time scaling with multiple samples \citep{chen2021evaluating,brown2024largelanguagemonkeysscaling}. 

First, we establish the power law scaling of the Pass@K rate. Assume the \(i\)-th problem has an unknown “difficulty’’ \(p_i\in[0,1]\), the probability that one response is correct.  With \(K\) independent samples, the chance we get at least one correct response is
\begin{equation*}
q_i(p_i, K) = 1 - (1 - p_i)^K \approx 1 - \exp(-p_i \cdot K) \quad \text{for small } p_i
\end{equation*}

Define the indicator variable:
\[
X_i = 
\begin{cases}
1, & \text{if the i-th query is solved at least once (with probability $q_i$)}, \\
0, & \text{otherwise},
\end{cases}
\]
and let the total number of solved problems be $Y = \sum_{i=1}^N X_i$.

The expected coverage $(\text{Pass@K})$ is the expected fraction of problems solved after trying $K$ times per problem:
\begin{equation*}
\text{Pass@K} := \mathbb{E}[Y]/N = \frac{1}{N}\sum_{i=1}^N (1 - (1 - p_i)^K)
\end{equation*}

To model population-level variation in problem difficulty, we assume each problem’s correctness probability $p_i$ is drawn from a Beta distribution: $p_i \sim \text{Beta}(\alpha, \beta)$.  This captures the idea that some problems are easier (high $p_i$) while others are harder (low $p_i$), with the overall distribution controlled by the shape parameters $\alpha,~\beta$. Then, the fraction of problem that can be solved in $K$ attempts follows,
\begin{align}
\text{Pass@K} &= \mathbb{E}_{p \sim \text{Beta}(\alpha, ~\beta)}[1- (1 - p)^K] \nonumber \\
&= 1 - \mathbb{E}_{p \sim \text{Beta}(\alpha, ~\beta)}[(1 - p)^K] = 1 - \frac{B(\alpha,~\beta + K)}{B(\alpha,~\beta)}
\label{eqn:coverage_beta}
\end{align}
by the definition of the Beta function $B(\cdot, \cdot)$. Taking logarithm:
\[
\log \text{Pass@K} = \log\left(1 - \frac{B(\alpha,~\beta + K)}{B(\alpha,~\beta)}\right) \approx - \frac{B(\alpha, ~\beta + K)}{B(\alpha,~\beta)}
\]
by $\log(1 - x) \approx -x$ when $x$ is small, which holds for large $K$. Then, expressing the Beta function in terms of the Gamma function leads to:
\[
\log \text{Pass@K} \approx - \frac{\Gamma(\beta + K)\Gamma(\alpha + \beta)}{\Gamma(\beta)\Gamma(\alpha + \beta + K)}
\]

For large $K$, we can apply Stirling's approximation of the Gamma function $\log \Gamma(x) \approx x \log x - x + \frac{1}{2} \log(2\pi) + \frac{1}{2} \log x $:
\begin{align*}
\log[-\log \text{Pass@K}] 
&= \log \Gamma(\beta + K) + \log \Gamma(\alpha + \beta) - \log \Gamma(\beta) - \log \Gamma(\alpha + \beta + K) \\
&\approx (\beta+K)\log (\beta+K) - (\alpha+\beta+K) \log(\alpha+\beta+K) + \frac{1}{2} \log \left(\frac{\beta+K}{\alpha + \beta +K}\right)\\
&\approx (\beta + K) \log K - (\alpha + \beta + K) \log K + \text{const} \\
&= -\alpha \log K + \log \zeta
\end{align*}

when we retain the leading term. 
In turn, the log of the expected coverage follows a power law in $K$, scaling as:
\begin{equation}
\log \text{Pass@K} = - \exp \left(-\alpha~\log K + \log \zeta\right) = -\zeta K^{-\alpha}    
\label{eqn:coverage_scaling}
\end{equation}

\paragraph{Verifier Success Modeling} 
Now suppose we pass the \(K\) candidates through a scoring model ("verifier'') which selects the top-scoring answer.  The verification process succeeds if (i) at least one correct answer was generated and (ii) the verifier ranks a correct answer highest.

\begin{equation}
\text{Selection@1}(K) := \mathbb{P}[\text{top-scoring response is correct}] 
\end{equation}

Assume the verifier assigns scores such that correct responses are drawn from a score distribution \( f_1 \), and the incorrect responses from a distribution \( f_0 \). Let \( s^{(1)} = \{s_j : y_j = 1\} \) and \( s^{(0)} = \{s_j : y_j = 0\} \) denote the scores of correct and incorrect responses, respectively. Then a query is successfully verified if:
\[
\text{Selection@1} = \mathbb{P}\left[ \max s^{(1)} > \max s^{(0)} \right]
\]
Our goal is to compute the probability that the maximum of \( c \) i.i.d draws from \( f_1 \) exceeds the maximum of \( K - c \) draws from \( f_0 \). 

To model the correctness of responses, we assume each query \( i \) has a latent correctness probability \( p_i \sim \text{Beta}(\alpha, \beta) \), reflecting query-specific difficulty. Given \( p_i \), each of the \( K \) responses is sampled independently as:
\[
y_{ij} \sim \text{Bernoulli}(p_i), \quad j = 1, \dots, K
\]
This implies the number of correct responses follows a Binomial distribution:
\[
C_i = \sum_{j=1}^K y_{ij} \sim \text{Binomial}(K, p_i)
\]
assuming (1) conditional independence of responses given \( p_i \), (2) identical correctness probabilities within a query, and (3) a fixed number of responses \( K \).

Because the correctness probability \( p \) varies across queries, the dataset-level Selection@1 curve requires marginalizing over $p$:
\[
\text{Selection@1}(K) = \mathbb{E}_{p \sim \text{Beta}(\alpha, \beta)} \left[ \text{Selection@1}(K \mid p) \right]
\]
Combined with the need to model max comparisons over verifier scores, it renders the exact calculation of Selection@1 analytically intractable.

To enable tractable, smooth modeling of Selection@1, we introduce the following parametric form:
\begin{equation}
\text{Selection@1}(K) \approx \exp(-\zeta K^{-\alpha}) \cdot \left(1 - (1 - \pi)^{K^\gamma} \right)
\label{eqn:select1_scaling}
\end{equation}
\begin{itemize}
\item The \textbf{coverage term} \( \exp(-\zeta K^{-\alpha}) \) approximates the probability that at least one correct response is generated.
\item The \textbf{verification term} \( 1 - (1 - \pi)^{K^\gamma} \) approximates the chance that the top-scoring response is correct, given that at least one correct response exists. The parameter \( \gamma \) controls whether verifier performance improves sublinearly or superlinearly with \( K \). The parameter $\pi$ represents the effective per-response probability that a correct response is successfully selected by the verifier, conditioned on the response being correct and included in the candidate set. %
\end{itemize}

To obtain practical scaling trends, we fit parametric models in \cref{eqn:select1_scaling} to the empirical averages computed from $5$ independent runs for each value of $K$, across each dataset and verification strategy. Specifically, we use the L-BFGS-B algorithm to optimize a smooth approximation following \cite{hoffmann2022training}. To ensure numerical stability and robustness to outliers or heavy-tailed noise in the observed selection accuracies, we minimize the Huber loss between the predicted values and the empirical means. The Huber loss behaves quadratically for small residuals and linearly for large ones, making it less sensitive to outliers than mean squared error (MSE) while maintaining smooth differentiability for gradient-based optimization. It is defined as,
\begin{equation*}
L_\delta(r) =
\begin{cases}
\frac{1}{2} r^2 & \text{if } |r| \leq \delta \\
\delta \left( |r| - \frac{1}{2} \delta \right) & \text{otherwise}
\end{cases}    
\end{equation*}
where $\delta > 0$ is a tunable threshold that controls the transition between the two regimes. We search over $\delta \in \{0.01, 0.05, 0.1, 0.25, 0.5\}$ to select the value that yields the best fit.

Additionally, we introduce floor and ceiling parameters to bound the predicted values and model saturation behavior. The floor accounts for the irreducible failure rate even at high $K$, while the ceiling models the upper bound on achievable performance (e.g., due to imperfect verifiers or ambiguous problems). The final fitted form is:
\begin{equation}
\text{Selection@1}(K) \approx \text{floor} + (\text{ceil} - \text{floor}) \cdot \exp(-\zeta K^{-\alpha}) \cdot \left(1 - (1 - \pi)^{K^\gamma} \right)
\label{eqn:select1_scaling_bounded}
\end{equation}
We can use an unbiased estimator to evaluate best-of-$k$ selection accuracy when a fixed verifier is used to rank responses, as described in \cite{singhi2025whentosolve}.  However, in the case of \weaver{}, the development set constitutes 1\% of the data and is itself selected based on the value of $K$. In turn, the ranking of responses is no longer independent of $K$, introducing bias into the best-of-$k$ estimate.As a result, we instead rely on Monte Carlo estimates to approximate best-of-$k$ performance, sampling $k$ responses multiple times and computing the average accuracy of the top-ranked output under the $K$-dependent verifier. We use an unbiased estimator for coverage, as described in \cite{chen2021evaluating}.

\cref{fig:power_law_fit70B} and \cref{fig:power_law_fit8B} along with \cref{tab:power_law_fit}  illustrate how the different verification strategies scale with the number of generations and the fit to the parametric form in  \cref{eqn:select1_scaling}. Each method exhibits characteristic scaling behavior that aligns with \cref{eqn:select1_scaling}. \weaver{} demonstrates improved performance over naive ensembles and majority voting. The fitted parameters in \cref{tab:power_law_fit} quantitatively capture these trends across datasets, providing evidence that the parametric from in \cref{eqn:select1_scaling_bounded} closely model empirical outcomes.  \cref{fig:power_law_pred70B} and \cref{fig:power_law_pred8B} along with \cref{tab:power_law_pred} illustrate the predictive performance of the parametric form in \cref{eqn:select1_scaling_bounded}, showing that models fit on subsets of $K$ can extrapolate to unseen values of $K$.\par 

\input{tables_and_figures/scaling_trends}

\newpage
\subsection{Scaling Candidate Generations}
\label{app:scaling_candidate_generations}

\input{tables_and_figures/FalsePositiveRates}

\input{tables_and_figures/8B_verifier_ablations}

\input{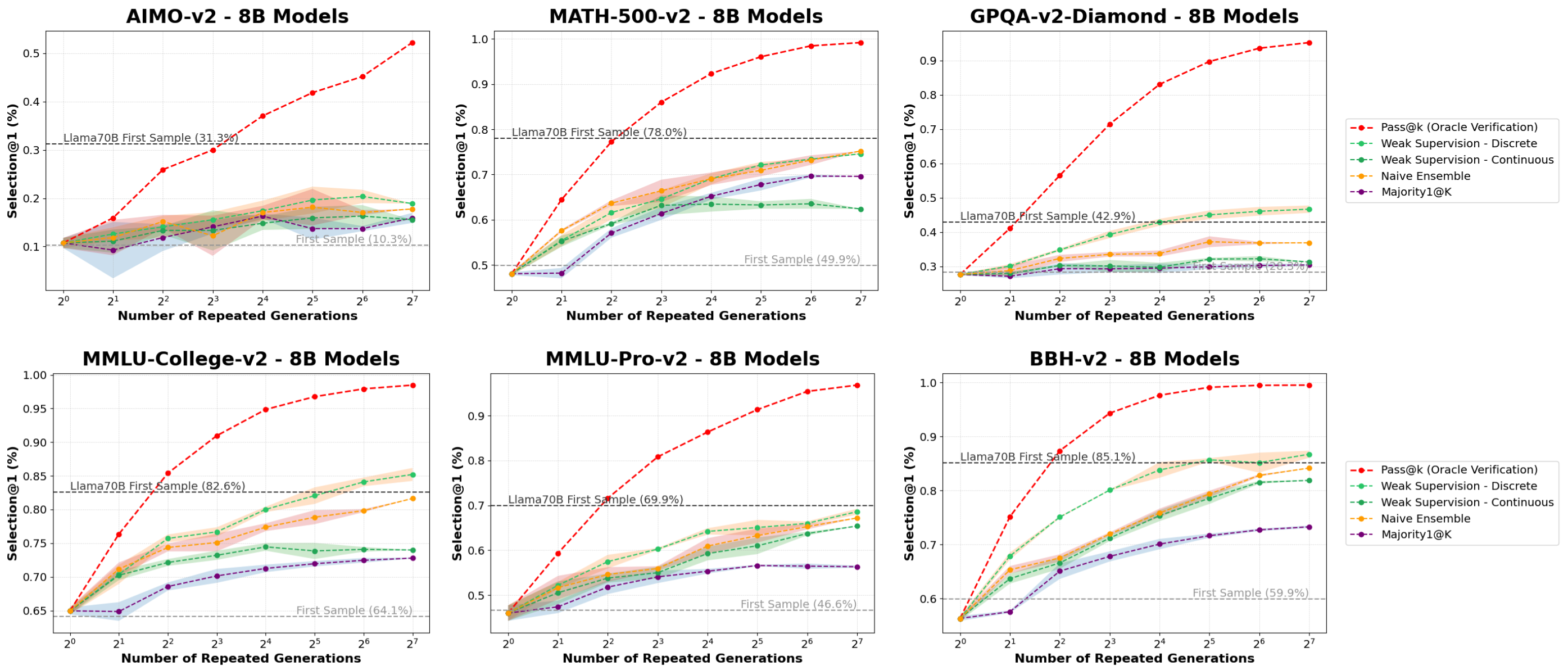}

\subsection{Scaling Verifier Count}
\label{app:scaling_verifier_count}

In Table \ref{tab:sampling_vs_ensembling}, we include results of scaling verifier scores.  We note that for reward models (RMs), which are typically deterministic \citep{lambert2024rewardbenchevaluatingrewardmodels, song2025prmbenchfinegrainedchallengingbenchmark}, multiple scores must be obtained by varying the prompt; for LM Judges, we can vary either the prompt or the sampling temperature to generate diverse outputs from the same model (Table \ref{tab:sampling_vs_ensembling}). We find that for both types of weak verifiers, RMs and LM judges, scaling the number of models yields better performance than sampling multiple evaluations from the same model via prompt tuning or temperature variation. 
{However, we note that these approaches are complementary.}

When breaking down weak verifiers into RMs or LM Judges, individually, we find that additional LMs leads average gains of 5.4\% and 6.1\%, respectively (Table \ref{tab:sampling_vs_ensembling}). 
In contrast, sampling additional scores from a single RM or LM judge yields only 0.8\% and 1.1\% gains on average. These results suggest that leveraging the complementary strengths of multiple verifiers can be more effective than eliciting multiple judgments from a single verifier. 
\cref{app:individual-verifier-optimization} provides additional details on the verifier prompting. 
Finally, Figure \ref{fig:Verifiers_vs_Generations_Tradeoff} illustrates the tradeoff of scaling the number of verifiers versus increasing the number of scores from a single verifier, showing that scaling verifiers is helpful when the coverage increases as we increase sample count.

\input{tables_and_figures/sampling_vs_ensembling}

\input{tables_and_figures/Verifiers_vs_Generations}

\subsection{\weaver{} Distillation}
\label{app:weaver_distillation}

\input{tables_and_figures/distillation_diagram}

For the loss function in \weaver{} distillation, we utilized cross-entropy loss with Adam \citep{kingma2015adam}. 
Our classification architecture comprises a single linear classification layer with 0.1 dropout applied to the input, which consists of the final hidden state from the $[CLS]$ token. 
Regarding learning dynamics, we implemented linear warmup and linear decay via the Sentence-Transformers library \citep{reimers-2020-multilingual-sentence-bert}, employing a learning rate of 5e-6 and training batch size of 64 across all experimental setups.

\input{tables_and_figures/WeaverDistilledParetoFrontiers}
\input{tables_and_figures/Weaver_Distillation_Ablations}

\newpage

\subsection{Individual Verifier Optimization}
\label{app:individual-verifier-optimization}

While \weaver{} primarily focuses on aggregating multiple weak verifiers to improve overall verification quality, this appendix explores complementary techniques for optimizing individual verifiers. As mentioned earlier in the paper, existing weak verifiers often suffer from high false positive rates \cite{stroebl2024inferencescalingflawslimits}, which can limit their effectiveness, even within an ensemble.

As we scale the number of repeated samples and employ multiple verifiers, the precision of each individual verifier becomes increasingly important relative to recall. When many candidate solutions are available, a verifier can afford to miss some correct solutions (false negatives) as long as its positive predictions are highly reliable (high precision).

This observation motivates exploring methods to enhance individual verifier quality through methods such as prompt optimization — tailoring verifier prompts to maximize performance, particularly precision, with minimal or no labeled data.

\subsubsection{LM Judge Prompt Optimization}
LM judges often suffer from biases such as position bias (favoring answers in certain positions), verbosity bias (preferring longer answers), and self-enhancement bias (preferring answers similar to their own generation patterns) \cite{zheng2023judging, alpaca_eval}, suggesting sensitivity to system and input prompt design.

Throughout our \weaver{} experiments, we used fixed, manually engineered prompts for our LM judge verifiers. However, optimizing these prompts could potentially improve individual verifier precision and reliability. Multi-Agent Verification \cite{lifshitz2025multiagentverificationscalingtesttime} demonstrates this by crafting specialized prompts for specific verification aspects.

We explored systematically optimizing verifier prompts using DSPy \cite{khattab2023dspy}, an open-source library that provides algorithms for optimizing language model prompts through discrete search over prompt candidates guided by a metric function. DSPy optimization works by generating, evaluating, and refining prompts that maximize task performance on a small labeled dataset.

\noindent \textbf{Experimental Setup}: We investigate two dimensions of prompt optimization: (1) \textbf{optimization space scaling}, where we progressively expand what the optimizer can modify from system instruction only (0-shot) to including 3 demonstrations (3-shot) and 5 demonstrations (5-shot); and (2) \textbf{training data size scaling}, where we vary labeled data from 1\% to 16\% to determine how much data is necessary for effective prompt optimization.

Our experimental setup uses training examples containing instruction-generation pairs. Since our datasets have multiple generations per instruction (up to 100), we group examples by instruction before splitting to prevent data leakage between train and validation sets. We hold out 50\% of the dataset instructions (each paired with 100 candidate generations) for evaluation. For the optimization space scaling experiment, we randomly select $n$ generations such that $n \times \text{len(dataset)}/2 = 250$, maximizing training set diversity while maintaining a fixed training set size. For the data scaling experiment, we train on different percentages (1\%, 2\%, 4\%, and 16\%) of the dataset by calculating the number of instructions as $\lceil\text{num\_problems\_in\_dataset} \times (\text{train\_percentage}/100)\rceil$ and selecting repeated samples for each instruction with $\text{samples} = \min(\max(4, \text{num\_problems} \times 2), 20)$ to avoid overfitting. We use a consistent random seed to ensure identical dataset splits between optimization runs.

\begin{figure}[t]
\centering
\includegraphics[width=\textwidth]{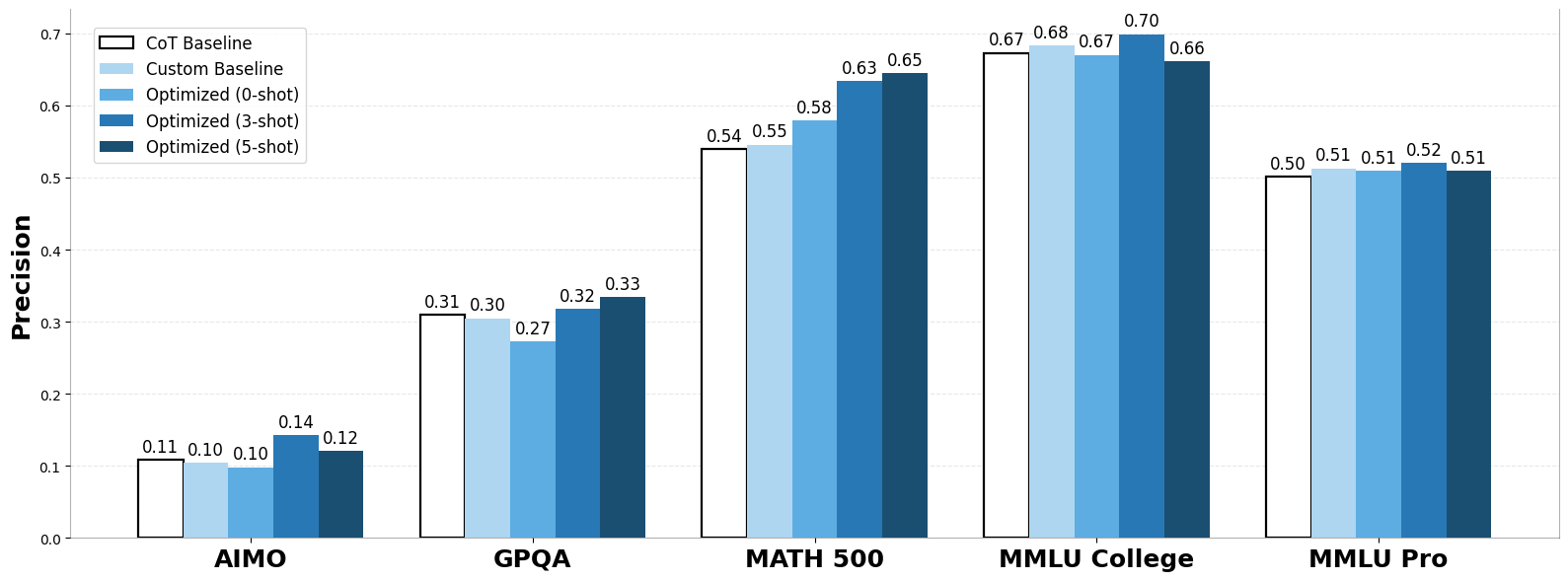}
\caption{\textbf{LM judge prompt optimization using 250 labeled examples consistently yields precision gains}. Baseline methods (CoT and Custom) are compared against DSPy-optimized prompts with varying numbers of demonstrations (0-shot, 3-shot, and 5-shot).}
\label{fig:judge_performance}
\end{figure}

\noindent \textbf{Results}: Figure~\ref{fig:judge_performance} shows results across different datasets and optimization configurations. While we don't observe clear scaling relationships across all datasets (possibly due to the increased stochasticity of LLM-based optimization), we observe an average precision gain of 3.8\% of the best judge over the chain-of-thought (CoT) baseline judge. MATH500 shows the largest jump in precision of 9\% and shows clear improvement in precision as the optimization space is scaled.

\begin{figure}[t]
\centering
\includegraphics[width=\textwidth]{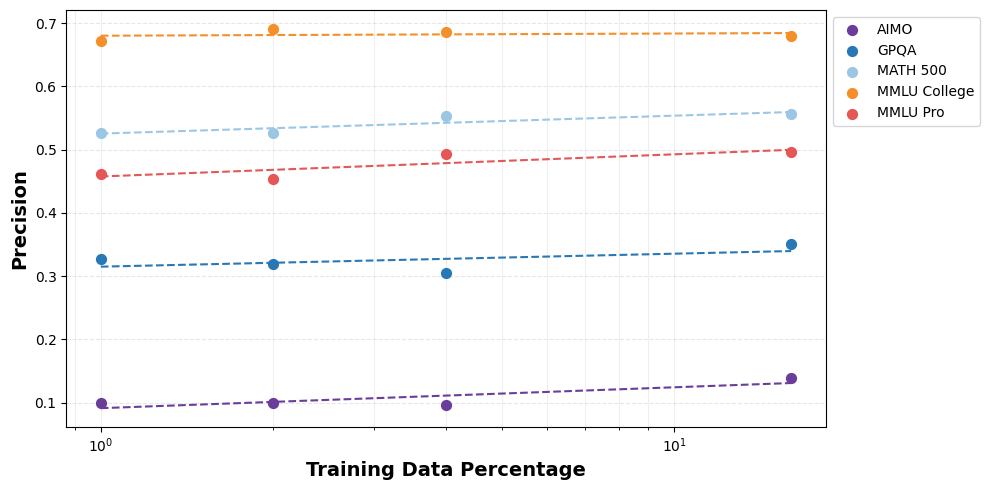}
\caption{\textbf{Scaling LM judge prompt optimization training data leads to modest precision gains.} The x-axis shows the percentage of training data used (log scale), and the y-axis shows precision.}
\label{fig:scaling_behavior}
\end{figure}

The scaling behavior with training data size (Figure~\ref{fig:scaling_behavior}) shows slight log-linear improvements in precision as we increase training data, though gains differ by dataset. MMLU-College shows minimal benefit from additional data, while the remaining datasets see an average boost of 3.2\% in precision when scaling the training data size from 1\% to 16\% of the original dataset. 

\begin{figure}[t]
\centering
\includegraphics[width=\textwidth]{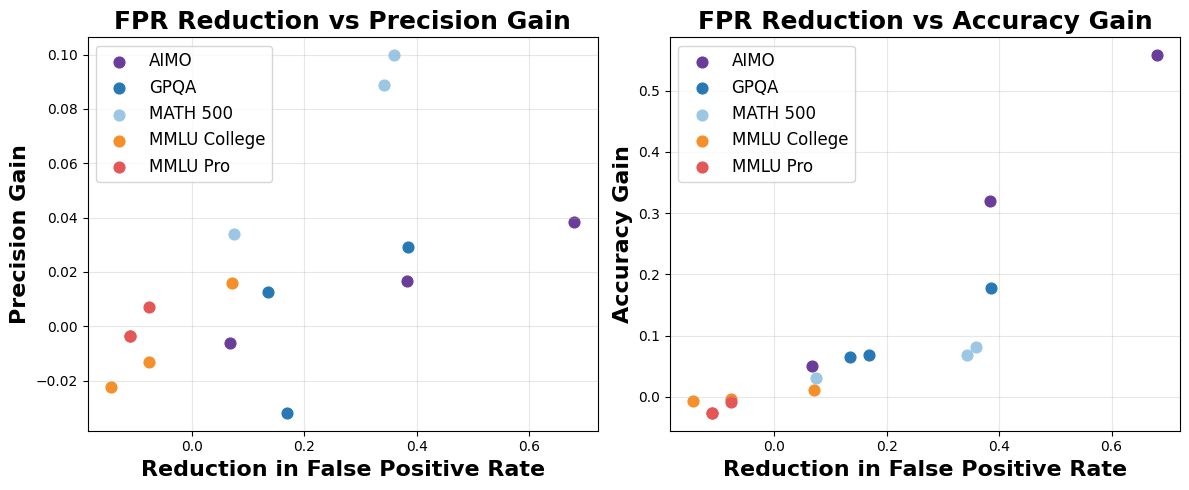}
\caption{\textbf{Optimized prompts often improve LM judge performance by reducing false positive rates.}}
\label{fig:false_positive}
\end{figure}

(Figure~\ref{fig:false_positive}) reveals that optimized prompts often improve both precision and accuracy by reducing false positive rates - essentially making judges more conservative in their correctness assessments. This is particularly valuable in the repeated sampling regime, where higher precision improves overall verification quality.

These findings suggest that prompt optimization can be a valuable complement to \weaver{}'s aggregation approach. Even with limited labeled data, targeted prompt engineering can enhance individual verifier quality, benefiting the ensemble as a whole. Further research is needed to define a more systematic recipe for verifier prompt optimization. Additionally, it remains a question of whether we can extend prompt optimization to discriminative reward models to enjoy similar gains in performance.

\section{Miscellaneous}
\label{app:miscellaneous}

\subsection{Compute Requirements}
\label{app:compute_requirements}

\textbf{Hardware Infrastructure.} Our experiments were conducted using 4 compute nodes, each equipped with 8 NVIDIA H100 GPUs (80GB HBM3 memory per GPU), for a total of 32 H100 GPUs. Each node was configured with high-bandwidth NVLink connections between GPUs and inter-node communication was facilitated via NVIDIA NVLink Switch System to minimize communication overhead during distributed training and inference.

\textbf{Model Parallelism and Distribution.} For our 72B parameter language models, we employed a hybrid parallelism strategy combining tensor parallelism, pipeline parallelism, and data parallelism:
\begin{itemize}
    \item 8-way tensor parallelism across GPUs within each node
    \item 4-way pipeline parallelism across nodes
    \item Data parallelism for batch processing
\end{itemize}

\textbf{Storage Requirements.} Processing datasets of 100GB+ required significant storage infrastructure:
\begin{itemize}
    \item 4TB NVMe SSDs per node for dataset caching and checkpoints
    \item 100TB shared network storage for full dataset repository
\end{itemize}

\textbf{Software Stack.} Our experiments were powered by:
\begin{itemize}
    \item NVIDIA CUDA 12.2
    \item PyTorch 2.1 with NVIDIA NCCL for distributed communication
    \item DeepSpeed ZeRO Stage 3 for memory optimization
    \item Distributed data loading with webdataset format for efficient streaming
\end{itemize}

%% file: tables_and_figures/binarization.tex
\begin{figure}[H]
   \centering
   \includegraphics[height=0.9\textheight,width=\linewidth]{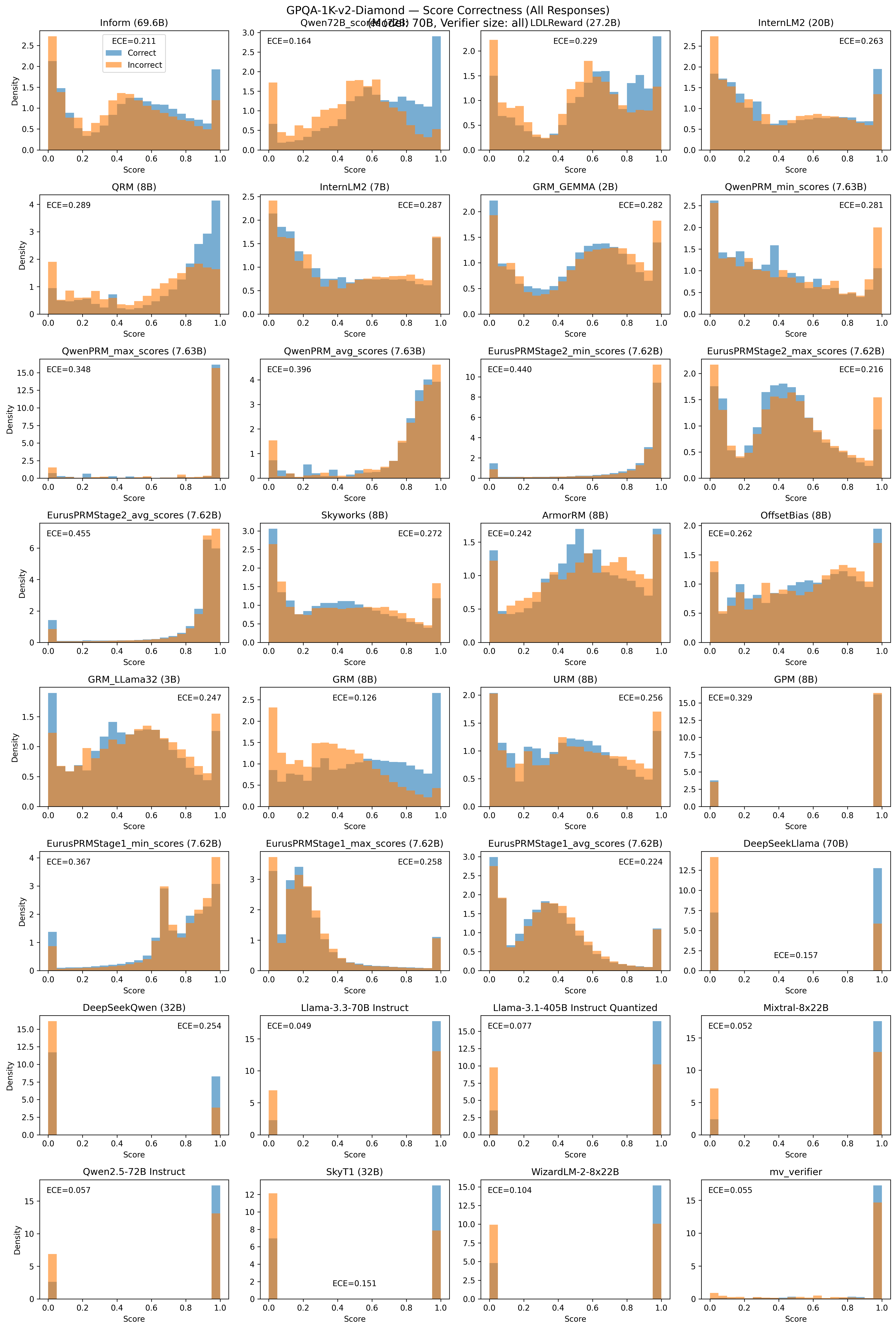}
   \caption{
   Accuracy of verifiers on the GPQA dataset. For each verifier, we compute the fraction of problems for which the top-ranked response (according to its score) is correct. While some verifiers consistently select high-quality answers, others perform near chance or worse, motivating the need to filter out low-quality verifiers before applying weak supervision.}
   \label{fig:histogram_verifier_accuracy_gpqa}
\end{figure}

\begin{figure}[H]
   \centering
   \includegraphics[width=\linewidth]{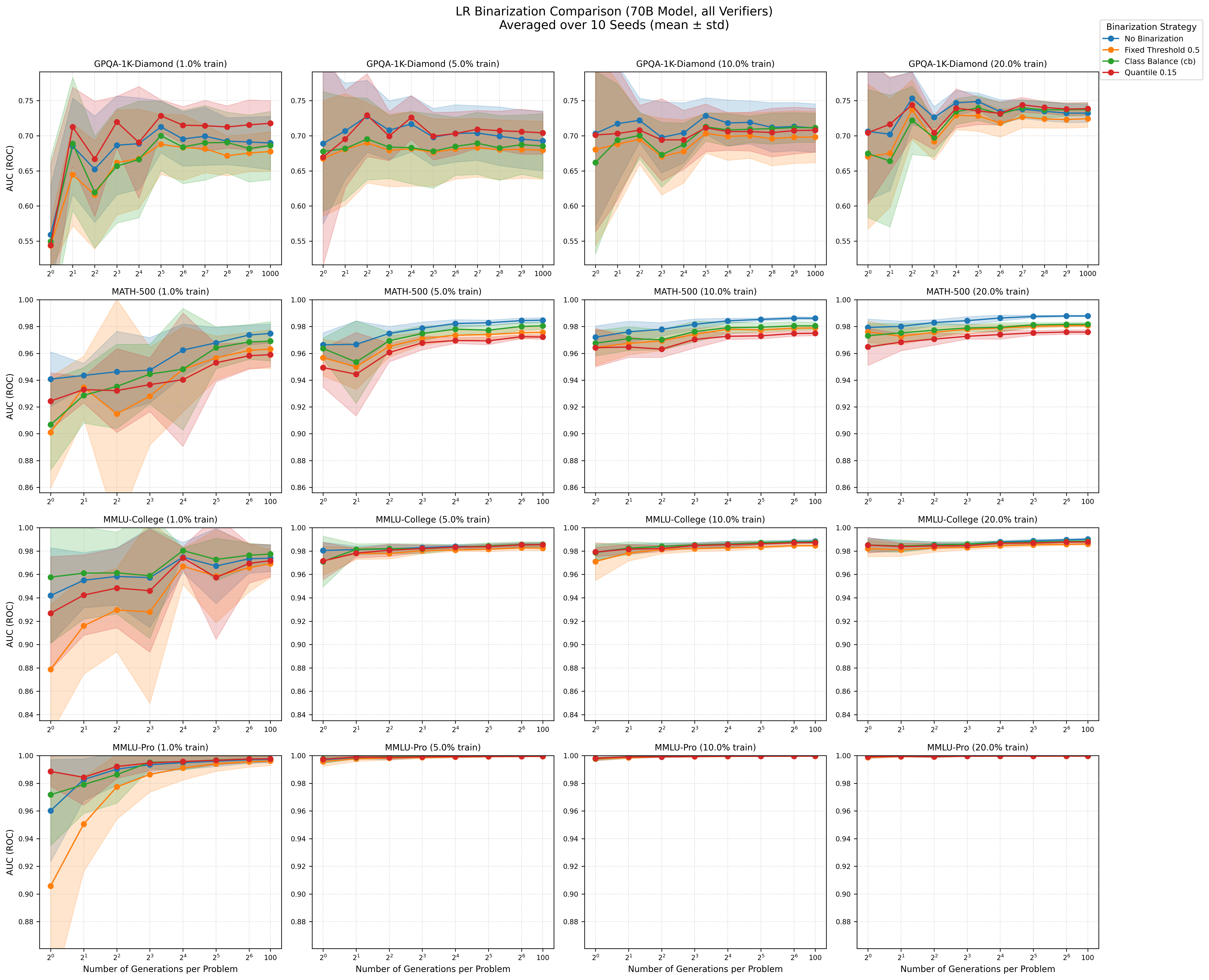}
   \caption{
   \textbf{}
   Same conventions as \cref{fig:lr_binarization_select_acc} where the performance reported is the area under the curve of the ROC.
   }
   \label{fig:lr_binarization_auc}
\end{figure}

\begin{figure}[H]
   \centering
   \includegraphics[width=\linewidth]{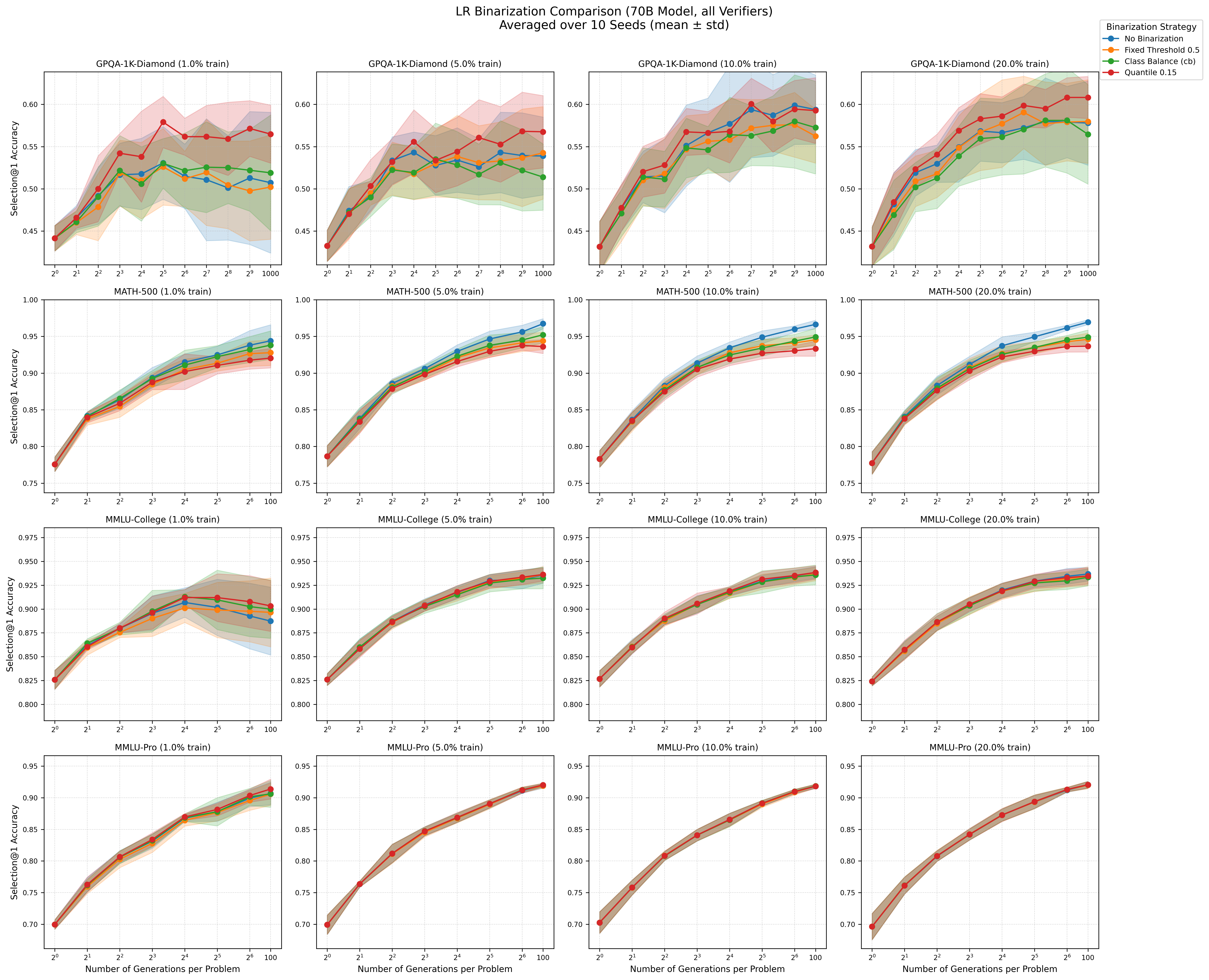}
   \caption{
   \textbf{}
   Performance for logistic regression models trained on the verifier outputs under different binarization strategies: (1) None: uses raw continuous scores without binarization, (2) Fixed Threshold: applies a uniform threshold across all data, (3) Class Balance: chooses a threshold per verifier so that the proportion of positive labels matches the true class distribution (4) Quantile: assigns positive labels to only the top ~15\% of scores, focusing on high-confidence predictions.  Results are shown across four datasets (GPQA-Diamond, MATH-500, MMLU-College, MMLU-Pro) and four training fractions $(1\%, 5\%, 10\%, 20\%)$. For each seed, we use a random subset of problems as the training set and report the performance on the remaining problems.  Curves report mean and standard deviation over multiple seeds. Performance is shown as a function of the number of generations per problem (x-axis, log scale).
   }
   \label{fig:lr_binarization_select_acc}
\end{figure}

%% file: sections/implementation_details.tex
\subsection{Adaptation Method}
\label{app:implementation_decisions}

\input{tables_and_figures/verifier_stats}

We now describe our proposed method for normalization, binarization and filtering of verifiers, after which the Weak Supervision algorithm described in Appendix~\ref{app:ws_discrete_knowndiff}.

\begin{enumerate}[leftmargin=*]
\item \textbf{Normalization}: To make verifier outputs comparable, we apply min-max normalization to each verifier:
\[
s' = \frac{s - \min(s)}{\max(s) - \min(s)} \in [0, 1].
\]
This ensures that all scores lie within the same numerical range and preserves relative orderings. For regression-style verifiers, normalization aligns their outputs with the scale of the labels and avoids numerical instability due to unbounded score ranges. Without normalization, aggregation methods may become biased or ill-conditioned due to disproportionate score magnitudes.

\item \textbf{Binarization}: We use a small amount of labeled samples $\Ddev$ (which we already are using to compute $\Pr(y=1)$) to determine a threshold for converting continuous verifier outputs to binary outputs.  \cref{fig:verifier_inv_cov_bin} illustrates the precision matrix of the verifier scores after binarization. It shows a damping of large off-diagonal dependencies and improved condition numbers. 
Table \ref{tab:weaver_dev_set_ablation} shows that with only 5 to 10 labeled queries from benchmark development sets (which $\leq$ 1\% of the evaluation set), we can estimate binarization thresholds that bolster performance by averages of 8.4\% for the 70B models when compared to binary splitting along the median score of the dataset.

\item \textbf{Filtering out low-quality verifiers}: To mitigate the impact of low-quality verifiers, we prune verifiers with extreme marginal behavior, depending on the class balance.  For datasets with estimated class balance between 20\% and 80\%, we filter out verifiers with positive rates outside this range. If a dataset has fewer than 20\% positive samples overall, we remove verifiers that predict positives more than 80\% of the time.
Conversely, for datasets with more than 80\% positives, we drop verifiers that predict positives less than 20\% of the time.

\Cref{fig:verifier_inv_cov_bin_drop} illustrates the precision matrix of the verifier scores after both binarization and dropping low-signal or redundant verifiers. Compared to \cref{fig:verifier_inv_cov_bin}, it shows further attenuation of off-diagonal structure. Illustrating that dropping contributes substantially to decorrelating the verifier set, which can improve identifiability and numerical stability for downstream weak supervision. As shown in Table \ref{tab:weaver_dropping_verifiers_ablation}, verifier pruning leads to 12.5\% performance improvement for the 70B model setting.

\end{enumerate}

\input{tables_and_figures/weaver_dev_set_ablation}

\input{tables_and_figures/weaver_dropping_verifiers_ablation}

\input{tables_and_figures/weaver_adaptive_threshold_ablation}

\input{tables_and_figures/discrete_vs_continuous_for_LR}

\input{tables_and_figures/8B_unique_answer_correlations}

\subsection{Exploration: Clustering by Difficulty to Improve \weaver{}}
\label{app:clustering_exploration}

Weak verifiers often behave inconsistently across the difficulty spectrum of input queries. For instance, most verifiers may have very high accuracy on easy queries and low accuracy on more difficult queries. \cref{fig:verifier_avg_accuracy} and \cref{fig:verifier_selection_accuracy} illustrate the average and selection accuracy of each verifier across multiple problems and datasets. This confirms that there is significant variation in verifier performance at the per-query level. 
To capture this heterogeneity, we explore clustering queries by difficulty and fitting one \weaver{}'s weak supervision model per cluster, independently.

We define \textit{query difficulty} as the empirical ratio of correct to incorrect generations for each query. Using this as a proxy for problem hardness, we partition each dataset into evenly sized clusters along the difficulty distribution and learn separate weak supervision models for each cluster. This is done in an oracle setting, where difficulty is computed using ground-truth correctness, but no label information is used when training the cluster-specific verifier models.

This approach is adaptive in two senses: (1) we adapt the weak supervision model to the difficulty class of the query, and (2) we adapt the threshold of each reward model independently per cluster to better reflect local verifier behavior. For each cluster, we perform a grid search over reward model thresholds ranging from 0.05 to 0.95 in increments of 0.05 (19 values total), selecting the threshold that maximizes accuracy on a held-out development set. We experiment with clustering the queries into between 1 and 5 difficulty levels per dataset, using the oracle difficulty distribution to divide the queries into equally sized bins. While the clustering in our study uses oracle difficulty, future work could explore unsupervised approximations or semi-supervised approaches using a small labeled subset (e.g., 10\%) to estimate difficulty distributions.

In Tables~\ref{tab:weaver_cluster_selection} and~\ref{tab:optimizing_clusters_threshold_70b},l, we analyze how difficulty-aware clustering and threshold adaptation affect \weaver{}’s performance at different model scales.

\begin{itemize}

\item \textbf{70B model:} At this scale, we find that optimizing a single reward model threshold captures most of the verification signal, with clustering yielding only marginal gains ($\sim$1\%). 
This suggests that verifier behavior is relatively stable across query difficulties at higher model capacities.

\item \textbf{8B model:} In contrast, the 8B setting exhibits a larger gain (4.8\%) from clustering and adaptive thresholding. We attribute this to three key factors:

\begin{enumerate}
    \item \textbf{Higher verifier variance:} As shown in Table~\ref{tab:verifier_accuracies_ranges}, verifier quality fluctuates more across queries in the 8B setting, making cluster-specific models more beneficial. 
    \item \textbf{Fewer positive generations:} Table~\ref{tab:8B_positive_negative_ratios} shows that the 8B generator produces fewer correct answers overall, increasing difficulty heterogeneity.
    \item \textbf{Larger generation-verification gap:} The Pass@1-to-Pass@K gap is more pronounced for 8B (e.g., 49.8\% to 99.2\% on MATH500), indicating greater room for selection-based improvements.
\end{enumerate}

\end{itemize}

Table~\ref{tab:weaver_cluster_selection} confirms that increasing cluster count improves accuracy for the 8B model but often degrades it for the 70B model. These results suggest that difficulty-aware modeling is especially useful when verifier behavior is unstable and when the generation model produces sparse correct candidates.

\noindent\textbf{Further gains from per-model thresholding.} Finally, in Table~\ref{tab:optimizing_clusters_threshold_70b_per_model}, we introduce a finer-grained tuning strategy where each reward model receives its own threshold, rather than using a single global threshold per cluster. This approach provides modest but consistent improvements for the 70B model (e.g., +0.5\% on GPQA Diamond, +0.8\% on MMLU Pro), and more substantial boosts for the 8B model across all datasets (+1.6\% to +2.8\%). These gains highlight the value of adapting verifier aggregation strategies not just by query difficulty but also by verifier-specific behavior, especially at smaller model scales where noise is more pronounced.

\input{tables_and_figures/weaver_cluster_selection}

\input{tables_and_figures/optimizing_cluster_threshold}

\input{tables_and_figures/optimizing_reward_model_threshold_per_model}

%% file: tables_and_figures/verifier_stats.tex
\begin{figure}[H]
   \centering
   \includegraphics[width=\linewidth]{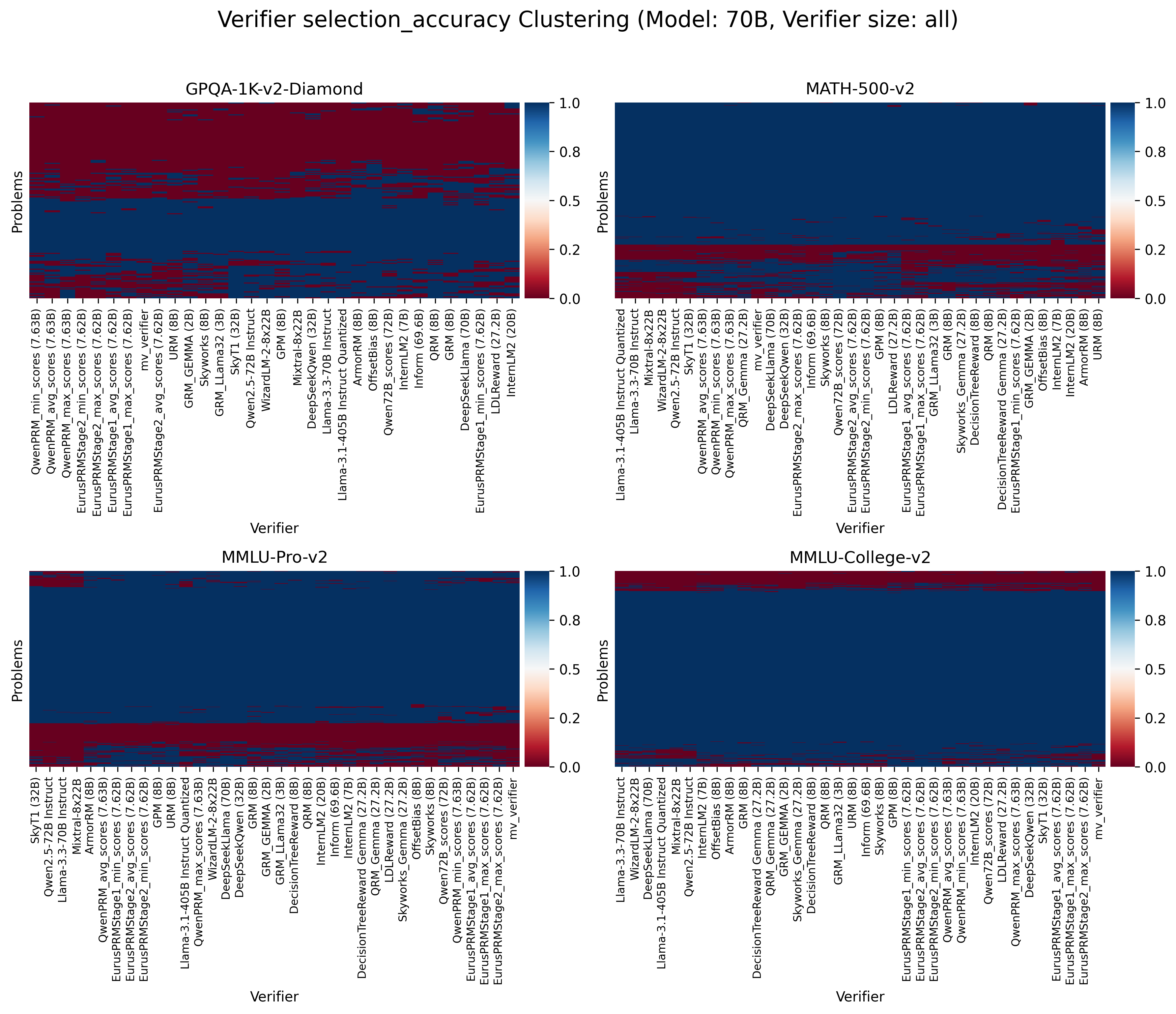}
   \caption{
   Selection accuracy for each verifier across problems and datasets. 
   }
   \label{fig:verifier_selection_accuracy}
\end{figure}

\begin{figure}[H]
   \centering
   \includegraphics[width=\linewidth]{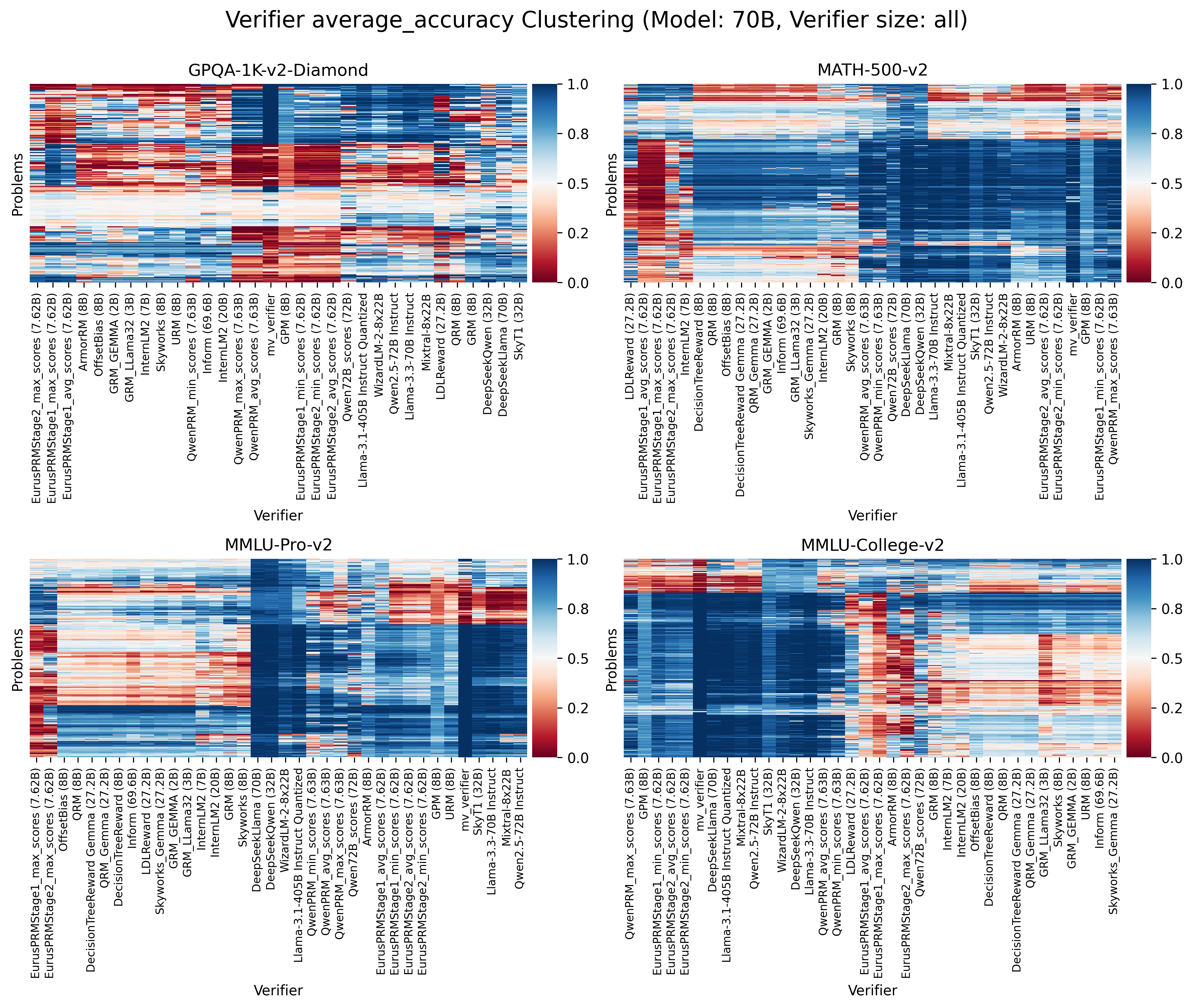}
   \caption{
   Average accuracy for each verifier across problems and datasets given Llama-70B responses scores.
   }
   \label{fig:verifier_avg_accuracy}
\end{figure}

\begin{figure}[H]
   \centering
   \includegraphics[width=\linewidth]{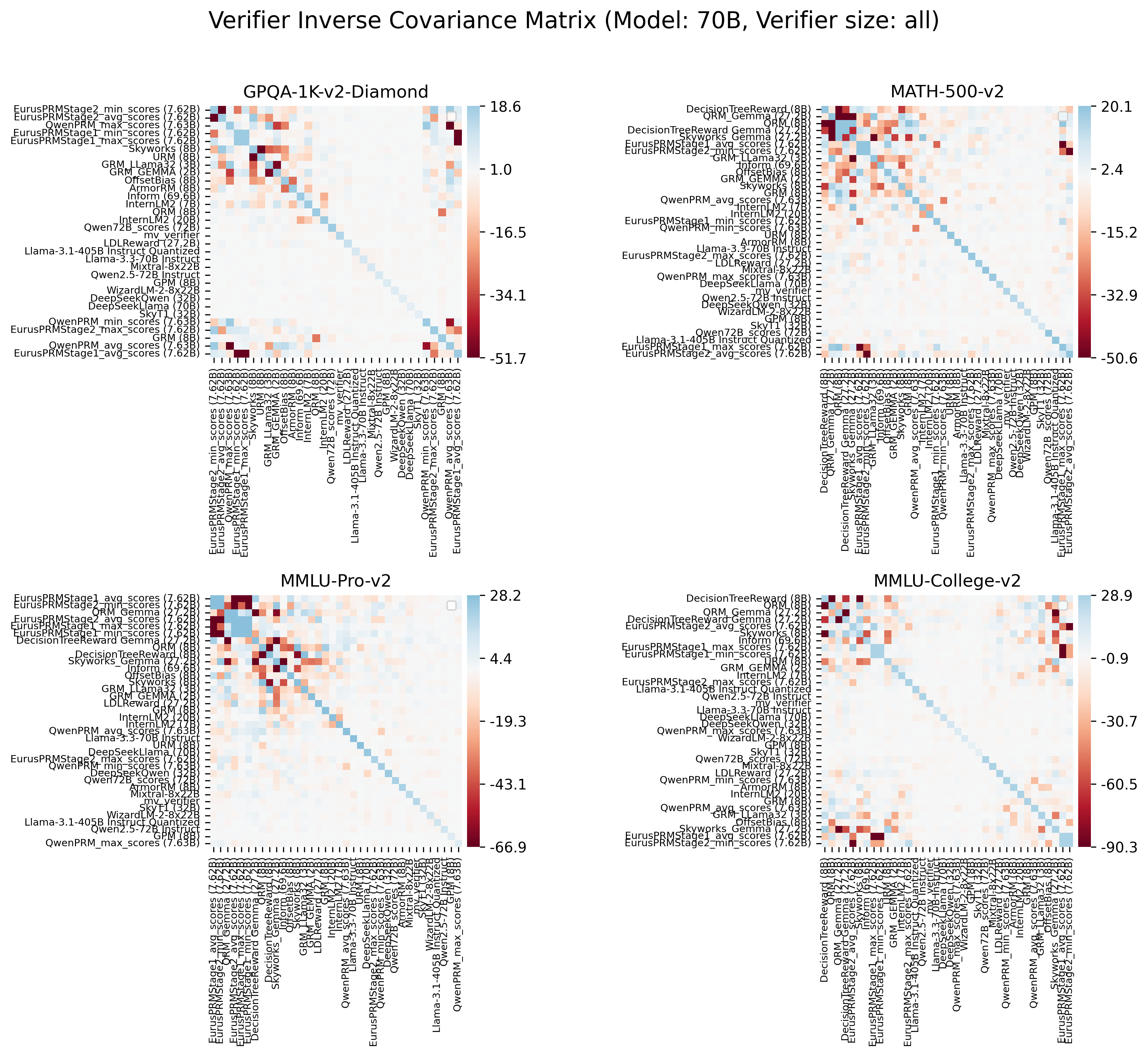}
   \caption{
   Inverse covariance matrix for each verifier across datasets given Llama-70B responses scores.
   }
   \label{fig:verifier_inv_cov_raw}
\end{figure}

\begin{figure}[H]
   \centering
   \includegraphics[width=\linewidth]{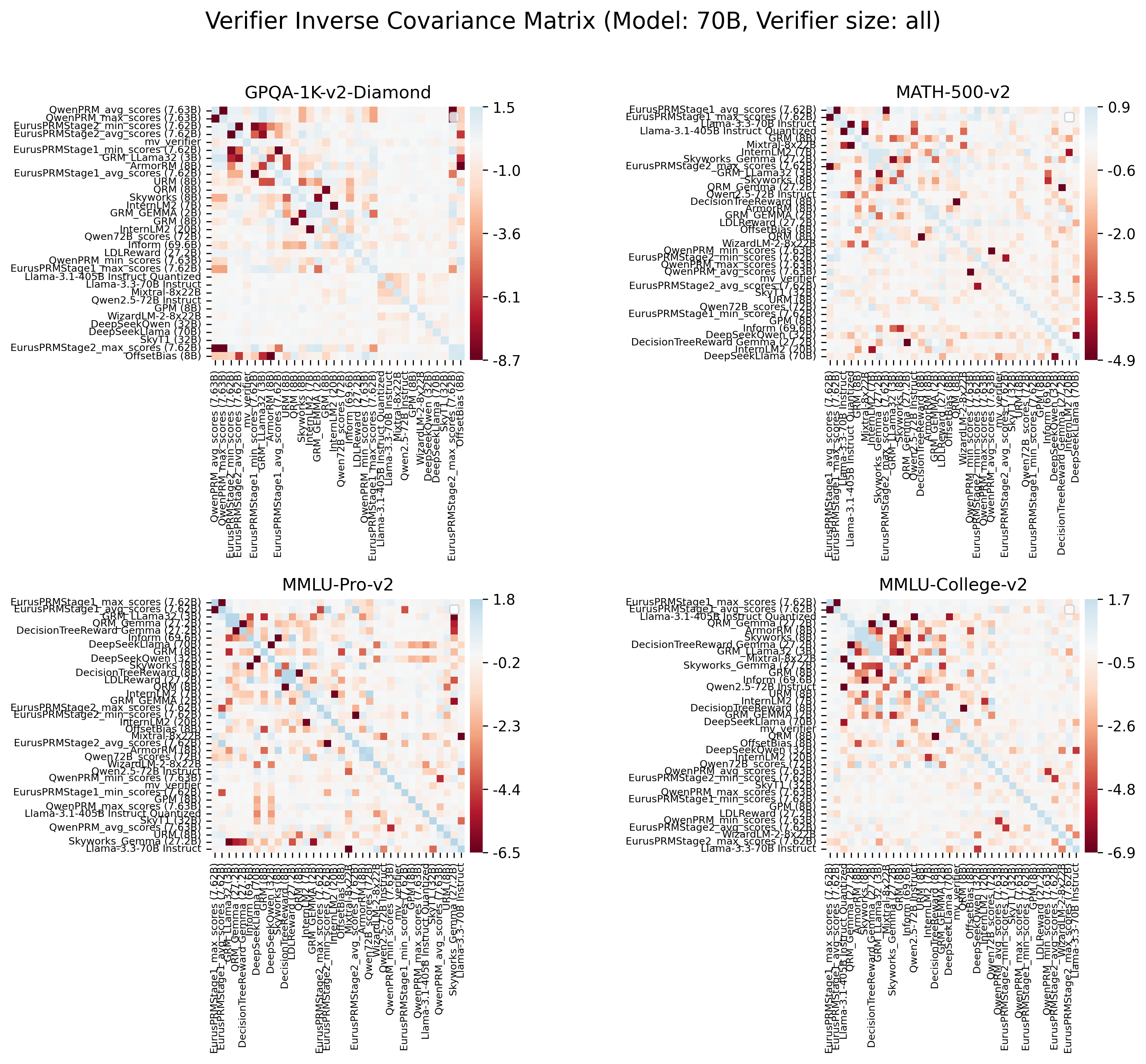}
   \caption{
   Inverse covariance matrix for each verifier across datasets given Llama-70B responses scores, after binarization
   }
   \label{fig:verifier_inv_cov_bin}
\end{figure}

\begin{figure}[H]
   \centering
   \includegraphics[width=\linewidth]{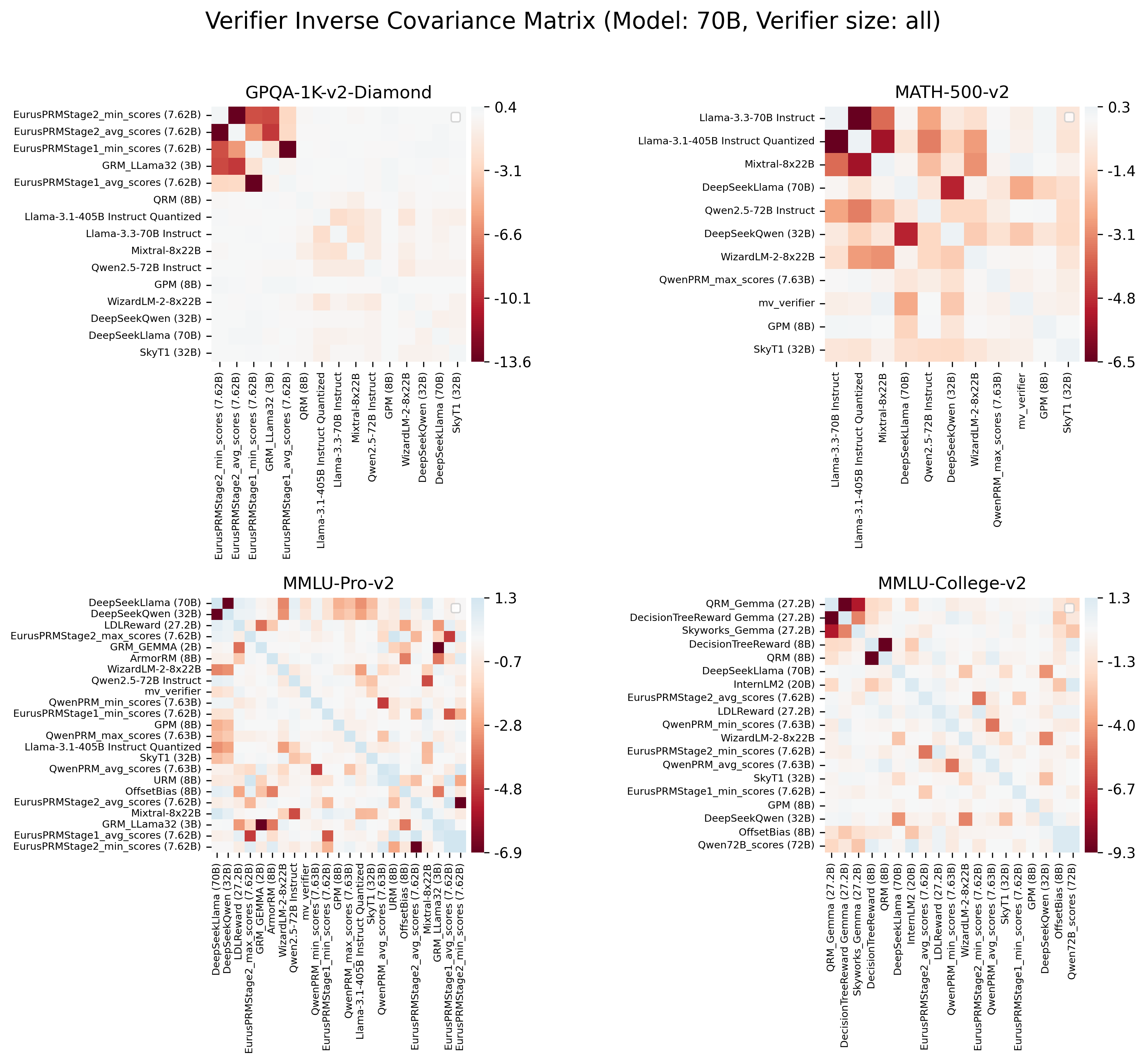}
   \caption{
   Inverse covariance matrix for each verifier across datasets given Llama-70B responses scores, after binarization and dropping.
   }
   \label{fig:verifier_inv_cov_bin_drop}
\end{figure}

%% file: tables_and_figures/weaver_dev_set_ablation.tex
\begin{table}
\small
\caption{\textbf{Ablation of Development Set used for \weaver{} Class Balance Estimation}: As our reward model threshold, we set it to the default of the $0.5$.
}
\centering
\setlength{\tabcolsep}{1.5pt}
\begin{tabular}{cccccccc}
\toprule
\multirow{2}{*}{\bf Approach} &
\multirow{2}{*}{\bf \begin{tabular}[c]{@{}c@{}}  Model\\Size\end{tabular}} &
\multirow{2}{*}{\bf \begin{tabular}[c]{@{}c@{}}  Dev Set\\Size\end{tabular}} &
\multicolumn{4}{c}{\bf Benchmarks} & \\
\cmidrule(lr){4-7}
& & & MATH500 & GPQA Diamond & MMLU & MMLU Pro & \textbf{Average} \\
\midrule
\weaver{} & 70B & \begin{tabular}[c]{@{}c@{}}  Naive Threshold \\(0.5 Threshold)\\\end{tabular} & 88.1\% & 52.0\% & 92.4\% & 83.5\% & 79.0\% \\ \midrule
\weaver{} & 70B & 1\% & 90.4\% & 67.1\% & 91.1\% & 87.0\% & 84.5\% \\
\weaver{} & 70B & 5\% & 92.4\% & 72.7\% & 93.5\% & 90.4\% & 87.5\% \\
\weaver{} & 70B & 20\% & 92.4\% & 72.7\% & 93.5\% & 90.4\% & 87.5\% \\
\weaver{} & 70B & 100\% & 92.4\% & 72.7\% & 93.5\% & 90.4\% & 87.5\% \\
\bottomrule
\end{tabular}
\label{tab:weaver_dev_set_ablation}
\end{table}

%% file: tables_and_figures/weaver_dropping_verifiers_ablation.tex
\begin{table}
\small
\caption{\textbf{Ablation of Verifier Selection Strategies for \weaver{}}: Comparison of different strategies for dropping faulty verifiers by dataset using each verifier's marginal probability.
}
\centering
\setlength{\tabcolsep}{1.5pt}
\begin{tabular}{ccccccccc}
\toprule
\multirow{2}{*}{\bf Approach} &
\multirow{2}{*}{\bf \begin{tabular}[c]{@{}c@{}}  Model\\Size\end{tabular}} &
\multirow{2}{*}{\bf \begin{tabular}[c]{@{}c@{}}  Verifier Selection\end{tabular}} &
\multicolumn{4}{c}{\bf Benchmarks} & \\
\cmidrule(lr){4-7}
& & & MATH500 & GPQA Diamond & MMLU & MMLU Pro & \textbf{Average} \\
\midrule
\weaver{} & 70B & No Dropped Verifiers & 90.4\% & 52.0\% & 91.1\% & 84.2\% & 79.4\% \\ \midrule
\weaver{} & 70B & \begin{tabular}[c]{@{}c@{}}Low Marginals Dropped\\(Mostly Negative Verifiers)\end{tabular} & \textbf{93.4\%} & 60.6\% & 91.7\% & \textbf{91.0\%} & 84.2\% \\ \midrule
\weaver{} & 70B & \begin{tabular}[c]{@{}c@{}}High Marginals Dropped\\(Mostly Positive Verifiers)\end{tabular} & 83.4\% & 69.7\% & 87.9\% & 78.4\% & 79.9\% \\ \midrule
\weaver{} & 70B & \begin{tabular}[c]{@{}c@{}}Extreme Marginals Dropped\\(Mostly Positive or Negative)\end{tabular} & 90.8\% & \textbf{72.7\%} & \textbf{92.4\%} & 85.0\% & \textbf{85.2\%} \\
\bottomrule
\end{tabular}
\label{tab:weaver_dropping_verifiers_ablation}
\end{table}

%% file: tables_and_figures/weaver_adaptive_threshold_ablation.tex
\begin{table}
\small
\caption{\textbf{Ablation of Adaptive Threshold Dev Set Size for \weaver{}}.
}
\centering
\setlength{\tabcolsep}{1.5pt}
\begin{tabular}{ccccccccc}
\toprule
\multirow{2}{*}{\bf Approach} &
\multirow{2}{*}{\bf \begin{tabular}[c]{@{}c@{}}  Model\\Size\end{tabular}} &
\multirow{2}{*}{\bf \begin{tabular}[c]{@{}c@{}}  Adaptive Threshold \\ Dev Set Size\end{tabular}} &
\multicolumn{4}{c}{\bf Benchmarks} & \\
\cmidrule(lr){4-7}
& & & MATH500 & GPQA Diamond & MMLU & MMLU Pro & \textbf{Average} \\
\midrule
\weaver{} & 70B & 0.01 & 92.4\% & 72.7\% & 93.5\% & 90.4\% & 87.5\% \\
\weaver{} & 70B & 0.05 & 92.4\% & 72.7\% & 93.5\% & 90.4\% & 87.5\% \\
\weaver{} & 70B & 0.2 & 92.4\% & 72.7\% & 93.5\% & 90.4\% & 87.5\% \\
\weaver{} & 70B & 1.0 & 92.4\% & 72.7\% & 93.5\% & 90.4\% & 87.5\% \\
\bottomrule
\end{tabular}
\label{tab:weaver_adaptive_threshold_ablation}
\end{table}

%% file: tables_and_figures/discrete_vs_continuous_for_LR.tex
\begin{table}[H]
\centering
\caption{\textbf{Performance Comparison Between Continuous and Discrete Logistic Regression}:  
For supervised fine-tuning on the verifier scores, the continuous model consistently outperforms the discrete variant across all datasets by avoiding the lossy conversion of floats to binary votes required for the discrete variant.}
\small
\setlength{\tabcolsep}{5pt}
\begin{tabular}{lccccc}
\toprule
\multicolumn{6}{c}{\textbf{Continuous vs.\ Discrete Logistic Regression Performance (\%)}}\\
\midrule
\multirow{2}{*}{\textbf{Method}} &
\multicolumn{5}{c}{\textbf{Dataset}}\\
\cmidrule(lr){2-6}
& MATH 500 & GPQA & MMLU College & MMLU Pro & BBH\\
\midrule
Discrete LR   & 93.1 & 74.3 & 87.5 & 87.1 & 90.1\\
Continuous LR & \textbf{97.2} & \textbf{78.1} & \textbf{90.4} & \textbf{92.0} & \textbf{96.5}\\ \midrule
\textbf{Improvement} & +4.1 & +3.8 & +2.9 & +4.9 & +6.4\\
\bottomrule
\end{tabular}
\label{tab:discrete_vs_continuous_for_LR}
\end{table}

%% file: tables_and_figures/8B_unique_answer_correlations.tex
\begin{table}[H]
  \centering
  \caption{\textbf{Number of Unique Extracted Answers vs. Positive}: Negative Sample Ratio per Query - Correlations for Llama 3.1 Instruct Models}
  \begin{minipage}{0.48\textwidth}
    \centering
    \small
    \setlength{\tabcolsep}{2pt}{
    \begin{tabular}{lccc}
      \toprule
      \multicolumn{4}{c}{\textbf{Llama 3.1 8B Instruct}} \\
      \multicolumn{4}{c}{\textbf{Correlation Metrics}} \\
      \midrule
      \multirow{2}{*}{\bf \begin{tabular}[c]{@{}c@{}}Dataset\end{tabular}} &
      \multicolumn{3}{c}{\bf Metric Type} \\
      \cmidrule(lr){2-4}
      & Pearson & Spearman & Kendall's Tau \\
      \midrule
      MATH 500 & -0.676 & -0.745 & -0.565 \\
      GPQA & -0.312 & -0.117 & -0.096 \\
      MMLU College & -0.595 & -0.700 & -0.591 \\
      MMLU Pro & -0.590 & -0.555 & -0.425 \\
      BBH & -0.365 & -0.386 & -0.300 \\
      \bottomrule
    \end{tabular}
    }
  \end{minipage}
  \hfill
  \begin{minipage}{0.48\textwidth}
    \centering
    \small
    \setlength{\tabcolsep}{2pt}{
    \begin{tabular}{lccc}
      \toprule
      \multicolumn{4}{c}{\textbf{Llama 3.1 70B Instruct}} \\
      \multicolumn{4}{c}{\textbf{Correlation Metrics}} \\
      \midrule
      \multirow{2}{*}{\bf \begin{tabular}[c]{@{}c@{}}Dataset\end{tabular}} &
      \multicolumn{3}{c}{\bf Metric Type} \\
      \cmidrule(lr){2-4}
      & Pearson & Spearman & Kendall's Tau \\
      \midrule
      MATH 500 & -0.631 & -0.842 & -0.709 \\
      GPQA & -0.148 & -0.093 & -0.089 \\
      MMLU College & -0.551 & -0.862 & -0.769 \\
      MMLU Pro & -0.446 & -0.693 & -0.585 \\
      BBH & -0.268 & -0.594 & -0.474 \\
      \bottomrule
    \end{tabular}
    }
  \end{minipage}
  \label{tab:unique_answer_correlations}
\end{table}

%% file: tables_and_figures/weaver_cluster_selection.tex
\begin{table}[H]
  \centering
  \small
  \caption{\textbf{Performance of \weaver{} with Different Clusters Counts}: Utilizes Llama 3.1 70B Instruct Generations with the verifier threshold optimized prior to clustering. We create the clusters based on query difficulty: \textit{the ratio of correct to incorrect generations for each queries}. We create cluster in evenly sized chunks from the distribution of each task.}
  \setlength{\tabcolsep}{5pt}{
  \begin{tabular}{lccccc}
    \toprule
    \multicolumn{6}{c}{\textbf{Clusters for \weaver{} Dataset}} \\
    \midrule
    \multirow{2}{*}{\bf \begin{tabular}[c]{@{}c@{}}Dataset\end{tabular}} &
    \multicolumn{5}{c}{\bf Cluster Count} \\
    \cmidrule(lr){2-6}
    & 1 & 2 & 3 & 4 & 5 \\
    \midrule
    MATH 500 & \textbf{93.4} & 87.6 & 83.8 & 82.8 & 81.2 \\
    GPQA & \textbf{66.4} & \textbf{66.4} & \textbf{66.4} & \textbf{66.4} & \textbf{66.4} \\
    MMLU College & \textbf{94.9} & 91.7 & 90.1 & 89.6 & 89.8 \\
    MMLU Pro & 88.4 & \textbf{90.2} & 87.1 & 84.6 & 79.8 \\
    \midrule
    \textbf{Average} & \textbf{85.8} & 84.0 & 81.9 & 80.9 & 79.3  \\
    \bottomrule
  \end{tabular}
  }
  \label{tab:weaver_cluster_selection}
\end{table}

%% file: tables_and_figures/optimizing_cluster_threshold.tex
\begin{table}[H]
  \centering
  \caption{\textbf{Optimizing Clusters and Adaptive Thresholds for \weaver{}}. We report selection performance across different evaluation strategies using weak supervision, with and without difficulty-based clustering and threshold tuning. Clustering is based on oracle query difficulty. Thresholds are selected via grid search from 0.05 to 0.95 in increments of 0.05.}
  \vspace{0.5em}
  \begin{minipage}{0.47\textwidth}
    \centering
    \scriptsize
    \vspace{0.3em}
    \setlength{\tabcolsep}{3pt}{
    \begin{tabular}{lccc}
      \toprule
      \multicolumn{4}{c}{\parbox{0.9\linewidth}{\centering\textbf{Performance Across Different Evaluation Methods\\ with Llama 3.1 70B Instruct}}} \\
      \midrule
      \multirow{2}{*}{\bf Dataset} &
      \multicolumn{3}{c}{\bf Clustering / Adaptive Threshold} \\
      \cmidrule(lr){2-4}
      & \begin{tabular}[c]{@{}c@{}}No Clusters /\\0.5 Threshold\end{tabular} & \begin{tabular}[c]{@{}c@{}}Best Found \\by Search\end{tabular} & Pass@K \\
      \midrule
      MATH 500 & 93.4\% & 95.2\% & 98.6\% \\
      GPQA Diamond & 72.4\% & 74.1\% & 81.0\% \\
      MMLU College & 94.9\% & 95.1\% & 96.0\% \\
      MMLU Pro & 90.2\% & 90.2\% & 92.0\% \\
      \midrule
      \textbf{Average} & 87.7\% & 88.7\% & 91.9\% \\
      \bottomrule
    \end{tabular}
    }
    \label{tab:optimizing_clusters_threshold_70b}
  \end{minipage}
  \hfill
  \begin{minipage}{0.47\textwidth}
    \centering
    \scriptsize
    \vspace{0.3em}
    \setlength{\tabcolsep}{3pt}{
    \begin{tabular}{lccc}
      \toprule
      \multicolumn{4}{c}{\parbox{0.9\linewidth}{\centering\textbf{Performance Across Different Evaluation Methods\\ with Llama 3.1 8B Instruct}}} \\
      \midrule
      \multirow{2}{*}{\bf Dataset} &
      \multicolumn{3}{c}{\bf Clustering / Adaptive Threshold} \\
      \cmidrule(lr){2-4}
      & \begin{tabular}[c]{@{}c@{}}No Clusters /\\0.5 Threshold\end{tabular} & \begin{tabular}[c]{@{}c@{}}Best Found \\by Search\end{tabular} & Pass@K \\
      \midrule
      MATH 500 & 80.0\% & 84.3\% & 99.2\% \\
      GPQA Diamond & 47.1\% & 52.7\% & 95.2\% \\
      MMLU College & 85.7\% & 89.9\% & 98.5\% \\
      MMLU Pro & 67.2\% & 72.3\% & 96.8\% \\
      \midrule
      \textbf{Average} & 70.0\% & 74.8\% & 97.4\% \\
      \bottomrule
    \end{tabular}
    }
    \label{tab:optimizing_clusters_threshold_8b}
  \end{minipage}
\end{table}

%% file: tables_and_figures/optimizing_reward_model_threshold_per_model.tex
\begin{table}[H]
  \centering
  \caption{\textbf{Optimizing Clusters and Per-Model Adaptive Thresholds for \weaver{}}. In this setting, each reward model receives its own optimized threshold (rather than a global threshold per cluster). Thresholds are selected via grid search from 0.05 to 0.95 in steps of 0.05. Clustering is still based on oracle query difficulty. This finer-grained tuning yields small improvements for 70B models and more substantial gains for 8B models, especially where verifier accuracy is highly variable.}
  \vspace{0.5em}
  \begin{minipage}{0.47\textwidth}
    \centering
    \scriptsize
    \vspace{0.3em}
    \setlength{\tabcolsep}{3pt}{
    \begin{tabular}{lccc}
      \toprule
      \multicolumn{4}{c}{\parbox{0.9\linewidth}{\centering{Performance Across Evaluation Methods\\ with Llama 3.1 70B Instruct (Per-Model Thresholds)}}} \\
      \midrule
      \multirow{2}{*}{\bf Dataset} &
      \multicolumn{3}{c}{\bf Clustering / Adaptive Threshold} \\
      \cmidrule(lr){2-4}
      & \begin{tabular}[c]{@{}c@{}}No Clusters /\\0.5 Threshold\end{tabular} & \begin{tabular}[c]{@{}c@{}}Best Found \\by Search\end{tabular} & Pass@K \\
      \midrule
      MATH 500 & 93.4\% & 95.2\% & 98.6\% \\
      GPQA Diamond & 72.4\% & {74.6\%} & 81.0\% \\
      MMLU College & 94.9\% & 95.1\% & 96.0\% \\
      MMLU Pro & 90.2\% & {91.0\%} & 92.0\% \\
      \midrule
      {Average} & 87.7\% & {89.0\%} & 91.9\% \\
      \bottomrule
    \end{tabular}
    }
    \label{tab:optimizing_clusters_threshold_70b_per_model}
  \end{minipage}
  \hfill
  \begin{minipage}{0.47\textwidth}
    \centering
    \scriptsize
    \vspace{0.3em}
    \setlength{\tabcolsep}{3pt}{
    \begin{tabular}{lccc}
      \toprule
      \multicolumn{4}{c}{\parbox{0.9\linewidth}{\centering{Performance Across Evaluation Methods\\ with Llama 3.1 8B Instruct (Per-Model Thresholds)}}} \\
      \midrule
      \multirow{2}{*}{\bf Dataset} &
      \multicolumn{3}{c}{\bf Clustering / Adaptive Threshold} \\
      \cmidrule(lr){2-4}
      & \begin{tabular}[c]{@{}c@{}}No Clusters /\\0.5 Threshold\end{tabular} & \begin{tabular}[c]{@{}c@{}}Best Found \\by Search\end{tabular} & Pass@K \\
      \midrule
      MATH 500 & 80.0\% & {86.5\%} & 99.2\% \\
      GPQA Diamond & 47.1\% & {55.5\%} & 95.2\% \\
      MMLU College & 85.7\% & {91.5\%} & 98.5\% \\
      MMLU Pro & 67.2\% & {74.5\%} & 96.8\% \\
      \midrule
      {Average} & 70.0\% & {77.0\%} & 97.4\% \\
      \bottomrule
    \end{tabular}
    }
    \label{tab:optimizing_clusters_threshold_8b_per_model}
  \end{minipage}
\end{table}

%% file: tables_and_figures/benchmarks_overview.tex
\begin{table}[H]
\centering
\small
\caption{\textbf{Benchmark Overview}: Evaluation configurations for AlpacaEval 2.0, Arena-Hard-Auto, AIMO, MATH500, GPQA, MMLU, MMLU Pro, and Big-Bench Hard (BBH). 
}
\setlength{\tabcolsep}{4pt}
\begin{tabular}{ccccc}
\toprule
\textbf{Benchmark} & \textbf{Dataset Size} & \textbf{Scoring Type} & \textbf{Metric} & \textbf{License} \\ \midrule
MATH500 & 500 & Ground Truth & Pass@1 & Apache 2.0 \\ \midrule
GPQA & 646 & Ground Truth & Pass@1 & CC BY 4.0 \\ \midrule
MMLU College & 719 & Ground Truth & Pass@1 & MIT \\ \midrule
MMLU Pro & 500 & Ground Truth & Pass@1 & MIT \\ \midrule
\end{tabular}
\label{tab:benchmarks_overview}
\end{table}

%% file: tables_and_figures/Growth_of_RMs.tex
\begin{figure}[H]
   \centering
   \includegraphics[width=\linewidth]{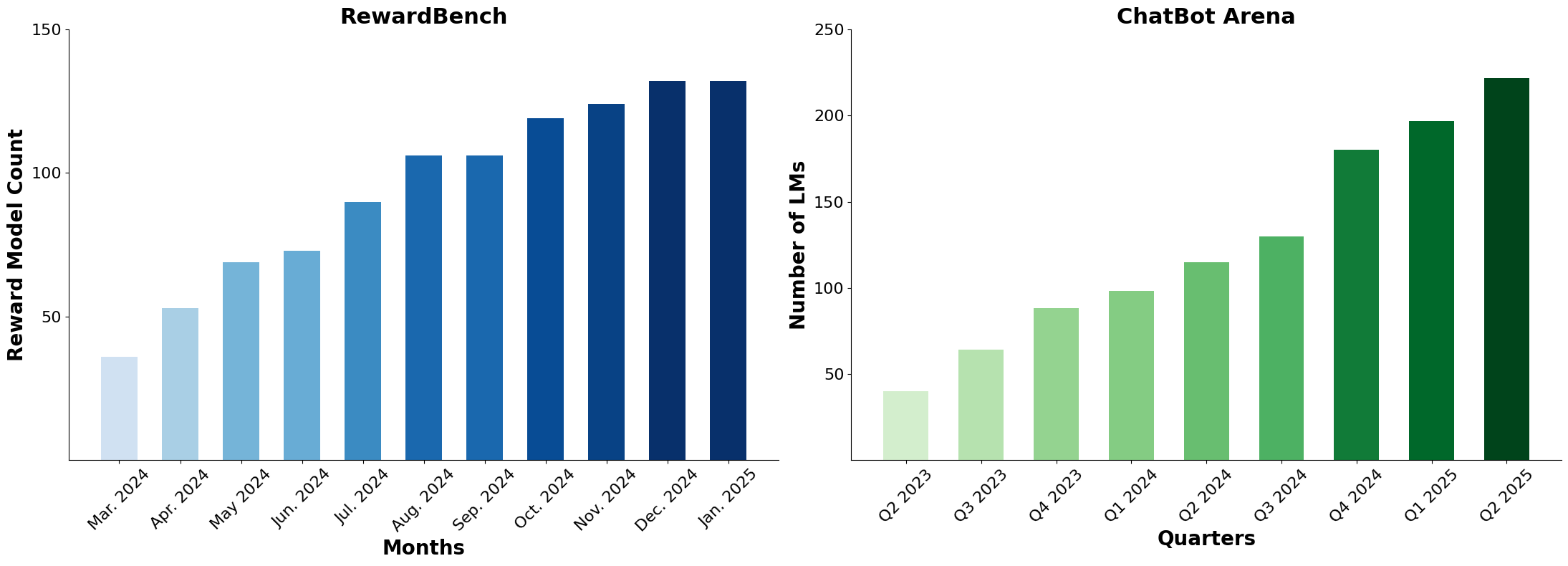}
   \caption{\textbf{Growth of Open-Source RMs and LMs}: As more and more RMs and LM judges become available, the need for better selection and utilization strategies for these models at test-time continues to grow.}
   \label{fig:growth_of_rms}
\end{figure}

%% file: tables_and_figures/8B_positive_negative_ratios.tex
\begin{table}[H]
  \centering
    \caption{\textbf{Distribution of Generation Accuracy for Llama 3.1 8B Instruct.} Each row shows the fraction of queries falling into deciles of correctness—i.e., the proportion of correct generations out of 100 samples per query. The final column reports the overall correct-to-incorrect (\textbf{C/I}) ratio for each dataset. These distributions highlight variation in query difficulty and motivate our clustering-by-difficulty approach in Appendix~\ref{app:clustering_exploration}.}
  \small
  \setlength{\tabcolsep}{1.8pt}{
  \begin{tabular}{lcccccccccc|c}
    \toprule
    \multicolumn{12}{c}{\textbf{Llama 3.1 8B Instruct}} \\
    \multicolumn{12}{c}{\textbf{Distribution of Correct/Incorrect Generations}} \\
    \midrule
    \multirow{2}{*}{\bf \begin{tabular}[c]{@{}c@{}}Dataset\end{tabular}} &
    \multicolumn{10}{c}{\bf Percentage of Total Dataset} & \multirow{2}{*}{\bf \begin{tabular}[c]{@{}c@{}}C/I\\ Ratio\end{tabular}} \\
    \cmidrule(lr){2-11}
    & 0.0-0.1 & 0.1-0.2 & 0.2-0.3 & 0.3-0.4 & 0.4-0.5 & 0.5-0.6 & 0.6-0.7 & 0.7-0.8 & 0.8-0.9 & 0.9-1.0 & \\
    \midrule
    AIMO & 72.2\% &	7.8\% &	3.3\% &	7.8\% &	2.2\% &	3.3\% &	2.2\% &	0.0\% &	1.1\% &	0.0\% &	10.3\% \\
    MATH 500 & 12.2\% & 11.4\% & 11.6\% & 8.2\% & 6.4\% & 8.0\% & 8.4\% & 6.60\% & 11.2\% & 15.0\% & 49.9\% \\
    GPQA & 28.5\% & 21.2\% & 16.1\% & 7.4\% & 7.3\% & 5.3\% & 3.7\% & 3.3\% & 2.9\% & 2.6\% & 28.3\% \\
    MMLU College & 7.8\% & 7.6\% & 7.1\% & 6.5\% & 6.7\% & 5.3\% & 7.2\% & 6.3\% & 8.6\% & 22.9\% & 64.1\% \\
    MMLU Pro & 22.4\% & 9.8\% & 10.6\% & 5.4\% & 6.4\% & 7.2\% & 4.0\% & 7.6\% & 7.6\% & 15.2\% & 46.6\% \\
    BBH & 3.2\% & 6.7\% & 10.3\% & 8.3\% & 10.3\% & 14.8\% & 11.6\% & 10.7\% & 10.4\% & 12.1\% & 56.9\% \\
    \bottomrule
  \end{tabular}
  }
  \label{tab:8B_positive_negative_ratios}
\end{table}

%% file: tables_and_figures/70B_positive_negative_ratios.tex
\begin{table}[H]
  \centering
  \small
    \caption{\textbf{Distribution of Generation Accuracy for Llama 3.1 70B Instruct.} Each row shows the fraction of queries falling into deciles of correctness—i.e., the proportion of correct generations out of 100 samples per query. The final column reports the overall correct-to-incorrect (\textbf{C/I}) ratio for each dataset. These distributions highlight variation in query difficulty and motivate our clustering-by-difficulty approach in Appendix~\ref{app:clustering_exploration}.}
  \setlength{\tabcolsep}{1.8pt}{
  \begin{tabular}{lcccccccccc|c}
    \toprule
    \multicolumn{12}{c}{\textbf{Llama 3.1 70B Instruct}} \\
    \multicolumn{12}{c}{\textbf{Distribution of Correct/Incorrect Generations}} \\
    \midrule
    \multirow{2}{*}{\bf \begin{tabular}[c]{@{}c@{}}Dataset\end{tabular}} &
    \multicolumn{10}{c}{\bf Percentage of Total Dataset} & \multirow{2}{*}{\bf \begin{tabular}[c]{@{}c@{}}C/I\\ Ratio\end{tabular}} \\
    \cmidrule(lr){2-11}
    & 0.0-0.1 & 0.1-0.2 & 0.2-0.3 & 0.3-0.4 & 0.4-0.5 & 0.5-0.6 & 0.6-0.7 & 0.7-0.8 & 0.8-0.9 & 0.9-1.0 & \\
    \midrule
    MATH 500 & 7.0\% & 4.2\% & 3.8\% & 2.0\% & 2.2\% & 3.8\% & 4.2\% & 4.0\% & 7.8\% & 61.0\% & 78.0\% \\
    GPQA & 36.8\% & 5.6\% & 5.1\% & 3.7\% & 6.2\% & 4.3\% & 4.5\% & 4.8\% & 8.0\% & 20.9\% & 42.9\% \\
    MMLU College & 8.1\% & 3.2\% & 1.8\% & 2.2\% & 1.5\% & 1.7\% & 1.8\% & 3.2\% & 2.4\% & 74.1\% & 82.6\% \\
    MMLU Pro & 16.4\% & 4.2\% & 1.8\% & 3.4\% & 3.0\% & 3.0\% & 3.2\% & 4.8\% & 6.0\% & 54.2\% & 69.9\% \\
    \bottomrule
  \end{tabular}
  }

  \label{tab:70B_positive_negative_ratios}
\end{table}

%% file: tables_and_figures/models_overview.tex
\begin{table}[H]
\centering
\caption{\textbf{Models Tested for \weaver{}.}} 
\scriptsize
\begin{tabular}{@{}c@{}ccccc@{}}
\toprule
& \textbf{Model} & \textbf{Source Code} & \textbf{\begin{tabular}[c]{@{}c@{}}Parameter \\ Count\end{tabular}} & \textbf{License} & \textbf{Loss Function} \\ \midrule
\multirow{23}*{\rotatebox[origin=c]{90}{\textbf{LM Judges}}}
& Llama-3.1-70B-Instruct & Open-Source & 70B & Llama 3.1 Community & Cross-Entropy Loss \\
& Llama-3.1-405B-Instruct & Open-Source & 405B & Llama 3.1 Community & Cross-Entropy Loss \\
& Llama-3.3-70B-Instruct & Open-Source & 70B & Llama 3.1 Community & Cross-Entropy Loss \\
& Meta-Llama-3.1-405B-Instruct-quantized.w8a16 & Open-Source & 405B & Llama 3.1 Community & Cross-Entropy Loss \\
& DeepSeek LLM 67B Chat & Open-Source & 67B & DeepSeek License & Cross-Entropy Loss \\
& DeepSeekLlama70B & Open-Source & 70B & DeepSeek License & Cross-Entropy Loss \\
& DeepSeekQwen32B & Open-Source & 32B & DeepSeek License & Cross-Entropy Loss \\
& DeepSeekLlama8B & Open-Source & 8B & DeepSeek License & Cross-Entropy Loss \\
& DeepSeekQwen7B & Open-Source & 7B & DeepSeek License & Cross-Entropy Loss \\
& Qwen2 72B Instruct & Open-Source & 72B & Tongyi Qianwen & Cross-Entropy Loss \\
& Qwen2.5-72B-Instruct & Open-Source & 72B & Tongyi Qianwen & Cross-Entropy Loss \\
& Qwen/Qwen2.5-72B-Instruct & Open-Source & 72B & Tongyi Qianwen & Cross-Entropy Loss \\
& QwQ-32B & Open-Source & 32B & Apache 2.0 & Cross-Entropy Loss \\
& Qwen1.5 110B Chat & Open-Source & 110B & Tongyi Qianwen & Cross-Entropy Loss \\
& Qwen1.5 72B Chat & Open-Source & 72B & Tongyi Qianwen & Cross-Entropy Loss \\
& Qwen-2.5-7B-Instruct & Open-Source & 7B & Tongyi Qianwen & Cross-Entropy Loss \\
& Qwen-2.5-Math-7B-Instruct & Open-Source & 7B & Tongyi Qianwen & Cross-Entropy Loss \\
& Mixtral 8x22B v0.1 & Open-Source & 176B & Apache 2.0 & Cross-Entropy Loss \\
& Mixtral-8x22B-Instruct-v0.1 & Open-Source & 176B & Apache 2.0 & Cross-Entropy Loss \\
& WizardLM 8x22B & Open-Source & 176B & Apache 2.0 & Cross-Entropy Loss \\
& WizardLM-2-8x22B & Open-Source & 176B & Apache 2.0 & Cross-Entropy Loss \\
& dbrx-instruct & Open-Source & 132B & Databricks Open Model & Cross-Entropy Loss \\
& SkyT1 & Open-Source & 32B & Apache 2.0 & Cross-Entropy Loss \\ %
\cmidrule{1-6}
\multirow{14}*{\rotatebox[origin=c]{90}{\textbf{RMs (8B and below)}}}
& GRM-Llama3-8B-rewardmodel-ft & Open-Source & 8B & MIT & Pairwise Ranking Loss \\
& GRM-Llama3.2-3B-rewardmodel-ft & Open-Source & 3B & Apache 2.0 & Pairwise Ranking Loss \\
& GRM-Gemma2-2B-rewardmodel-ft & Open-Source & 2B & Apache 2.0 & Pairwise Ranking Loss \\
& Skywork-Reward-Llama-3.1-8B-v0.2 & Open-Source & 8B & Skywork License & Pairwise Ranking Loss \\
& QRM-Llama3.1-8B-v2 & Open-Source & 8B & MIT & Quantile Regression Loss \\
& URM-LLaMa-3.1-8B & Open-Source & 8B & Skywork License & Uncertainty-Aware Loss \\
& GPM-Llama-3.1-8B & Open-Source & 8B & MIT & Pairwise Ranking Loss \\
& Llama-3-OffsetBias-RM-8B & Open-Source & 8B & Llama 3.1 Community & Pairwise Ranking Loss \\
& ArmoRM-Llama3-8B-v0.1 & Open-Source & 8B & Llama 3.1 Community & Pairwise Ranking Loss \\
& Qwen2.5-Math-PRM-7B & Open-Source & 7B & Tongyi Qianwen & Cross-Entropy Loss \\
& EurusPRM-Stage1 & Open-Source & 7B & Apache 2.0 & Cross-Entropy Loss \\
& EurusPRM-Stage2 & Open-Source & 7B & Apache 2.0 & Cross-Entropy Loss \\
& internlm2-7b-reward & Open-Source & 7B & Apache 2.0 & Pairwise Ranking Loss \\
& Decision-Tree-Reward-Llama-3.1-8B & Open-Source & 8B & Skywork License & Decision Tree Loss \\ \cmidrule{1-6}
\multirow{7}*{\rotatebox[origin=c]{90}{\textbf{RMs (27B--72B)}}}
& Skywork-Reward-Gemma-2-27B-v0.2 & Open-Source & 27B & Skywork License & Pairwise Ranking Loss \\
& QRM-Gemma-2-27B & Open-Source & 27B & MIT & Quantile Regression Loss \\
& INF-ORM-Llama3.1-70B & Open-Source & 70B & Custom License & Binary Cross-Entropy Loss \\
& Qwen2.5-Math-RM-72B & Open-Source & 72B & Tongyi Qianwen & Cross-Entropy Loss \\
& Qwen2.5-Math-PRM-72B & Open-Source & 72B & Tongyi Qianwen & Cross-Entropy Loss \\
& internlm2-20b-reward & Open-Source & 20B & Apache 2.0 & Pairwise Ranking Loss \\
& Decision-Tree-Reward-Gemma-2-27B & Open-Source & 27B & Skywork License & Pairwise Ranking Loss \\
\bottomrule                                                                 
\end{tabular}
\label{tab:models_overview}
\end{table}

%% file: tables_and_figures/verifier_accuracies_ranges.tex
\begin{table}
\small
\caption{\textbf{\weaver{} Verifier Accuracies and Score Correlations.} 
We report the range of individual verifier accuracies and the average pairwise Pearson correlation between verifier scores. 
Each verifier's outputs are flattened across all query–candidate pairs, and correlations are computed across all \( \binom{m}{2} \) verifier pairs.
Lower correlation indicates greater diversity in how verifiers score responses, which supports the effectiveness of ensembling under \weaver{}.}
\centering
\setlength{\tabcolsep}{1.5pt}
\begin{tabular}{ccccccccc}
\toprule
\multirow{2}{*}{\bf Metric} &
\multirow{2}{*}{\bf \begin{tabular}[c]{@{}c@{}}  Model\\Size\end{tabular}} &
\multicolumn{4}{c}{\bf Benchmarks} \\
\cmidrule(lr){3-6}
& & MATH500 & GPQA & MMLU Pro & \textbf{Average} \\
\midrule
Verifier Accuracy Range & 8B & 34.2\% & 40.7\% & 36.4\% & \textbf{37.1\%} \\
Avg. Score Correlation  & 8B & 0.0253 & 0.0349 & 0.0312 & \textbf{0.0305} \\ \midrule
Verifier Accuracy Range & 70B & 27.4\% & 29.0\% & 31.6\% & \textbf{29.3\%} \\
Avg. Score Correlation  & 70B & 0.0211 & 0.0372 & 0.0240 & \textbf{0.0274} \\
\bottomrule
\end{tabular}
\label{tab:verifier_accuracies_ranges}
\end{table}

%% file: tables_and_figures/LR_and_NB_across_datasizes.tex
\begin{table}
\small
\centering
\caption{\textbf{Logistic Regression and Naive Bayes Performances across Datasets and Dev Set Sizes}}
\setlength{\tabcolsep}{1.5pt}
\begin{tabular}{cccccccc}
\toprule
\multirow{2}{*}{\bf Dataset} &
\multirow{2}{*}{\bf Approach} &
\multirow{2}{*}{\bf \begin{tabular}[c]{@{}c@{}}  Model\\Size\end{tabular}} &
\multicolumn{5}{c}{\bf Dev Set Size} \\
\cmidrule(lr){4-8}
& & & 0.01 & 0.05 & 0.2 & 0.5 & 1.0 \\
\midrule
\multirow{2}{*}{MATH-500} & Logistic Regression & 70B & 70.5\% & 74.7\% & 81.4\% & 93.1\% & 97.2\% \\
& Naive Bayes & 70B & 67.4\% & 78.1\% & 85.0\% & 89.2\% & 92.2\% \\ 
\midrule
\multirow{2}{*}{GPQA Diamond} & Logistic Regression & 70B & 55.9\% & 59.4\% & 69.8\% & 71.4\% & 72.9\% \\
& Naive Bayes & 70B & 47.2\% & 49.2\% & 57.6\% & 62.1\% & 64.3\% \\ 
\midrule
\multirow{2}{*}{MMLU Pro} & Logistic Regression & 70B & 72.1\% & 81.0\% & 84.6\% & 86.0\% & 92.0\% \\
& Naive Bayes & 70B & 60.2\% & 73.1\% & 73.1\% & 78.6\% & 78.6\% \\ 
\bottomrule
\end{tabular}
\label{tab:LR_and_NB_across_datasizes}
\end{table}

%% file: tables_and_figures/scaling_trends.tex
\begin{figure}[htbp]
\centering
\includegraphics[height=0.45\textheight,width=\textwidth]{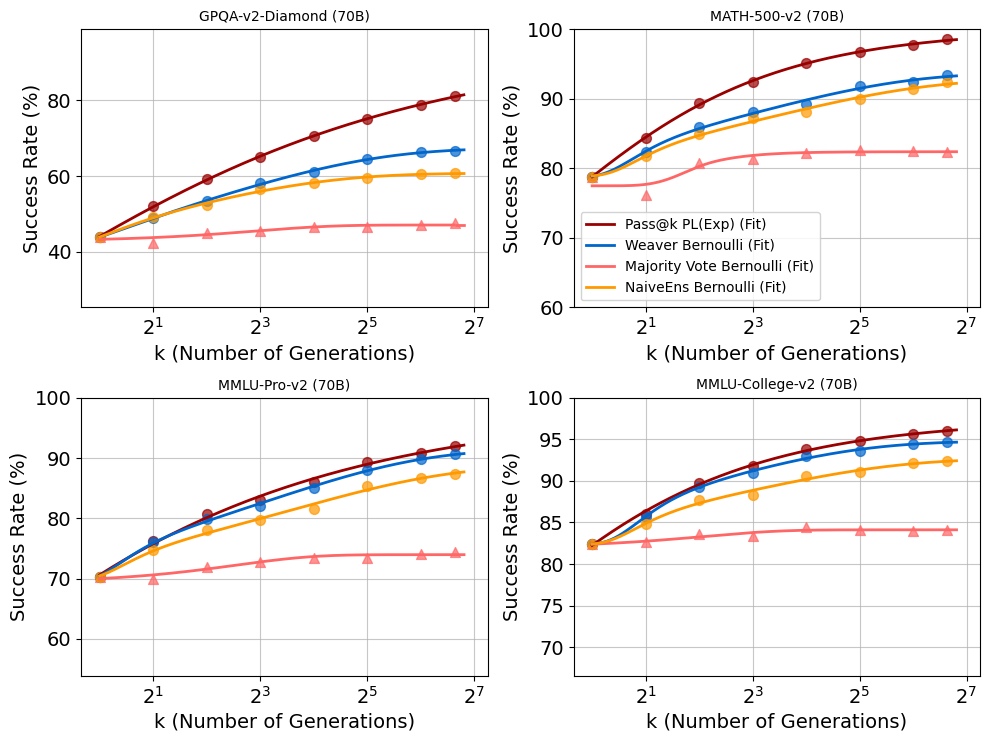}
\caption{\textbf{\weaver{} Scaling trend fit for 70B models}
}
\label{fig:power_law_fit70B}
\end{figure}

\begin{figure}[h]
\centering
\includegraphics[height=0.45\textheight,width=\textwidth]{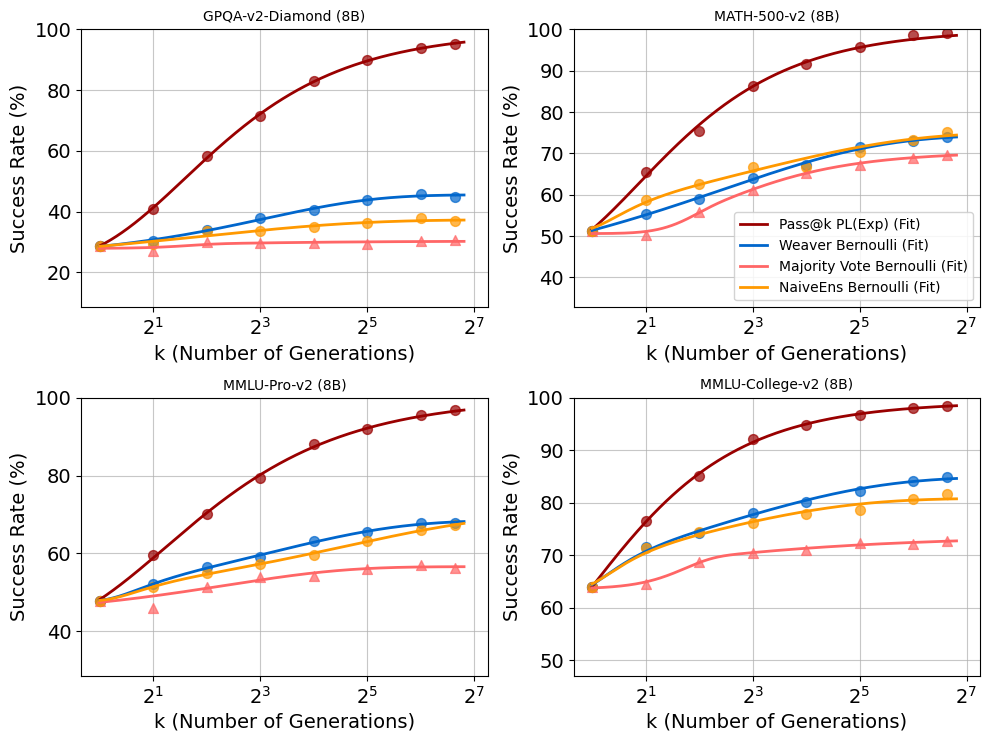}
\caption{\textbf{\weaver{} Scaling trend fit for 8B models.}
}
\label{fig:power_law_fit8B}
\end{figure}

\begin{table}[h]
\small
\centering
\caption{Fitted parameters for Scaling Trends in \cref{fig:power_law_fit70B} and  \cref{fig:power_law_fit8B}.}
\label{tab:power_law_fit}
\resizebox{\textwidth}{!}{%
\begin{tabular}{lllrrrrllrrr}
\toprule
Dataset & Approach & Equation & floor & ceil & $\zeta$ & $\alpha$ & $\pi$ & $\gamma$ & R2 fit & MSE fit & $\delta$ \\
\midrule
GPQA-v2-Diamond (70B) & Pass@K & $y = \mathrm{floor} + (\mathrm{ceil} - \mathrm{floor}) \cdot \exp(-\zeta \cdot K^{-\alpha})$ & 0.0000 & 0.9429 & 0.7603 & 0.3475 & X & X & 0.9999 & 0.0000 & 0.5000 \\
GPQA-v2-Diamond (70B) & Weaver & $y = \mathrm{floor} + (\mathrm{ceil} - \mathrm{floor}) \cdot \exp(-\zeta \cdot K^{-\alpha}) (1 - (1-\pi)^{K^{\gamma}})$ & 0.3958 & 0.6728 & 0.7320 & 1.5865 & 0.3250 & 0.5053 & 0.9994 & 0.0000 & 0.1000 \\
GPQA-v2-Diamond (70B) & Majority1@K & $y = \mathrm{floor} + (\mathrm{ceil} - \mathrm{floor}) \cdot \exp(-\zeta \cdot K^{-\alpha}) (1 - (1-\pi)^{K^{\gamma}})$ & 0.4283 & 0.4710 & 0.0499 & 1.0000 & 0.1217 & 1.0091 & 0.8634 & 0.0000 & 0.0100 \\
GPQA-v2-Diamond (70B) & Naive Ensemble & $y = \mathrm{floor} + (\mathrm{ceil} - \mathrm{floor}) \cdot \exp(-\zeta \cdot K^{-\alpha}) (1 - (1-\pi)^{K^{\gamma}})$ & 0.3921 & 0.6071 & 0.6553 & 1.9147 & 0.4224 & 0.5000 & 0.9975 & 0.0000 & 0.2500 \\
MATH-500-v2 (70B) & Pass@K & $y = \mathrm{floor} + (\mathrm{ceil} - \mathrm{floor}) \cdot \exp(-\zeta \cdot K^{-\alpha})$ & 0.6262 & 1.0000 & 0.8394 & 0.6427 & X & X & 0.9994 & 0.0000 & 0.2500 \\
MATH-500-v2 (70B) & Weaver & $y = \mathrm{floor} + (\mathrm{ceil} - \mathrm{floor}) \cdot \exp(-\zeta \cdot K^{-\alpha}) (1 - (1-\pi)^{K^{\gamma}})$ & 0.7870 & 0.9371 & 3.3908 & 3.0000 & 0.2869 & 0.5000 & 0.9958 & 0.0000 & 0.1000 \\
MATH-500-v2 (70B) & Majority1@K & $y = \mathrm{floor} + (\mathrm{ceil} - \mathrm{floor}) \cdot \exp(-\zeta \cdot K^{-\alpha}) (1 - (1-\pi)^{K^{\gamma}})$ & 0.7747 & 0.8238 & 10.0000 & 2.1433 & 0.0885 & 2.4951 & 0.8655 & 0.0001 & 0.1000 \\
MATH-500-v2 (70B) & Naive Ensemble & $y = \mathrm{floor} + (\mathrm{ceil} - \mathrm{floor}) \cdot \exp(-\zeta \cdot K^{-\alpha}) (1 - (1-\pi)^{K^{\gamma}})$ & 0.7883 & 0.9282 & 4.0033 & 3.0000 & 0.2573 & 0.5000 & 0.9961 & 0.0000 & 0.0100 \\
MMLU-Pro-v2 (70B) & Pass@K & $y = \mathrm{floor} + (\mathrm{ceil} - \mathrm{floor}) \cdot \exp(-\zeta \cdot K^{-\alpha})$ & 0.0000 & 0.9828 & 0.3303 & 0.3465 & X & X & 0.9967 & 0.0000 & 0.2500 \\
MMLU-Pro-v2 (70B) & Weaver & $y = \mathrm{floor} + (\mathrm{ceil} - \mathrm{floor}) \cdot \exp(-\zeta \cdot K^{-\alpha}) (1 - (1-\pi)^{K^{\gamma}})$ & 0.6912 & 0.9148 & 1.5284 & 3.0000 & 0.2764 & 0.5000 & 0.9987 & 0.0000 & 0.1000 \\
MMLU-Pro-v2 (70B) & Majority1@K & $y = \mathrm{floor} + (\mathrm{ceil} - \mathrm{floor}) \cdot \exp(-\zeta \cdot K^{-\alpha}) (1 - (1-\pi)^{K^{\gamma}})$ & 0.6933 & 0.7399 & 0.0498 & 1.0001 & 0.1531 & 1.0123 & 0.9451 & 0.0000 & 0.0100 \\
MMLU-Pro-v2 (70B) & Naive Ensemble & $y = \mathrm{floor} + (\mathrm{ceil} - \mathrm{floor}) \cdot \exp(-\zeta \cdot K^{-\alpha}) (1 - (1-\pi)^{K^{\gamma}})$ & 0.6969 & 0.8874 & 1.7834 & 3.0000 & 0.2403 & 0.5000 & 0.9944 & 0.0000 & 0.2500 \\
MMLU-College-v2 (70B) & Pass@K & $y = \mathrm{floor} + (\mathrm{ceil} - \mathrm{floor}) \cdot \exp(-\zeta \cdot K^{-\alpha})$ & 0.5924 & 0.9744 & 0.5071 & 0.5682 & X & X & 0.9982 & 0.0000 & 0.0100 \\
MMLU-College-v2 (70B) & Weaver & $y = \mathrm{floor} + (\mathrm{ceil} - \mathrm{floor}) \cdot \exp(-\zeta \cdot K^{-\alpha}) (1 - (1-\pi)^{K^{\gamma}})$ & 0.8234 & 0.9477 & 4.4129 & 3.0000 & 0.3622 & 0.5000 & 0.9987 & 0.0000 & 0.0100 \\
MMLU-College-v2 (70B) & Majority1@K & $y = \mathrm{floor} + (\mathrm{ceil} - \mathrm{floor}) \cdot \exp(-\zeta \cdot K^{-\alpha}) (1 - (1-\pi)^{K^{\gamma}})$ & 0.8197 & 0.8412 & 0.0498 & 1.0001 & 0.2057 & 1.0173 & 0.8912 & 0.0000 & 0.0100 \\
MMLU-College-v2 (70B) & Naive Ensemble & $y = \mathrm{floor} + (\mathrm{ceil} - \mathrm{floor}) \cdot \exp(-\zeta \cdot K^{-\alpha}) (1 - (1-\pi)^{K^{\gamma}}) $ & 0.8235 & 0.9266 & 3.8766 & 3.0000 & 0.3012 & 0.5000 & 0.9925 & 0.0000 & 0.0500 \\
GPQA-v2-Diamond (8B) & Pass@K & $y = \mathrm{floor} + (\mathrm{ceil} - \mathrm{floor}) \cdot \exp(-\zeta \cdot K^{-\alpha})$ & 0.2262 & 0.9926 & 2.5454 & 0.8474 & X & X & 0.9996 & 0.0000 & 0.1000 \\
GPQA-v2-Diamond (8B) & Weaver & $y = \mathrm{floor} + (\mathrm{ceil} - \mathrm{floor}) \cdot \exp(-\zeta \cdot K^{-\alpha}) (1 - (1-\pi)^{K^{\gamma}})$ & 0.2463 & 0.4549 & 0.0534 & 1.0020 & 0.1953 & 0.7089 & 0.9948 & 0.0000 & 0.0500 \\
GPQA-v2-Diamond (8B) & Majority1@K & $y = \mathrm{floor} + (\mathrm{ceil} - \mathrm{floor}) \cdot \exp(-\zeta \cdot K^{-\alpha}) (1 - (1-\pi)^{K^{\gamma}})$ & 0.2783 & 0.3029 & 1.2431 & 0.6756 & 0.0630 & 2.5000 & 0.5929 & 0.0000 & 0.0500 \\
GPQA-v2-Diamond (8B) & Naive Ensemble & $y = \mathrm{floor} + (\mathrm{ceil} - \mathrm{floor}) \cdot \exp(-\zeta \cdot K^{-\alpha}) (1 - (1-\pi)^{K^{\gamma}})$ & 0.2359 & 0.3727 & 0.0701 & 1.0016 & 0.3871 & 0.5000 & 0.9408 & 0.0001 & 0.0100 \\
MATH-500-v2 (8B) & Pass@K & $y = \mathrm{floor} + (\mathrm{ceil} - \mathrm{floor}) \cdot \exp(-\zeta \cdot K^{-\alpha})$ & 0.4099 & 1.0000 & 1.7182 & 0.8949 & X & X & 0.9976 & 0.0001 & 0.0100 \\
MATH-500-v2 (8B) & Weaver & $y = \mathrm{floor} + (\mathrm{ceil} - \mathrm{floor}) \cdot \exp(-\zeta \cdot K^{-\alpha}) (1 - (1-\pi)^{K^{\gamma}})$ & 0.4110 & 0.7440 & 0.0785 & 1.0033 & 0.3333 & 0.5036 & 0.9984 & 0.0000 & 0.0100 \\
MATH-500-v2 (8B) & Majority1@K & $y = \mathrm{floor} + (\mathrm{ceil} - \mathrm{floor}) \cdot \exp(-\zeta \cdot K^{-\alpha}) (1 - (1-\pi)^{K^{\gamma}})$ & 0.5058 & 0.7038 & 5.1187 & 1.0175 & 0.0666 & 2.5000 & 0.9964 & 0.0000 & 0.0500 \\
MATH-500-v2 (8B) & Naive Ensemble & $y = \mathrm{floor} + (\mathrm{ceil} - \mathrm{floor}) \cdot \exp(-\zeta \cdot K^{-\alpha}) (1 - (1-\pi)^{K^{\gamma}})$ & 0.5071 & 0.7508 & 1.8068 & 3.0000 & 0.2890 & 0.5000 & 0.9796 & 0.0001 & 0.0100 \\
MMLU-Pro-v2 (8B) & Pass@K & $y = \mathrm{floor} + (\mathrm{ceil} - \mathrm{floor}) \cdot \exp(-\zeta \cdot K^{-\alpha})$ & 0.3906 & 1.0000 & 1.9045 & 0.7590 & X & X & 0.9991 & 0.0000 & 0.0100 \\
MMLU-Pro-v2 (8B) & Weaver & $y = \mathrm{floor} + (\mathrm{ceil} - \mathrm{floor}) \cdot \exp(-\zeta \cdot K^{-\alpha}) (1 - (1-\pi)^{K^{\gamma}})$ & 0.4764 & 0.6846 & 2.7876 & 3.0000 & 0.2136 & 0.6141 & 0.9985 & 0.0000 & 0.2500 \\
MMLU-Pro-v2 (8B) & Majority1@K & $y = \mathrm{floor} + (\mathrm{ceil} - \mathrm{floor}) \cdot \exp(-\zeta \cdot K^{-\alpha}) (1 - (1-\pi)^{K^{\gamma}})$ & 0.4439 & 0.5662 & 0.1136 & 0.9916 & 0.2787 & 0.6659 & 0.9084 & 0.0001 & 0.0100 \\
MMLU-Pro-v2 (8B) & Naive Ensemble & $y = \mathrm{floor} + (\mathrm{ceil} - \mathrm{floor}) \cdot \exp(-\zeta \cdot K^{-\alpha}) (1 - (1-\pi)^{K^{\gamma}})$ & 0.4771 & 0.7025 & 2.8863 & 3.0000 & 0.1728 & 0.5181 & 0.9986 & 0.0000 & 0.1000 \\
MMLU-College-v2 (8B) & Pass@K & $y = \mathrm{floor} + (\mathrm{ceil} - \mathrm{floor}) \cdot \exp(-\zeta \cdot K^{-\alpha})$ & 0.4316 & 0.9924 & 0.9887 & 0.9123 & X & X & 0.9994 & 0.0000 & 0.5000 \\
MMLU-College-v2 (8B) & Weaver & $y = \mathrm{floor} + (\mathrm{ceil} - \mathrm{floor}) \cdot \exp(-\zeta \cdot K^{-\alpha}) (1 - (1-\pi)^{K^{\gamma}})$ & 0.6226 & 0.8494 & 1.2646 & 3.0000 & 0.3346 & 0.5000 & 0.9958 & 0.0000 & 0.2500 \\
MMLU-College-v2 (8B) & Majority1@K & $y = \mathrm{floor} + (\mathrm{ceil} - \mathrm{floor}) \cdot \exp(-\zeta \cdot K^{-\alpha}) (1 - (1-\pi)^{K^{\gamma}})$ & 0.6359 & 0.7368 & 1.0929 & 0.5130 & 0.0576 & 2.5000 & 0.9949 & 0.0000 & 0.0500 \\
MMLU-College-v2 (8B) & Naive Ensemble & $y = \mathrm{floor} + (\mathrm{ceil} - \mathrm{floor}) \cdot \exp(-\zeta \cdot K^{-\alpha}) (1 - (1-\pi)^{K^{\gamma}})$ & 0.6283 & 0.8085 & 1.4279 & 3.0000 & 0.3914 & 0.5000 & 0.9845 & 0.0000 & 0.0100 \\
\bottomrule
\end{tabular}
}
\end{table}

\begin{figure}[H]
\centering
\includegraphics[height=0.45\textheight,width=\linewidth]{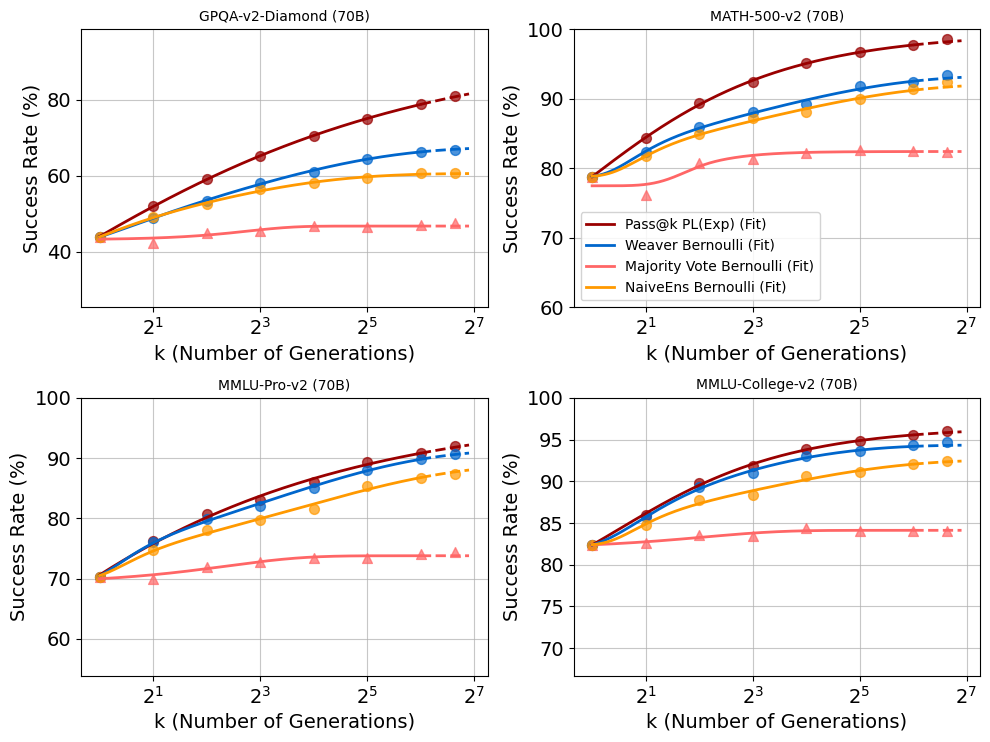}
\caption{\textbf{\weaver{} Scaling trend predicted for 70B models}}
\label{fig:power_law_pred70B}
\end{figure}

\begin{figure}[H]
\centering
\includegraphics[height=0.45\textheight,width=\linewidth]{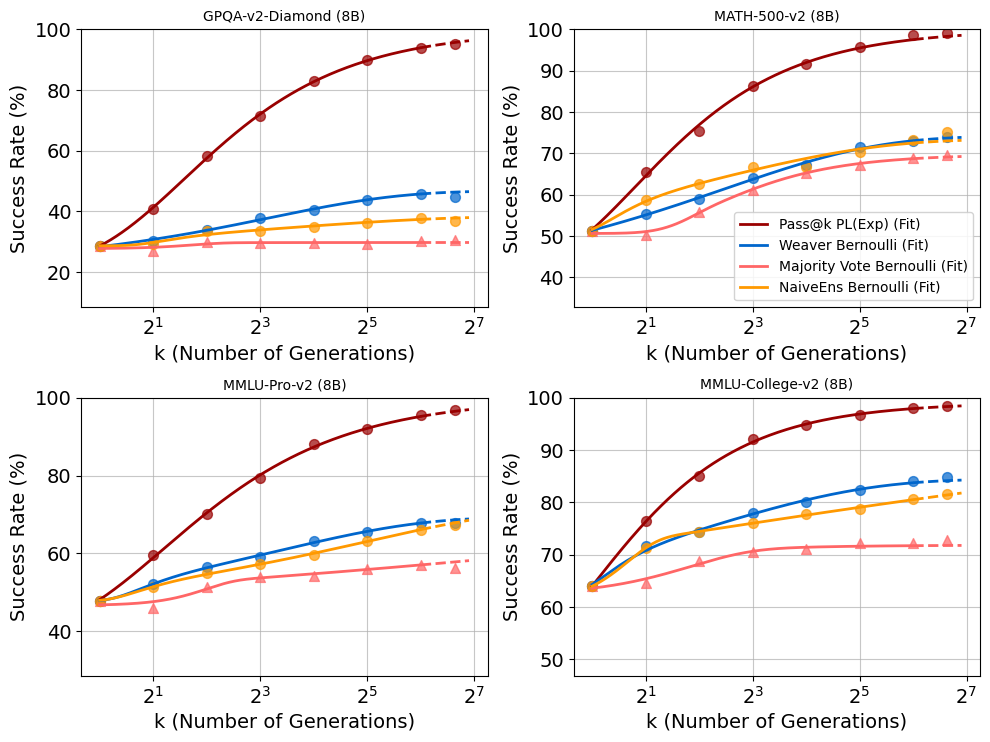}
\caption{\textbf{\weaver{} Scaling trend predicted for 8B models}}
\label{fig:power_law_pred8B}
\end{figure}

\begin{table}[h]
\small
\centering
\caption{Fitted parameters for Scaling Trends with 90\% of data in \cref{fig:power_law_pred70B} and \cref{fig:power_law_pred8B}.}
\label{tab:power_law_pred}
\resizebox{\textwidth}{!}{%
\begin{tabular}{lllrrrrllrrrr}
\toprule
Dataset & Approach & Equation & floor & ceil & $\zeta$ & $\alpha$ & $\pi$ & $\gamma$ & R2 fit & MSE fit & MSE pred & $\delta$ \\
\midrule
GPQA-v2-Diamond (70B) & Pass@K & $y = \mathrm{floor} + (\mathrm{ceil} - \mathrm{floor}) \cdot \exp(-\zeta \cdot K^{-\alpha})$ & 0.0000 & 0.9357 & 0.7534 & 0.3537 & X & X & 0.9999 & 0.0000 & 0.0000 & 0.1000 \\
GPQA-v2-Diamond (70B) & Weaver & $y = \mathrm{floor} + (\mathrm{ceil} - \mathrm{floor}) \cdot \exp(-\zeta \cdot K^{-\alpha}) (1 - (1-\pi)^{K^{\gamma}}) $ & 0.4050 & 0.6756 & 0.9195 & 1.6227 & 0.3163 & 0.5000 & 0.9993 & 0.0000 & 0.0000 & 0.0500 \\
GPQA-v2-Diamond (70B) & Majority1@K & $y = \mathrm{floor} + (\mathrm{ceil} - \mathrm{floor}) \cdot \exp(-\zeta \cdot K^{-\alpha}) (1 - (1-\pi)^{K^{\gamma}}) $ & 0.4320 & 0.4678 & 0.0707 & 0.9864 & 0.0414 & 1.6541 & 0.8601 & 0.0000 & 0.0001 & 0.0100 \\
GPQA-v2-Diamond (70B) & Naive Ensemble & $y = \mathrm{floor} + (\mathrm{ceil} - \mathrm{floor}) \cdot \exp(-\zeta \cdot K^{-\alpha}) (1 - (1-\pi)^{K^{\gamma}}) $ & 0.3809 & 0.6061 & 0.5211 & 1.9075 & 0.4360 & 0.5000 & 0.9972 & 0.0000 & 0.0000 & 0.0500 \\
MATH-500-v2 (70B) & Pass@K & $y = \mathrm{floor} + (\mathrm{ceil} - \mathrm{floor}) \cdot \exp(-\zeta \cdot K^{-\alpha})$ & 0.6639 & 0.9952 & 0.9836 & 0.6936 & X & X & 0.9995 & 0.0000 & 0.0000 & 0.0100 \\
MATH-500-v2 (70B) & Weaver & $y = \mathrm{floor} + (\mathrm{ceil} - \mathrm{floor}) \cdot \exp(-\zeta \cdot K^{-\alpha}) (1 - (1-\pi)^{K^{\gamma}}) $ & 0.7869 & 0.9339 & 3.5000 & 3.0000 & 0.2985 & 0.5000 & 0.9954 & 0.0000 & 0.0000 & 0.0500 \\
MATH-500-v2 (70B) & Majority1@K & $y = \mathrm{floor} + (\mathrm{ceil} - \mathrm{floor}) \cdot \exp(-\zeta \cdot K^{-\alpha}) (1 - (1-\pi)^{K^{\gamma}}) $ & 0.7747 & 0.8241 & 10.0000 & 2.1272 & 0.0888 & 2.4999 & 0.8560 & 0.0001 & 0.0000 & 0.0500 \\
MATH-500-v2 (70B) & Naive Ensemble & $y = \mathrm{floor} + (\mathrm{ceil} - \mathrm{floor}) \cdot \exp(-\zeta \cdot K^{-\alpha}) (1 - (1-\pi)^{K^{\gamma}}) $ & 0.7879 & 0.9221 & 4.2081 & 3.0000 & 0.2789 & 0.5000 & 0.9966 & 0.0000 & 0.0000 & 0.1000 \\
MMLU-Pro-v2 (70B) & Pass@K & $y = \mathrm{floor} + (\mathrm{ceil} - \mathrm{floor}) \cdot \exp(-\zeta \cdot K^{-\alpha})$ & 0.0000 & 0.9785 & 0.3263 & 0.3543 & X & X & 0.9959 & 0.0000 & 0.0000 & 0.1000 \\
MMLU-Pro-v2 (70B) & Weaver & $y = \mathrm{floor} + (\mathrm{ceil} - \mathrm{floor}) \cdot \exp(-\zeta \cdot K^{-\alpha}) (1 - (1-\pi)^{K^{\gamma}}) $ & 0.6912 & 0.9148 & 1.5277 & 3.0000 & 0.2765 & 0.5000 & 0.9984 & 0.0000 & 0.0000 & 0.2500 \\
MMLU-Pro-v2 (70B) & Majority1@K & $y = \mathrm{floor} + (\mathrm{ceil} - \mathrm{floor}) \cdot \exp(-\zeta \cdot K^{-\alpha}) (1 - (1-\pi)^{K^{\gamma}}) $ & 0.6923 & 0.7379 & 0.0495 & 1.0001 & 0.1711 & 1.0148 & 0.9481 & 0.0000 & 0.0000 & 0.0500 \\
MMLU-Pro-v2 (70B) & Naive Ensemble & $y = \mathrm{floor} + (\mathrm{ceil} - \mathrm{floor}) \cdot \exp(-\zeta \cdot K^{-\alpha}) (1 - (1-\pi)^{K^{\gamma}}) $ & 0.6979 & 0.8907 & 1.8452 & 3.0000 & 0.2327 & 0.5000 & 0.9931 & 0.0000 & 0.0000 & 0.1000 \\
MMLU-College-v2 (70B) & Pass@K & $y = \mathrm{floor} + (\mathrm{ceil} - \mathrm{floor}) \cdot \exp(-\zeta \cdot K^{-\alpha})$ & 0.7723 & 0.9655 & 1.3237 & 0.7746 & X & X & 0.9987 & 0.0000 & 0.0000 & 0.1000 \\
MMLU-College-v2 (70B) & Weaver & $y = \mathrm{floor} + (\mathrm{ceil} - \mathrm{floor}) \cdot \exp(-\zeta \cdot K^{-\alpha}) (1 - (1-\pi)^{K^{\gamma}}) $ & 0.8184 & 0.9436 & 2.2897 & 2.0761 & 0.4148 & 0.5000 & 0.9979 & 0.0000 & 0.0000 & 0.2500 \\
MMLU-College-v2 (70B) & Majority1@K & $y = \mathrm{floor} + (\mathrm{ceil} - \mathrm{floor}) \cdot \exp(-\zeta \cdot K^{-\alpha}) (1 - (1-\pi)^{K^{\gamma}}) $ & 0.8196 & 0.8413 & 0.0498 & 1.0001 & 0.2051 & 1.0153 & 0.8814 & 0.0000 & 0.0000 & 0.0100 \\
MMLU-College-v2 (70B) & Naive Ensemble & $y = \mathrm{floor} + (\mathrm{ceil} - \mathrm{floor}) \cdot \exp(-\zeta \cdot K^{-\alpha}) (1 - (1-\pi)^{K^{\gamma}}) $ & 0.8234 & 0.9262 & 3.9001 & 3.0000 & 0.3036 & 0.5000 & 0.9909 & 0.0000 & 0.0000 & 0.1000 \\
GPQA-v2-Diamond (8B) & Pass@K & $y = \mathrm{floor} + (\mathrm{ceil} - \mathrm{floor}) \cdot \exp(-\zeta \cdot K^{-\alpha})$ & 0.2223 & 0.9976 & 2.4975 & 0.8328 & X & X & 0.9996 & 0.0000 & 0.0000 & 0.2500 \\
GPQA-v2-Diamond (8B) & Weaver & $y = \mathrm{floor} + (\mathrm{ceil} - \mathrm{floor}) \cdot \exp(-\zeta \cdot K^{-\alpha}) (1 - (1-\pi)^{K^{\gamma}}) $ & 0.2316 & 0.4678 & 0.0559 & 1.0027 & 0.2347 & 0.5940 & 0.9965 & 0.0000 & 0.0002 & 0.0100 \\
GPQA-v2-Diamond (8B) & Majority1@K & $y = \mathrm{floor} + (\mathrm{ceil} - \mathrm{floor}) \cdot \exp(-\zeta \cdot K^{-\alpha}) (1 - (1-\pi)^{K^{\gamma}}) $ & 0.2773 & 0.2979 & 0.1335 & 0.9633 & 0.0479 & 2.5000 & 0.5476 & 0.0000 & 0.0001 & 0.0500 \\
GPQA-v2-Diamond (8B) & Naive Ensemble & $y = \mathrm{floor} + (\mathrm{ceil} - \mathrm{floor}) \cdot \exp(-\zeta \cdot K^{-\alpha}) (1 - (1-\pi)^{K^{\gamma}}) $ & 0.2871 & 0.3845 & 8.5633 & 3.0000 & 0.2431 & 0.5000 & 0.9643 & 0.0000 & 0.0001 & 0.0100 \\
MATH-500-v2 (8B) & Pass@K & $y = \mathrm{floor} + (\mathrm{ceil} - \mathrm{floor}) \cdot \exp(-\zeta \cdot K^{-\alpha})$ & 0.3986 & 1.0000 & 1.6393 & 0.8786 & X & X & 0.9976 & 0.0001 & 0.0001 & 0.0100 \\
MATH-500-v2 (8B) & Weaver & $y = \mathrm{floor} + (\mathrm{ceil} - \mathrm{floor}) \cdot \exp(-\zeta \cdot K^{-\alpha}) (1 - (1-\pi)^{K^{\gamma}}) $ & 0.4149 & 0.7417 & 0.0750 & 1.0042 & 0.3261 & 0.5179 & 0.9981 & 0.0000 & 0.0000 & 0.0100 \\
MATH-500-v2 (8B) & Majority1@K & $y = \mathrm{floor} + (\mathrm{ceil} - \mathrm{floor}) \cdot \exp(-\zeta \cdot K^{-\alpha}) (1 - (1-\pi)^{K^{\gamma}}) $ & 0.5061 & 0.6976 & 5.9901 & 1.1205 & 0.0782 & 2.5000 & 0.9960 & 0.0000 & 0.0000 & 0.0500 \\
MATH-500-v2 (8B) & Naive Ensemble & $y = \mathrm{floor} + (\mathrm{ceil} - \mathrm{floor}) \cdot \exp(-\zeta \cdot K^{-\alpha}) (1 - (1-\pi)^{K^{\gamma}}) $ & 0.5009 & 0.7342 & 1.6358 & 3.0000 & 0.3321 & 0.5000 & 0.9803 & 0.0001 & 0.0005 & 0.0100 \\
MMLU-Pro-v2 (8B) & Pass@K & $y = \mathrm{floor} + (\mathrm{ceil} - \mathrm{floor}) \cdot \exp(-\zeta \cdot K^{-\alpha})$ & 0.3877 & 1.0000 & 1.8798 & 0.7549 & X & X & 0.9990 & 0.0000 & 0.0000 & 0.1000 \\
MMLU-Pro-v2 (8B) & Weaver & $y = \mathrm{floor} + (\mathrm{ceil} - \mathrm{floor}) \cdot \exp(-\zeta \cdot K^{-\alpha}) (1 - (1-\pi)^{K^{\gamma}}) $ & 0.4770 & 0.6937 & 3.1648 & 3.0000 & 0.2172 & 0.5705 & 0.9986 & 0.0000 & 0.0001 & 0.0500 \\
MMLU-Pro-v2 (8B) & Majority1@K & $y = \mathrm{floor} + (\mathrm{ceil} - \mathrm{floor}) \cdot \exp(-\zeta \cdot K^{-\alpha}) (1 - (1-\pi)^{K^{\gamma}}) $ & 0.4663 & 1.0000 & 2.4923 & 0.1016 & 0.0362 & 2.5000 & 0.9600 & 0.0001 & 0.0002 & 0.0500 \\
MMLU-Pro-v2 (8B) & Naive Ensemble & $y = \mathrm{floor} + (\mathrm{ceil} - \mathrm{floor}) \cdot \exp(-\zeta \cdot K^{-\alpha}) (1 - (1-\pi)^{K^{\gamma}}) $ & 0.4777 & 0.7207 & 3.0170 & 3.0000 & 0.1613 & 0.5000 & 0.9987 & 0.0000 & 0.0000 & 0.0100 \\
MMLU-College-v2 (8B) & Pass@K & $y = \mathrm{floor} + (\mathrm{ceil} - \mathrm{floor}) \cdot \exp(-\zeta \cdot K^{-\alpha})$ & 0.4442 & 0.9912 & 1.0258 & 0.9265 & X & X & 0.9993 & 0.0000 & 0.0000 & 0.5000 \\
MMLU-College-v2 (8B) & Weaver & $y = \mathrm{floor} + (\mathrm{ceil} - \mathrm{floor}) \cdot \exp(-\zeta \cdot K^{-\alpha}) (1 - (1-\pi)^{K^{\gamma}}) $ & 0.6178 & 0.8447 & 1.1473 & 3.0000 & 0.3523 & 0.5000 & 0.9957 & 0.0000 & 0.0001 & 0.2500 \\
MMLU-College-v2 (8B) & Majority1@K & $y = \mathrm{floor} + (\mathrm{ceil} - \mathrm{floor}) \cdot \exp(-\zeta \cdot K^{-\alpha}) (1 - (1-\pi)^{K^{\gamma}}) $ & 0.6254 & 0.7186 & 0.5765 & 0.9281 & 0.2058 & 1.2287 & 0.9736 & 0.0000 & 0.0001 & 0.0100 \\
MMLU-College-v2 (8B) & Naive Ensemble & $y = \mathrm{floor} + (\mathrm{ceil} - \mathrm{floor}) \cdot \exp(-\zeta \cdot K^{-\alpha}) (1 - (1-\pi)^{K^{\gamma}}) $ & 0.6074 & 0.9813 & 1.2560 & 0.1636 & 0.3060 & 2.4651 & 0.9988 & 0.0000 & 0.0000 & 0.0100 \\
\bottomrule
\end{tabular}
}
\end{table}

%% file: tables_and_figures/FalsePositiveRates.tex
\begin{figure}[H]
   \centering
   \includegraphics[width=\linewidth]{tables_and_figures/FalsePositiveRate_Fig.pdf}
   \caption{\textbf{False Positive Rates across Verification Systems}
   }
   \label{fig:FalsePositiveRates}
\end{figure}

%% file: tables_and_figures/8B_verifier_ablations.tex
\begin{table*}%
  \caption{\textbf{\weaver{} with 8B Models Exceeds Majority Voting and Naive Ensemble across All Datasets}: Candidate are generated with Llama 3.1 8B Instruct while the weak verifiers are 8B parameters or smaller in size.}
  \centering
  \resizebox{\textwidth}{!}{%
  \setlength{\tabcolsep}{5pt}{
  \begin{tabular}{ccccccccc}
    \toprule
    & \multirow{2}{*}{\bf \begin{tabular}[c]{@{}c@{}}  \\ \\Methodology\end{tabular} } &
    \multirow{2}{*}{\bf \begin{tabular}[c]{@{}c@{}}  \\ \\Generations ($K$)\end{tabular} } & 
    \multicolumn{4}{c}{\bf Datasets} & 
    
    \multirow{2}{*}{\bf \begin{tabular}[c]{@{}c@{}}  \\ \\Average\end{tabular} } \\
    \cmidrule(lr){4-7}
    & & & \begin{tabular}[c]{@{}c@{}}  MATH\\500\end{tabular} & GPQA & \begin{tabular}[c]{@{}c@{}}  MMLU\\College\end{tabular} & \begin{tabular}[c]{@{}c@{}}  MMLU\\Pro\end{tabular} \\
    \midrule
    \multirow{6}{*}{\rotatebox[origin=c]{90}{\scriptsize \begin{tabular}[c]{@{}c@{}} \textbf{Baselines}\end{tabular}}} & First Sample & 1 & 	49.8\% &		28.3\% &		64.1\% &		46.6\% & 47.2\%		   \\
    & Majority Voting & 100 &		69.0\% &		30.5\% &		72.7\% &		56.4\% &	\textbf{57.2\%}			  \\
    & Top-Ranked RM from RewardBench \citep{lambert2024rewardbenchevaluatingrewardmodels} & 100 & 73.8\% & 25.4\% & 70.1\% & 53.4\% &  55.7\%   \\
    & Top-10 RM Ensemble from RewardBench \citep{lambert2024rewardbenchevaluatingrewardmodels} & 100 & 70.2\% & 22.1\% & 73.9\% & 49.4\% & 53.9\%    \\
    & Multi-Agent Verification \citep{lifshitz2025multiagentverificationscalingtesttime} & 100 & 65.4\% &  31.4\% & 70.5\% & 55.2\% & 55.6\%   \\
    & Self-Verification \citep{zhao2025sample} & 100 & 71.4\% & 32.2\% & 70.4\% &  53.0\% & \underline{56.8\%}    \\
    \midrule
    & \begin{tabular}[c]{@{}c@{}}  \weaver{}\end{tabular} & 100 & 80.0\% & 47.1\% & 85.7\% & 67.2\% & \textbf{70.0\%}   \\
    \midrule
    & GPT-4o-mini & 1 &		76.8\% &		38.4\% &		82.2\% &		61.8\% & 64.8\%	\\
    & Claude 3.5 Haiku & 1 &		70.0\% &		36.4\% &		75.9\% &		65.2\% & 61.9\%	\\
    & Oracle Verifier (Pass@100) & 100 &		99.2\% &		95.2\% &		98.5\% &		96.8\% & \textbf{97.4\%}		  \\ 
    \bottomrule
  \end{tabular}
  }
  }
  \label{tab:8B_verifier_ablations}
\end{table*}

%% file: tables_and_figures/8B_ScalingPlots.tex
\begin{figure}[H]
   \centering
   \includegraphics[width=\linewidth]{tables_and_figures/8B_ScalingPlots_Fig.pdf}
   \caption{
   \textbf{\weaver{} Scaling - 8B Generations and Models}
   }
   \label{fig:8B_ScalingPlots}
\end{figure}

%% file: tables_and_figures/sampling_vs_ensembling.tex
\begin{table}
\scriptsize
\caption{\textbf{Ensembling with Multiple Verifiers Outperforms Increased Sampling with Single Verifier}: Candidate responses are generated with Llama 3.3 70B Instruct while the weak verifiers range in size from 8B to 72B parameters. 
For details on prompting, please see Appendix \ref{app:individual-verifier-optimization}.}
\centering
{%
\setlength{\tabcolsep}{5pt}{
\begin{tabular}{cccc}
\toprule
\multirow{2}{*}{\bf Methodology} &
\multicolumn{3}{c}{\bf Benchmarks} \\
\cmidrule(lr){2-4}
& MATH500 & GPQA & MMLU Pro \\
\midrule
First Sample & 78.0\% & 42.9\%  & 69.9\%  \\ \midrule
Majority Voting & 83.0\% & 47.4\% & 74.4\% \\ \midrule
\begin{tabular}[c]{@{}c@{}} Best Reward Model \\ (1 Score)\end{tabular} & 94.4\% & 58.4\% & 81.8\% \\ \midrule
\begin{tabular}[c]{@{}c@{}} Best Reward Model \\ (5 Scores, 5 Prompts)\end{tabular} & 93.2\% & 55.3\% & 82.5\%  \\ \midrule
\begin{tabular}[c]{@{}c@{}} Top-5 Most Accurate \\ Reward Models\end{tabular} & \textbf{95.4\%} & \underline{64.1\%} & \textbf{87.3\%} \\ \midrule
\begin{tabular}[c]{@{}c@{}} Best LM Judge \\ (1 Score)\end{tabular} & 90.2\% & 61.1\% & 79.5\% \\ \midrule
\begin{tabular}[c]{@{}c@{}} Best LM Judge \\ (5 Scores, 5 Prompts)\end{tabular} & 88.1\% & 57.2\% & 80.8\% \\ \midrule
\begin{tabular}[c]{@{}c@{}} Top-5 Most Accurate \\ LM Judges\end{tabular} & \underline{93.4\%} & \textbf{65.2\%} & \underline{85.4\%} \\
\bottomrule
\end{tabular}
}
}
\label{tab:sampling_vs_ensembling}
\end{table}

%% file: tables_and_figures/Verifiers_vs_Generations.tex
\begin{figure*}[t]
   \centering
   \includegraphics[width=.90\linewidth]%
{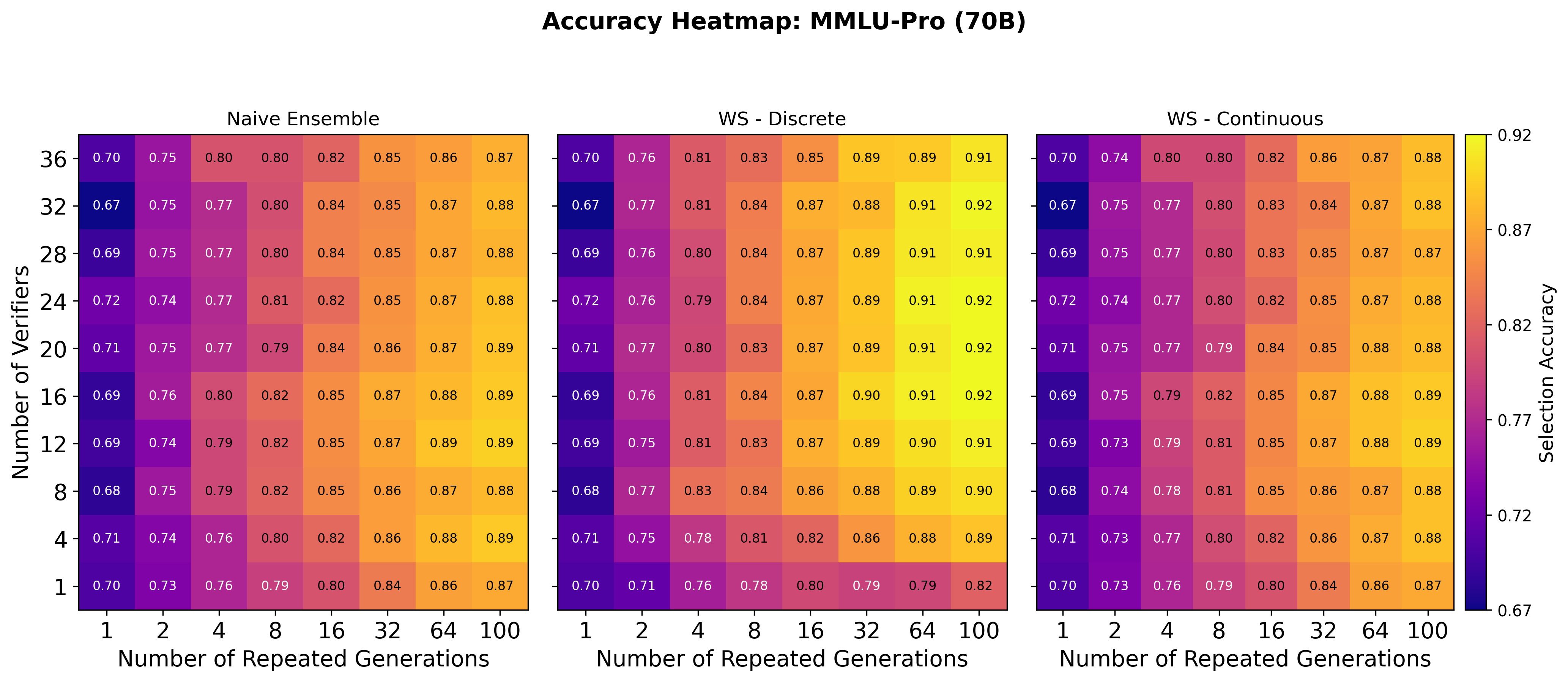}
\caption{\textbf{\weaver{} Performance Improvements from Scaling Generations and Verifiers}: Increased candidate generations and weak verifiers available generally improves performance. %
}
  \label{fig:Verifiers_vs_Generations_Tradeoff}
\end{figure*}

In \cref{fig:Verifiers_vs_Generations_Tradeoff} illustrates how the number of verifiers and repeated generations interact to influence success rate. We observe that increasing the number of generations tends to be more effective than increasing the number of verifiers alone—but only when paired with the right verification strategy. For example, naive ensembling of verifiers plateaus in performance even as more generations are added, whereas \weaver{} continues to improve with both axes. This highlights that generation diversity is a stronger driver of performance than verifier count alone, and that weak supervision methods like \weaver{} are essential to fully leverage this diversity. We illustrate the verification generation tradeoff for additional datasets in Appendix \ref{app:scaling_trends_of_weaver}.

%% file: tables_and_figures/distillation_diagram.tex
\begin{figure}[H]
   \centering
   \includegraphics[width=1.0\linewidth]{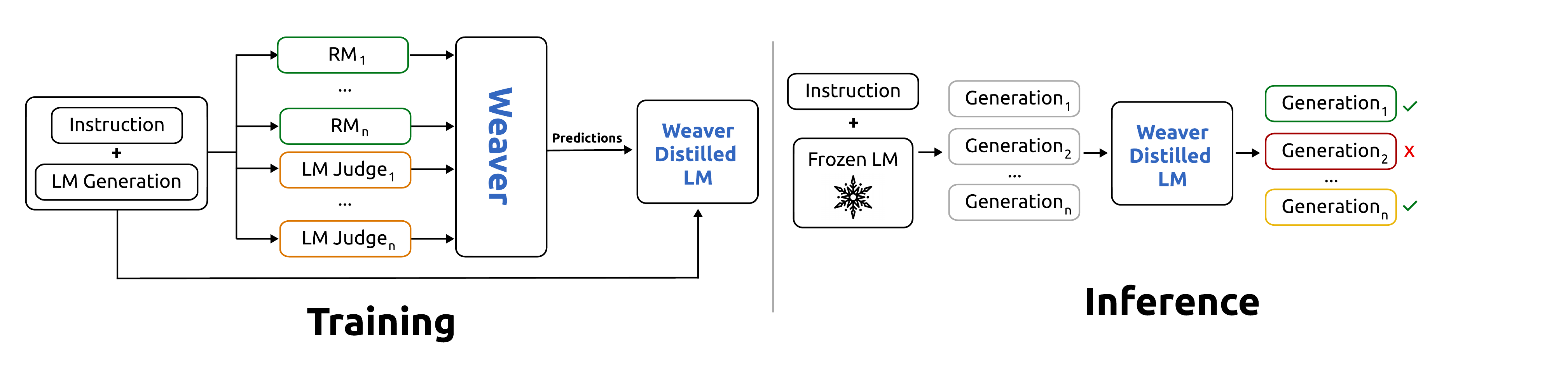}
\caption{\textbf{Overview of \weaver{} Distillation (Section \ref{sec:weaver_distillation})}}
   \label{fig:distillation_diagram}
\end{figure}

%% file: tables_and_figures/WeaverDistilledParetoFrontiers.tex
\begin{figure}[h]
   \centering
   \includegraphics[width=1.0\linewidth]{tables_and_figures/WeaverDistilledParetoFrontiers_fig.pdf}
\caption{\textbf{\weaver{} Distilled - Pareto Frontiers}: $^*$We train/evaluate on an 80:20 split.}
   \label{fig:WeaverDistilledParetoFrontiers}
\end{figure}

%% file: tables_and_figures/Weaver_Distillation_Ablations.tex
\begin{table}[h]
  \small
  \caption{\textbf{Distillation Comparison of \weaver{} and Naive Ensemble Across Different Training Set Sizes}}
  \centering
  {
  \begin{tabularx}{\textwidth}{cc>{\centering\arraybackslash}X>{\centering\arraybackslash}X>{\centering\arraybackslash}X>{\centering\arraybackslash}X>{\centering\arraybackslash}Xc}
    \toprule
    \multirow{2}{*}{\bf \begin{tabular}[c]{@{}c@{}}  \\ Methodology\end{tabular} } &
    \multirow{2}{*}{\bf \begin{tabular}[c]{@{}c@{}}  \\ Dataset\end{tabular} } & 
    \multicolumn{5}{c}{\bf Training Set as Percentage of Entire Dataset} &
    \multirow{2}{*}{\bf \begin{tabular}[c]{@{}c@{}}  Full\\System\end{tabular} } \\
    \cmidrule(lr){3-7}
    & & 5\% & 10\% & 20\% & 50\% & 80\% & \\
    \midrule
    \multirow{4}{*}{\weaver{}} & MATH500 & 78.4\% & 80.7\% & 83.9\% & 88.2\% & {91.4\%} & 93.4\% \\
    & GPQA Diamond & 42.6\% & 46.8\% & 52.7\% & 63.1\% & {71.8\%} & 73.2\% \\
    & MMLU College & 83.5\% & 85.2\% & 87.6\% & 91.0\% & 93.1\% & 94.9\% \\
    & MMLU Pro & 69.2\% & 72.5\% & 76.8\% & 83.7\% & {87.8\%} & 90.2\% \\
    \midrule
    \multirow{4}{*}{NaiveEnsemble} & MATH500 & 77.8\% & 79.6\% & 82.1\% & 86.4\% & 89.1\% & 92.4\% \\
    & GPQA Diamond & 42.1\% & 44.7\% & 48.9\% & 56.2\% & 62.8\% & 66.2\% \\
    & MMLU College & 84.0\% & 85.3\% & 87.2\% & 90.8\% & {93.5\%} & 95.1\% \\
    & MMLU Pro & 69.5\% & 71.8\% & 74.9\% & 80.3\% & 84.7\% & 87.4\% \\
    \bottomrule
  \end{tabularx}
  }
  \label{tab:Weaver_Distillation_Ablations}
\end{table}

%% file: neurips_2025.bbl
\begin{thebibliography}{100}

\bibitem{abebuchanan2024}
Taiga Abe, E.~Kelly Buchanan, Geoff Pleiss, and John~Patrick Cunningham.
\newblock Pathologies of predictive diversity in deep ensembles.
\newblock {\em Transactions on Machine Learning Research}, 2024.
\newblock Featured Certification.

\bibitem{Anthropic2025Claude37}
{Anthropic}.
\newblock Claude 3.7 sonnet and claude code, February 2025.
\newblock Announcement blog post, 5 min read.

\bibitem{arora2022askanythingsimplestrategy}
Simran Arora, Avanika Narayan, Mayee~F. Chen, Laurel Orr, Neel Guha, Kush Bhatia, Ines Chami, Frederic Sala, and Christopher Ré.
\newblock Ask me anything: A simple strategy for prompting language models, 2022.

\bibitem{askell2021general}
Amanda Askell, Yuntao Bai, Anna Chen, Dawn Drain, Deep Ganguli, Tom Henighan, Andy Jones, Nicholas Joseph, Ben Mann, Nova DasSarma, et~al.
\newblock {A General Language Assistant as a Laboratory for Alignment}.
\newblock {\em ArXiv Preprint arXiv:2112.00861}, 2021.

\bibitem{bradley1952rank}
Ralph~Allan Bradley and Milton~E Terry.
\newblock {Rank Analysis of Incomplete Block Designs: I. The Method of Paired Comparisons}.
\newblock {\em Biometrika}, 39(3/4):324--345, 1952.

\bibitem{brown2024largelanguagemonkeysscaling}
Bradley Brown, Jordan Juravsky, Ryan Ehrlich, Ronald Clark, Quoc~V. Le, Christopher Ré, and Azalia Mirhoseini.
\newblock Large language monkeys: Scaling inference compute with repeated sampling, 2024.

\bibitem{cao2007learning}
Zhe Cao, Tao Qin, Tie-Yan Liu, Ming-Feng Tsai, and Hang Li.
\newblock Learning to rank: from pairwise approach to listwise approach.
\newblock In {\em Proceedings of the 24th international conference on Machine learning}, pages 129--136, 2007.

\bibitem{chaudhari2024rlhfdecipheredcriticalanalysis}
Shreyas Chaudhari, Pranjal Aggarwal, Vishvak Murahari, Tanmay Rajpurohit, Ashwin Kalyan, Karthik Narasimhan, Ameet Deshpande, and Bruno~Castro da~Silva.
\newblock {RLHF Deciphered: A Critical Analysis of Reinforcement Learning from Human Feedback for LLMs}, 2024.

\bibitem{chen2024humans}
Guiming~Hardy Chen, Shunian Chen, Ziche Liu, Feng Jiang, and Benyou Wang.
\newblock {Humans or LLMs as the Judge? A Study on Judgement Bias}.
\newblock {\em ArXiv Preprint arXiv:2402.10669}, 2024.

\bibitem{chen2025sets}
Jiefeng Chen, Jie Ren, Xinyun Chen, Chengrun Yang, Ruoxi Sun, and Sercan~{\"O} Ar{\i}k.
\newblock Sets: Leveraging self-verification and self-correction for improved test-time scaling.
\newblock {\em arXiv preprint arXiv:2501.19306}, 2025.

\bibitem{chen2024more}
Lingjiao Chen, Jared~Quincy Davis, Boris Hanin, Peter Bailis, Ion Stoica, Matei Zaharia, and James Zou.
\newblock Are more llm calls all you need? towards scaling laws of compound inference systems.
\newblock {\em arXiv preprint arXiv:2403.02419}, 2024.

\bibitem{chen2024llmcallsneedscaling}
Lingjiao Chen, Jared~Quincy Davis, Boris Hanin, Peter Bailis, Ion Stoica, Matei Zaharia, and James Zou.
\newblock Are more llm calls all you need? towards scaling laws of compound inference systems, 2024.

\bibitem{chen2021evaluating}
Mark Chen, Jerry Tworek, Heewoo Jun, Qiming Yuan, Henrique~Ponde de~Oliveira~Pinto, Jared Kaplan, Harri Edwards, Yuri Burda, Nicholas Joseph, Greg Brockman, Alex Ray, Raul Puri, Gretchen Krueger, Michael Petrov, Heidy Khlaaf, Girish Sastry, Pamela Mishkin, Brooke Chan, Scott Gray, Nick Ryder, Mikhail Pavlov, Alethea Power, Lukasz Kaiser, Mohammad Bavarian, Clemens Winter, Philippe Tillet, Felipe~Petroski Such, Dave Cummings, Matthias Plappert, Fotios Chantzis, Elizabeth Barnes, Ariel Herbert-Voss, William~Hebgen Guss, Alex Nichol, Alex Paino, Nikolas Tezak, Jie Tang, Igor Babuschkin, Suchir Balaji, Shantanu Jain, William Saunders, Christopher Hesse, Andrew~N. Carr, Jan Leike, Josh Achiam, Vedant Misra, Evan Morikawa, Alec Radford, Matthew Knight, Miles Brundage, Mira Murati, Katie Mayer, Peter Welinder, Bob McGrew, Dario Amodei, Sam McCandlish, Ilya Sutskever, and Wojciech Zaremba.
\newblock Evaluating large language models trained on code, 2021.

\bibitem{chen2022shoring}
Mayee~F. Chen, Daniel~Y. Fu, Dyah Adila, Michael Zhang, Frederic Sala, Kayvon Fatahalian, and Christopher R\'e.
\newblock Shoring up the foundations: fusing model embeddings and weak supervision.
\newblock In James Cussens and Kun Zhang, editors, {\em Proceedings of the Thirty-Eighth Conference on Uncertainty in Artificial Intelligence}, volume 180 of {\em Proceedings of Machine Learning Research}, pages 357--367. PMLR, 01--05 Aug 2022.

\bibitem{chiang2024chatbot}
Wei-Lin Chiang, Lianmin Zheng, Ying Sheng, Anastasios~Nikolas Angelopoulos, Tianle Li, Dacheng Li, Hao Zhang, Banghua Zhu, Michael Jordan, Joseph~E. Gonzalez, and Ion Stoica.
\newblock Chatbot arena: An open platform for evaluating llms by human preference, 2024.

\bibitem{christiano2017deep}
Paul~F Christiano, Jan Leike, Tom Brown, Miljan Martic, Shane Legg, and Dario Amodei.
\newblock {Deep Reinforcement Learning from Human Preferences}.
\newblock {\em {Advances in Neural Information Processing Systems}}, 30, 2017.

\bibitem{clark2021thatshumangoldevaluating}
Elizabeth Clark, Tal August, Sofia Serrano, Nikita Haduong, Suchin Gururangan, and Noah~A. Smith.
\newblock All that's 'human' is not gold: Evaluating human evaluation of generated text, 2021.

\bibitem{cui2025process}
Ganqu Cui, Lifan Yuan, Zefan Wang, Hanbin Wang, Wendi Li, Bingxiang He, Yuchen Fan, Tianyu Yu, Qixin Xu, Weize Chen, et~al.
\newblock Process reinforcement through implicit rewards.
\newblock {\em arXiv preprint arXiv:2502.01456}, 2025.

\bibitem{dorka2024quantile}
Nicolai Dorka.
\newblock Quantile regression for distributional reward models in rlhf.
\newblock {\em arXiv preprint arXiv:2409.10164}, 2024.

\bibitem{dubois2024length}
Yann Dubois, Bal{\'a}zs Galambosi, Percy Liang, and Tatsunori~B Hashimoto.
\newblock {Length-Controlled AlpacaEval: A Simple Way to Debias Automatic Evaluators}.
\newblock {\em ArXiv Preprint arXiv:2404.04475}, 2024.

\bibitem{dubois2024alpacafarm}
Yann Dubois, Chen~Xuechen Li, Rohan Taori, Tianyi Zhang, Ishaan Gulrajani, Jimmy Ba, Carlos Guestrin, Percy~S Liang, and Tatsunori~B Hashimoto.
\newblock {AlpacaFarm: A Simulation Framework for Methods that Learn from Human Feedback}.
\newblock {\em Advances in Neural Information Processing Systems}, 36, 2024.

\bibitem{eisenstein2023reward}
Jacob Eisenstein, Jonathan Berant, Chirag Nagpal, Alekh Agarwal, Ahmad Beirami, Alexander~Nicholas D'Amour, Krishnamurthy~Dj Dvijotham, Katherine~A Heller, Stephen~Robert Pfohl, and Deepak Ramachandran.
\newblock {Reward Model Underspecification in Language Model Alignment}.
\newblock In {\em NeurIPS 2023 Workshop on Distribution Shifts: New Frontiers with Foundation Models}, 2023.

\bibitem{eisenstein2023helping}
Jacob Eisenstein, Chirag Nagpal, Alekh Agarwal, Ahmad Beirami, Alex D'Amour, DJ~Dvijotham, Adam Fisch, Katherine Heller, Stephen Pfohl, Deepak Ramachandran, et~al.
\newblock Helping or herding? reward model ensembles mitigate but do not eliminate reward hacking.
\newblock {\em arXiv preprint arXiv:2312.09244}, 2023.

\bibitem{es2023ragas}
Shahul Es, Jithin James, Luis Espinosa-Anke, and Steven Schockaert.
\newblock {RAGAs: Automated Evaluation of Retrieval Augmented Generation}.
\newblock {\em ArXiv Preprint arXiv:2309.15217}, 2023.

\bibitem{phan2025humanitysexam}
Long~Phan et~al.
\newblock Humanity's last exam, 2025.

\bibitem{fu2020fast}
Daniel Fu, Mayee Chen, Frederic Sala, Sarah Hooper, Kayvon Fatahalian, and Christopher R{\'e}.
\newblock Fast and three-rious: Speeding up weak supervision with triplet methods.
\newblock In {\em International conference on machine learning}, pages 3280--3291. PMLR, 2020.

\bibitem{fu2023gptscore}
Jinlan Fu, See-Kiong Ng, Zhengbao Jiang, and Pengfei Liu.
\newblock {GPTScore: Evaluate as You Desire}.
\newblock {\em Proceedings of the 2024 Conference of the North American Chapter of the Association for Computational Linguistics: Human Language Technologies (Volume 1: Long Papers)}, 2023.

\bibitem{guhasmoothie}
Neel Guha, Mayee~F Chen, Trevor Chow, Ishan~S Khare, and Christopher Re.
\newblock Smoothie: Label free language model routing.
\newblock In {\em The Thirty-eighth Annual Conference on Neural Information Processing Systems}, 2024.

\bibitem{hall2003generalized}
Alastair~R Hall.
\newblock Generalized method of moments.
\newblock {\em A companion to theoretical econometrics}, pages 230--255, 2003.

\bibitem{HazyResearchMetal}
HazyResearch.
\newblock metal.
\newblock \url{https://https://github.com/HazyResearch/metal}, 2018.

\bibitem{hendrycks2021measuring}
Dan Hendrycks, Collin Burns, Saurav Kadavath, Akul Arora, Steven Basart, Eric Tang, Dawn Song, and Jacob Steinhardt.
\newblock Measuring mathematical problem solving with the math dataset.
\newblock {\em arXiv preprint arXiv:2103.03874}, 2021.

\bibitem{hoffmann2022training}
Jordan Hoffmann, Sebastian Borgeaud, Arthur Mensch, Elena Buchatskaya, Trevor Cai, Eliza Rutherford, Diego de~Las Casas, Lisa~Anne Hendricks, Johannes Welbl, Aidan Clark, et~al.
\newblock Training compute-optimal large language models.
\newblock {\em arXiv preprint arXiv:2203.15556}, 2022.

\bibitem{hoffmann2022trainingcomputeoptimallargelanguage}
Jordan Hoffmann, Sebastian Borgeaud, Arthur Mensch, Elena Buchatskaya, Trevor Cai, Eliza Rutherford, Diego de~Las~Casas, Lisa~Anne Hendricks, Johannes Welbl, Aidan Clark, Tom Hennigan, Eric Noland, Katie Millican, George van~den Driessche, Bogdan Damoc, Aurelia Guy, Simon Osindero, Karen Simonyan, Erich Elsen, Jack~W. Rae, Oriol Vinyals, and Laurent Sifre.
\newblock Training compute-optimal large language models, 2022.

\bibitem{hosking2024humanfeedbackgoldstandard}
Tom Hosking, Phil Blunsom, and Max Bartolo.
\newblock Human feedback is not gold standard, 2024.

\bibitem{jain2024livecodebenchholisticcontaminationfree}
Naman Jain, King Han, Alex Gu, Wen-Ding Li, Fanjia Yan, Tianjun Zhang, Sida Wang, Armando Solar-Lezama, Koushik Sen, and Ion Stoica.
\newblock Livecodebench: Holistic and contamination free evaluation of large language models for code, 2024.

\bibitem{kalra2025verdictlibraryscalingjudgetime}
Nimit Kalra and Leonard Tang.
\newblock Verdict: A library for scaling judge-time compute, 2025.

\bibitem{kaplan2020scalinglawsneurallanguage}
Jared Kaplan, Sam McCandlish, Tom Henighan, Tom~B. Brown, Benjamin Chess, Rewon Child, Scott Gray, Alec Radford, Jeffrey Wu, and Dario Amodei.
\newblock Scaling laws for neural language models, 2020.

\bibitem{karpinska2021perilsusingmechanicalturk}
Marzena Karpinska, Nader Akoury, and Mohit Iyyer.
\newblock The perils of using mechanical turk to evaluate open-ended text generation, 2021.

\bibitem{khattab2023dspy}
Omar Khattab, Arnav Singhvi, Paridhi Maheshwari, Zhiyuan Zhang, Keshav Santhanam, Sri Vardhamanan, Saiful Haq, Ashutosh Sharma, Thomas~T. Joshi, Hanna Moazam, Heather Miller, Matei Zaharia, and Christopher Potts.
\newblock Dspy: Compiling declarative language model calls into self-improving pipelines.
\newblock {\em arXiv preprint arXiv:2310.03714}, 2023.

\bibitem{kingma2015adam}
Diederik~P. Kingma and Jimmy Ba.
\newblock Adam: A method for stochastic optimization, 2017.

\bibitem{kirchner2024prover}
Jan~Hendrik Kirchner, Yining Chen, Harri Edwards, Jan Leike, Nat McAleese, and Yuri Burda.
\newblock Prover-verifier games improve legibility of llm outputs.
\newblock {\em arXiv preprint arXiv:2407.13692}, 2024.

\bibitem{lambert2023alignment}
Nathan Lambert and Roberto Calandra.
\newblock The alignment ceiling: Objective mismatch in reinforcement learning from human feedback.
\newblock {\em ArXiv Preprint arXiv:2311.00168}, 2023.

\bibitem{lambert2024rewardbenchevaluatingrewardmodels}
Nathan Lambert, Valentina Pyatkin, Jacob Morrison, LJ~Miranda, Bill~Yuchen Lin, Khyathi Chandu, Nouha Dziri, Sachin Kumar, Tom Zick, Yejin Choi, Noah~A. Smith, and Hannaneh Hajishirzi.
\newblock {RewardBench: Evaluating Reward Models for Language Modeling}, 2024.

\bibitem{alpaca_eval}
Xuechen Li, Tianyi Zhang, Yann Dubois, Rohan Taori, Ishaan Gulrajani, Carlos Guestrin, Percy Liang, and Tatsunori~B. Hashimoto.
\newblock Alpacaeval: An automatic evaluator of instruction-following models.
\newblock \url{https://github.com/tatsu-lab/alpaca_eval}, 2023.

\bibitem{lifshitz2025multiagentverificationscalingtesttime}
Shalev Lifshitz, Sheila~A. McIlraith, and Yilun Du.
\newblock Multi-agent verification: Scaling test-time compute with multiple verifiers, 2025.

\bibitem{lightman2023letsverifystepstep}
Hunter Lightman, Vineet Kosaraju, Yura Burda, Harri Edwards, Bowen Baker, Teddy Lee, Jan Leike, John Schulman, Ilya Sutskever, and Karl Cobbe.
\newblock Let's verify step by step, 2023.

\bibitem{skyworkreward2024}
Chris~Yuhao Liu and Liang Zeng.
\newblock {Skywork Reward Model Series}.
\newblock \url{https://huggingface.co/Skywork}, September 2024.

\bibitem{liu2024skywork}
Chris~Yuhao Liu, Liang Zeng, Jiacai Liu, Rui Yan, Jujie He, Chaojie Wang, Shuicheng Yan, Yang Liu, and Yahui Zhou.
\newblock Skywork-reward: Bag of tricks for reward modeling in llms.
\newblock {\em arXiv preprint arXiv:2410.18451}, 2024.

\bibitem{liu20251bllmsurpass405b}
Runze Liu, Junqi Gao, Jian Zhao, Kaiyan Zhang, Xiu Li, Biqing Qi, Wanli Ouyang, and Bowen Zhou.
\newblock Can 1b llm surpass 405b llm? rethinking compute-optimal test-time scaling, 2025.

\bibitem{liu2025can}
Runze Liu, Junqi Gao, Jian Zhao, Kaiyan Zhang, Xiu Li, Biqing Qi, Wanli Ouyang, and Bowen Zhou.
\newblock Can 1b llm surpass 405b llm? rethinking compute-optimal test-time scaling.
\newblock {\em arXiv preprint arXiv:2502.06703}, 2025.

\bibitem{liu2023g}
Yang Liu, Dan Iter, Yichong Xu, Shuohang Wang, Ruochen Xu, and Chenguang Zhu.
\newblock G-eval: Nlg evaluation using gpt-4 with better human alignment.
\newblock {\em ArXiv Preprint arXiv:2303.16634}, 2023.

\bibitem{meta2025llama4}
{Meta}.
\newblock The {Llama} 4 herd: {The} beginning of a new era of natively multimodal {AI} innovation, 4 2025.
\newblock Accessed: 2025-05-12.

\bibitem{INF-ORM-Llama3.1-70B}
Xiaoyu~Tan Minghao~Yang, Chao~Qu.
\newblock Inf-orm-llama3.1-70b, 2024.

\bibitem{openaigpt4}
OpenAI.
\newblock {GPT-4 Technical Report}.
\newblock {\em ArXiv Preprint arXiv:2303.08774}, 2023.

\bibitem{OpenAIo3mini2025}
{OpenAI}.
\newblock Openai o3-mini system card.
\newblock Technical report, OpenAI, January 2025.
\newblock Publication.

\bibitem{ouyang2022training}
Long Ouyang, Jeffrey Wu, Xu~Jiang, Diogo Almeida, Carroll Wainwright, Pamela Mishkin, Chong Zhang, Sandhini Agarwal, Katarina Slama, Alex Ray, et~al.
\newblock Training language models to follow instructions with human feedback.
\newblock {\em {Advances in Neural Information Processing Systems}}, 35:27730--27744, 2022.

\bibitem{pan2024human}
Qian Pan, Zahra Ashktorab, Michael Desmond, Mart{\'i}n~Santill{\'a}n Cooper, James Johnson, Rahul Nair, Elizabeth Daly, and Werner Geyer.
\newblock {Human-Centered Design Recommendations for LLM-as-a-judge}.
\newblock In {\em Proceedings of the 1st Human-Centered Large Language Modeling Workshop}, pages 16--29, 2024.

\bibitem{puri2025probabilisticinferenceapproachinferencetime}
Isha Puri, Shivchander Sudalairaj, Guangxuan Xu, Kai Xu, and Akash Srivastava.
\newblock A probabilistic inference approach to inference-time scaling of llms using particle-based monte carlo methods, 2025.

\bibitem{quan2025codeelobenchmarkingcompetitionlevelcode}
Shanghaoran Quan, Jiaxi Yang, Bowen Yu, Bo~Zheng, Dayiheng Liu, An~Yang, Xuancheng Ren, Bofei Gao, Yibo Miao, Yunlong Feng, Zekun Wang, Jian Yang, Zeyu Cui, Yang Fan, Yichang Zhang, Binyuan Hui, and Junyang Lin.
\newblock Codeelo: Benchmarking competition-level code generation of llms with human-comparable elo ratings, 2025.

\bibitem{ratner2020snorkel}
Alexander Ratner, Stephen~H Bach, Henry Ehrenberg, Jason Fries, Sen Wu, and Christopher R{\'e}.
\newblock Snorkel: rapid training data creation with weak supervision.
\newblock {\em The VLDB Journal}, 29(2):709--730, 2020.

\bibitem{ratner2019training}
Alexander Ratner, Braden Hancock, Jared Dunnmon, Frederic Sala, Shreyash Pandey, and Christopher R{\'e}.
\newblock Training complex models with multi-task weak supervision.
\newblock In {\em Proceedings of the AAAI Conference on Artificial Intelligence}, volume~33, pages 4763--4771, 2019.

\bibitem{ratner2016data}
Alexander~J Ratner, Christopher~M De~Sa, Sen Wu, Daniel Selsam, and Christopher R{\'e}.
\newblock Data programming: Creating large training sets, quickly.
\newblock {\em Advances in neural information processing systems}, 29, 2016.

\bibitem{ravi2024lynxopensourcehallucination}
Selvan~Sunitha Ravi, Bartosz Mielczarek, Anand Kannappan, Douwe Kiela, and Rebecca Qian.
\newblock {Lynx: An Open Source Hallucination Evaluation Model}, 2024.

\bibitem{reimers-2020-multilingual-sentence-bert}
Nils Reimers and Iryna Gurevych.
\newblock Making monolingual sentence embeddings multilingual using knowledge distillation.
\newblock In {\em Proceedings of the 2020 Conference on Empirical Methods in Natural Language Processing}. Association for Computational Linguistics, 11 2020.

\bibitem{rein2024gpqa}
David Rein, Betty~Li Hou, Asa~Cooper Stickland, Jackson Petty, Richard~Yuanzhe Pang, Julien Dirani, Julian Michael, and Samuel~R. Bowman.
\newblock {GPQA}: A graduate-level google-proof q\&a benchmark.
\newblock In {\em First Conference on Language Modeling}, 2024.

\bibitem{rosset2003margin}
Saharon Rosset, Ji~Zhu, and Trevor Hastie.
\newblock Margin maximizing loss functions.
\newblock {\em Advances in neural information processing systems}, 16, 2003.

\bibitem{saad2023ares}
Jon Saad-Falcon, Omar Khattab, Christopher Potts, and Matei Zaharia.
\newblock {ARES: An Automated Evaluation Framework for Retrieval-Augmented Generation Systems}.
\newblock {\em ArXiv Preprint arXiv:2311.09476}, 2023.

\bibitem{saad2024archon}
Jon Saad-Falcon, Adrian~Gamarra Lafuente, Shlok Natarajan, Nahum Maru, Hristo Todorov, Etash Guha, E~Kelly Buchanan, Mayee Chen, Neel Guha, Christopher R{\'e}, and Azalia Mirhoseini.
\newblock Archon: An architecture search framework for inference-time techniques.
\newblock {\em arXiv preprint arXiv:2409.15254}, 2024.

\bibitem{saadfalcon2024lmunitfinegrainedevaluationnatural}
Jon Saad-Falcon, Rajan Vivek, William Berrios, Nandita~Shankar Naik, Matija Franklin, Bertie Vidgen, Amanpreet Singh, Douwe Kiela, and Shikib Mehri.
\newblock Lmunit: Fine-grained evaluation with natural language unit tests, 2024.

\bibitem{schapire2013explaining}
Robert~E Schapire.
\newblock Explaining adaboost.
\newblock In {\em Empirical inference: festschrift in honor of vladimir N. Vapnik}, pages 37--52. Springer, 2013.

\bibitem{shin2021universalizing}
Changho Shin, Winfred Li, Harit Vishwakarma, Nicholas Roberts, and Frederic Sala.
\newblock Universalizing weak supervision.
\newblock {\em arXiv preprint arXiv:2112.03865}, 2021.

\bibitem{singhal2023long}
Prasann Singhal, Tanya Goyal, Jiacheng Xu, and Greg Durrett.
\newblock A long way to go: Investigating length correlations in rlhf.
\newblock {\em ArXiv Preprint arXiv:2310.03716}, 2023.

\bibitem{singhi2025solveverifycomputeoptimalproblem}
Nishad Singhi, Hritik Bansal, Arian Hosseini, Aditya Grover, Kai-Wei Chang, Marcus Rohrbach, and Anna Rohrbach.
\newblock When to solve, when to verify: Compute-optimal problem solving and generative verification for llm reasoning, 2025.

\bibitem{singhi2025whentosolve}
Nishad Singhi, Hritik Bansal, Arian Hosseini, Aditya Grover, Kai-Wei Chang, Marcus Rohrbach, and Anna Rohrbach.
\newblock When to solve, when to verify: Compute-optimal problem solving and generative verification for llm reasoning, 2025.

\bibitem{snell2024scalingllmtesttimecompute}
Charlie Snell, Jaehoon Lee, Kelvin Xu, and Aviral Kumar.
\newblock Scaling llm test-time compute optimally can be more effective than scaling model parameters, 2024.

\bibitem{song2025prmbenchfinegrainedchallengingbenchmark}
Mingyang Song, Zhaochen Su, Xiaoye Qu, Jiawei Zhou, and Yu~Cheng.
\newblock Prmbench: A fine-grained and challenging benchmark for process-level reward models, 2025.

\bibitem{song2025mindgapexaminingselfimprovement}
Yuda Song, Hanlin Zhang, Carson Eisenach, Sham Kakade, Dean Foster, and Udaya Ghai.
\newblock Mind the gap: Examining the self-improvement capabilities of large language models, 2025.

\bibitem{stroebl2024inferencescalingflawslimits}
Benedikt Stroebl, Sayash Kapoor, and Arvind Narayanan.
\newblock Inference scaling flaws: The limits of llm resampling with imperfect verifiers, 2024.

\bibitem{suzgun2022challengingbigbenchtaskschainofthought}
Mirac Suzgun, Nathan Scales, Nathanael Schärli, Sebastian Gehrmann, Yi~Tay, Hyung~Won Chung, Aakanksha Chowdhery, Quoc~V. Le, Ed~H. Chi, Denny Zhou, and Jason Wei.
\newblock Challenging big-bench tasks and whether chain-of-thought can solve them, 2022.

\bibitem{tang2024minicheckefficientfactcheckingllms}
Liyan Tang, Philippe Laban, and Greg Durrett.
\newblock {MiniCheck: Efficient Fact-Checking of LLMs on Grounding Documents}, 2024.

\bibitem{verga2024replacingjudgesjuriesevaluating}
Pat Verga, Sebastian Hofstatter, Sophia Althammer, Yixuan Su, Aleksandra Piktus, Arkady Arkhangorodsky, Minjie Xu, Naomi White, and Patrick Lewis.
\newblock Replacing judges with juries: Evaluating llm generations with a panel of diverse models, 2024.

\bibitem{vishwakarma2022lifting}
Harit Vishwakarma and Frederic Sala.
\newblock Lifting weak supervision to structured prediction.
\newblock {\em Advances in Neural Information Processing Systems}, 35:37563--37574, 2022.

\bibitem{wang2024secretsrlhflargelanguage}
Binghai Wang, Rui Zheng, Lu~Chen, Yan Liu, Shihan Dou, Caishuang Huang, Wei Shen, Senjie Jin, Enyu Zhou, Chenyu Shi, Songyang Gao, Nuo Xu, Yuhao Zhou, Xiaoran Fan, Zhiheng Xi, Jun Zhao, Xiao Wang, Tao Ji, Hang Yan, Lixing Shen, Zhan Chen, Tao Gui, Qi~Zhang, Xipeng Qiu, Xuanjing Huang, Zuxuan Wu, and Yu-Gang Jiang.
\newblock {Secrets of RLHF in Large Language Models Part II: Reward Modeling}, 2024.

\bibitem{ArmoRM}
Haoxiang Wang, Wei Xiong, Tengyang Xie, Han Zhao, and Tong Zhang.
\newblock Interpretable preferences via multi-objective reward modeling and mixture-of-experts.
\newblock {\em ArXiv Preprint arXiv:2406.12845}, 2024.

\bibitem{wang2023chatgpt}
Jiaan Wang, Yunlong Liang, Fandong Meng, Zengkui Sun, Haoxiang Shi, Zhixu Li, Jinan Xu, Jianfeng Qu, and Jie Zhou.
\newblock {Is ChatGPT a good NLG Evaluator? A Preliminary Study}.
\newblock {\em ArXiv Preprint arXiv:2303.04048}, 2023.

\bibitem{wang2025improvingmodelalignmentcollective}
Junlin Wang, Roy Xie, Shang Zhu, Jue Wang, Ben Athiwaratkun, Bhuwan Dhingra, Shuaiwen~Leon Song, Ce~Zhang, and James Zou.
\newblock Improving model alignment through collective intelligence of open-source llms, 2025.

\bibitem{wang2020vatexlargescalehighqualitymultilingual}
Xin Wang, Jiawei Wu, Junkun Chen, Lei Li, Yuan-Fang Wang, and William~Yang Wang.
\newblock Vatex: A large-scale, high-quality multilingual dataset for video-and-language research, 2020.

\bibitem{wang2024mmlu}
Yubo Wang, Xueguang Ma, Ge~Zhang, Yuansheng Ni, Abhranil Chandra, Shiguang Guo, Weiming Ren, Aaran Arulraj, Xuan He, Ziyan Jiang, et~al.
\newblock Mmlu-pro: A more robust and challenging multi-task language understanding benchmark.
\newblock {\em arXiv preprint arXiv:2406.01574}, 2024.

\bibitem{wang2023helpsteermultiattributehelpfulnessdataset}
Zhilin Wang, Yi~Dong, Jiaqi Zeng, Virginia Adams, Makesh~Narsimhan Sreedhar, Daniel Egert, Olivier Delalleau, Jane~Polak Scowcroft, Neel Kant, Aidan Swope, and Oleksii Kuchaiev.
\newblock {HelpSteer: Multi-Attribute Helpfulness Dataset for SteerLM}, 2023.

\bibitem{wang2024transforming}
Zihao Wang, Chirag Nagpal, Jonathan Berant, Jacob Eisenstein, Alex D'Amour, Sanmi Koyejo, and Victor Veitch.
\newblock Transforming and combining rewards for aligning large language models.
\newblock {\em arXiv preprint arXiv:2402.00742}, 2024.

\bibitem{modernbert}
Benjamin Warner, Antoine Chaffin, Benjamin Clavié, Orion Weller, Oskar Hallström, Said Taghadouini, Alexis Gallagher, Raja Biswas, Faisal Ladhak, Tom Aarsen, Nathan Cooper, Griffin Adams, Jeremy Howard, and Iacopo Poli.
\newblock Smarter, better, faster, longer: A modern bidirectional encoder for fast, memory efficient, and long context finetuning and inference, 2024.

\bibitem{wei2023chainofthoughtpromptingelicitsreasoning}
Jason Wei, Xuezhi Wang, Dale Schuurmans, Maarten Bosma, Brian Ichter, Fei Xia, Ed~Chi, Quoc Le, and Denny Zhou.
\newblock Chain-of-thought prompting elicits reasoning in large language models, 2023.

\bibitem{xu2024perfectblendredefiningrlhf}
Tengyu Xu, Eryk Helenowski, Karthik~Abinav Sankararaman, Di~Jin, Kaiyan Peng, Eric Han, Shaoliang Nie, Chen Zhu, Hejia Zhang, Wenxuan Zhou, Zhouhao Zeng, Yun He, Karishma Mandyam, Arya Talabzadeh, Madian Khabsa, Gabriel Cohen, Yuandong Tian, Hao Ma, Sinong Wang, and Han Fang.
\newblock The perfect blend: Redefining rlhf with mixture of judges, 2024.

\bibitem{yang2024qwen25mathtechnicalreportmathematical}
An~Yang, Beichen Zhang, Binyuan Hui, Bofei Gao, Bowen Yu, Chengpeng Li, Dayiheng Liu, Jianhong Tu, Jingren Zhou, Junyang Lin, Keming Lu, Mingfeng Xue, Runji Lin, Tianyu Liu, Xingzhang Ren, and Zhenru Zhang.
\newblock Qwen2.5-math technical report: Toward mathematical expert model via self-improvement.
\newblock {\em arXiv preprint arXiv:2409.12122}, 2024.

\bibitem{ying2015decision}
LU~Ying et~al.
\newblock Decision tree methods: applications for classification and prediction.
\newblock {\em Shanghai archives of psychiatry}, 27(2):130, 2015.

\bibitem{yuan2024implicitprm}
Lifan Yuan, Wendi Li, Huayu Chen, Ganqu Cui, Ning Ding, Kaiyan Zhang, Bowen Zhou, Zhiyuan Liu, and Hao Peng.
\newblock Free process rewards without process labels.
\newblock {\em arXiv preprint arXiv:2412.01981}, 2024.

\bibitem{zhang2024generativeverifiersrewardmodeling}
Lunjun Zhang, Arian Hosseini, Hritik Bansal, Mehran Kazemi, Aviral Kumar, and Rishabh Agarwal.
\newblock {Generative Verifiers: Reward Modeling as Next-Token Prediction}, 2024.

\bibitem{zhao2025sample}
Eric Zhao, Pranjal Awasthi, and Sreenivas Gollapudi.
\newblock Sample, scrutinize and scale: Effective inference-time search by scaling verification.
\newblock {\em arXiv preprint arXiv:2502.01839}, 2025.

\bibitem{zheng2022minif2fcrosssystembenchmarkformal}
Kunhao Zheng, Jesse~Michael Han, and Stanislas Polu.
\newblock Minif2f: a cross-system benchmark for formal olympiad-level mathematics, 2022.

\bibitem{zheng2023judging}
Lianmin Zheng, Wei-Lin Chiang, Ying Sheng, Siyuan Zhuang, Zhanghao Wu, Yonghao Zhuang, Zi~Lin, Zhuohan Li, Dacheng Li, Eric.~P Xing, Hao Zhang, Joseph~E. Gonzalez, and Ion Stoica.
\newblock Judging llm-as-a-judge with mt-bench and chatbot arena, 2023.

\bibitem{zheng2023evaluation}
Lianmin Zheng, Dacheng Xu, Jiajun Dong, Andy Zeng, Shuo Xie, Eric~P Xing, and Percy Liang.
\newblock {Evaluation Biases for Large Language Models}.
\newblock {\em ArXiv Preprint arXiv:2305.17926}, 2023.

\end{thebibliography}
